\documentclass[nohyperref]{article}

\usepackage{microtype}
\usepackage{graphicx}
\usepackage{subfigure}
\usepackage{booktabs} 

\usepackage{multirow}

\usepackage{hyperref}

\usepackage[accepted]{icml2022}

\usepackage{amsmath}
\usepackage{amssymb}
\usepackage{mathtools}
\usepackage{amsthm}

\usepackage[capitalize,noabbrev]{cleveref}

\usepackage[textsize=tiny]{todonotes}

\icmltitlerunning{HyperTransformer: Model Generation for Few-Shot Learning}

\usepackage{amsmath,amsfonts,bm}

\def\eqref#1{equation~\ref{#1}}

\def\1{\bm{1}}

\def\vtheta{{\bm{\theta}}}

\def\vb{{\bm{b}}}

\def\ve{{\bm{e}}}

\def\vy{{\bm{y}}}

\def\mK{{\bm{K}}}

\def\mQ{{\bm{Q}}}

\def\mV{{\bm{V}}}
\def\mW{{\bm{W}}}

\DeclareMathAlphabet{\mathsfit}{\encodingdefault}{\sfdefault}{m}{sl}
\SetMathAlphabet{\mathsfit}{bold}{\encodingdefault}{\sfdefault}{bx}{n}

\newcommand{\E}{\mathbb{E}}

\newcommand{\R}{\mathbb{R}}

\DeclareMathOperator*{\argmin}{arg\,min}

\begin{document}

\raggedbottom

\twocolumn[
\icmltitle{HyperTransformer: Model Generation for Supervised and Semi-Supervised Few-Shot Learning}




\icmlsetsymbol{equal}{*}

\begin{icmlauthorlist}
\icmlauthor{Andrey Zhmoginov}{comp}
\icmlauthor{Mark Sandler}{comp}
\icmlauthor{Max Vladymyrov}{comp}
\end{icmlauthorlist}

\icmlaffiliation{comp}{Google Research}

\icmlcorrespondingauthor{Andrey Zhmoginov}{azhmogin@google.com}

\icmlkeywords{few-shot learning; Transformer model}

\vskip 0.3in
]



\printAffiliationsAndNotice{}
\smallskip

\begin{abstract}
    In this work we propose a {\em HyperTransformer}, a Transformer-based model for supervised and semi-supervised few-shot learning that generates weights of a convolutional neural network (CNN) directly from support samples.
    Since the dependence of a small generated CNN model on a specific task is encoded by a high-capacity Transformer model, we effectively decouple the complexity of the large task space from the complexity of individual tasks.
    Our method is particularly effective for small target CNN architectures where learning a fixed universal task-independent embedding is not optimal and better performance is attained when the information about the task can modulate all model parameters.
    For larger models we discover that generating the last layer alone allows us to produce competitive or better results than those obtained with state-of-the-art methods while being end-to-end differentiable.
\end{abstract}
\smallskip

\definecolor{brickred}{RGB}{203,65,84}

\def\maml{\textsc{MAML}}
\def\mamlpp{\textsc{MAML++}}
\def\rfs{\textsc{RFS}}
\def\omni{\textsc{Omniglot}}
\def\miniim{\textsc{miniImageNet}}
\def\tiered{\textsc{tieredImageNet}}
\def\task{\tau}
\def\loss{L}
\def\tasks{\mathcal{T}}
\def\trtasks{\mathcal{T}_{\rm{train}}}
\def\tstasks{\mathcal{T}_{\rm{test}}}
\def\vphi{{\bm{\phi}}}
\def\hyper{\textsc{HT}}
\def\rfs{\textsc{RFS}}
\def\numexamples{t}
\def\numclasses{n}
\def\numshot{k}
\def\embeddingdim{d}

\def\smallskip{}
\def\tinyskip{}

\newcommand{\alert}[1]{{\color{brickred} #1}}

\smallskip
\section{Introduction}
    In few-shot learning, a conventional machine learning paradigm of fitting a parametric model to training data is taken to a limit of extreme data scarcity where entire categories are introduced with just one or few examples.
    A generic approach to solving this problem uses training data to identify parameters $\vphi$ of a {\em solver} $a_\vphi$ that given a small batch of examples for a particular task (called a {\em support} set) can solve this task on unseen data (called a {\em query} set).
    
    One broad family of few-shot image classification methods frequently referred to as {\em metric-based learning}, relies on pretraining an embedding $\ve_\vphi(\cdot)$ and then using some distance in the embedding space to label query samples based on their closeness to known labeled support samples.
    These methods proved effective on numerous benchmarks (see \citet{tian2020rethinking} for review and references), however the capabilities of the solver are limited by the capacity of the architecture itself, as these methods try to build a universal embedding function. 
    
    On the other hand, {\em optimization-based methods} such as seminal \maml{} algorithm \citep{finn2017maml} can fine-tune the embedding $\ve_\vphi$ by performing additional SGD updates on all parameters $\vphi$ of the model producing it.
    This partially addresses the constraints of metric-based methods by learning a new embedding for each new task.
    However, in many of these methods, all the knowledge extracted during training on different tasks and describing the solver $a_\vphi$ still has to ``fit'' into the same number of parameters as the model itself.
    Such limitation becomes more severe as the target models get smaller, while the richness of the task set increases.
    
    In this paper we propose a new few-shot learning approach that allows us to decouple the complexity of the {\em task space} from the complexity of individual tasks.
    The main idea is to use the Transformer model \citep{vaswani2017attention} that given a few-shot task episode, generates an entire inference model by producing all model weights in a single pass.
    This allows us to encode the intricacies of the available training data inside the Transformer model, while producing specialized tiny models for a given individual task.
    Reducing the size of the generated model and moving the computational overhead to the Transformer-based weight generator, we can lower the cost of the inference on new images.
    This can reduce the overall computation cost in cases where the tasks change infrequently and hence the weight generator is only used sporadically.
    Note that here we follow the {\em inductive inference} paradigm with test samples processed one-by-one (by the generated inference model) and do not target other settings like, for example, {\em transductive inference} \cite{liu2019transductive} that consider relationships between test samples.

    We start by observing that the self-attention mechanism is well suited to be an underlying mechanism for a few-shot CNN weight generator.
    In contrast with earlier CNN- \citep{zhao2020meta} or BiLSTM-based approaches \citep{ravi2017optimization}, the vanilla\footnote{without attention masking or positional encodings} Transformer model is invariant to sample permutations and can handle unbalanced datasets with a varying number of samples per category.
    Furthermore, we demonstrate that a single-layer self-attention model can replicate a simplified gradient-descent-based learning algorithm.
    Using a Transformer to generate the logits layer on top of a conventionally end-to-end learned embedding, we achieve competitive results on several common few-shot learning benchmarks.
    For smaller generated CNN models, our approach shows significantly better performance than \mamlpp{} \cite{antoniou2019maml} and \rfs{} \cite{tian2020rethinking}, while also closely matching the performance of many state-of-the-art methods for larger CNN models.
    Varying Transformer parameters we demonstrate that this high performance can be attributed to additional capacity of the Transformer model that decouples its complexity from that of the generated CNN.
    While this additional capacity proves to be very advantageous for smaller generated models, larger CNNs can accommodate sufficiently complex representations and our approach does not provide a clear advantage compared to other methods in this case.

    We additionally can extend our method to support unlabeled samples by appending a special input token that encodes unknown classes to all unlabeled examples.
    In our experiments outlined in Section~\ref{sec:semisupervised}, we observe that adding unlabeled samples can significantly improve model performance. Interestingly, the full benefit of using additional data is only realized if the Transformers use two or more layers.
    This result is consistent with the basic mechanism described in Section~\ref{sec:reasoning}, where we show that a Transformer model with at least two layers can encode the nearest-neighbor style algorithm that associates unlabeled samples with similar labeled examples. 
    In essence, by training the weight generator to produce CNN models with best possible performance on a query set, we teach the Transformer to utilize unlabeled samples without having to manually introduce additional optimization objectives.

    We also explore the capability of our approach to generate all weights of the CNN model, adjusting both the logits layer and all intermediate layers producing the sample embedding.
    We show that by generating all layers we can improve both the training and test accuracies of CNN models below a certain size.
    Above this model size threshold, however, generation of the logits layer alone on top of a episode-agnostic embedding appears to be sufficient for reaching peak performance (see Figure~\ref{fig:tiny}).
    This threshold is expected to depend on the variability and the complexity of the training tasks.

    Another important advantage of our method is that it allows to do learning end-to-end without relying on complex nested gradients optimization and other meta-learning approaches, where the number of unrolls steps is large. Our optimization is done in a single loop of updates to the Transformer (and feature extractor) parameters.
    The code for the paper can be found at {{\href{https://github.com/google-research/google-research/tree/master/hypertransformer}{https://github.com/google-research/google-research/tree/master/hypertransformer}}}.

\smallskip
\section{Related work}

    Few-shot learning received a lot of attention from the deep learning community and while there are hundreds of few-shot learning methods, several common themes emerged in the past years.
    Here we outline several existing approaches and show how they relate to our method.

    \smallskip
    \paragraph{Metric-Based Learning.}
    One family of approaches involves mapping input samples into an embedding space and then using some nearest neighbor algorithm to label query samples based on the distances from their embeddings to embeddings of labeled support samples.
    The metric used to compute the distances can either be the same for all tasks, or can be task-dependent.
    This family of methods includes, for example, such methods as Siamese networks \citep{koch2015siamese}, Matching Networks \citep{vinyals2016matching}, Prototypical Networks \citep{snell2017prototypical}, Relation Networks \citep{sung2018relation} and TADAM \citep{oreshkin2018tadam}.
    It has recently been argued \citep{tian2020rethinking} that methods based on building a powerful sample representation can frequently outperform numerous other approaches including many optimization-based methods.
    However, such approaches essentially amount to the ``one-model solves all'' approach and thus require larger models than needed to solve individual tasks.

    \smallskip
    \paragraph{Optimization-Based Learning.}
    An alternative approach that can adapt the embedding to a new task is to incorporate optimization within the learning process.
    A variety of such methods are based on the approach called {\em Model-Agnostic Meta-Learning}, or \maml{} \citep{finn2017maml}.
    The core idea of \maml{} is learning initial model parameters $\vtheta_0$ that produce good models for each episode after being adjusted with one or more gradient descent updates minimizing the corresponding episode classification loss.
    This approach was later refined \citep{antoniou2019maml} and built upon giving rise to Reptile \citep{nichol2018meta}, LEO \citep{rusu2019metalearning} and others.
    One limitation of various \maml{}-inspired methods is that the knowledge about the set of training tasks $\trtasks$ is distilled into parameters $\vphi=\vtheta_0$ that have the same dimensionality as the model parameters.
    Therefore, for a very lightweight model $f(x;\vtheta)$ the capacity of the solver $a_{\vphi}$ producing model weights from the support set is still limited by the size of $\vtheta$.
    Methods that use parameterized preconditioners that otherwise do not impact the model $f(x;\vtheta)$ can alleviate this issue, but as with \maml{}, such methods can be difficult to train \citep{antoniou2019maml}.
    
    \smallskip
    \paragraph{Weight Modulation and Generation.}
    The idea of using a task specification to directly generate or modulate model weights has been previously explored in the generalized supervised learning context \citep{requeima2019conditional, ratzlaff2019hypergan}, few-shot learning \citep{guo2020attentive} and in specific language models \citep{pilault2021mtlearning, mahabadi2021finetuning, tay2021hypergrid, ye2021adapters}.
    Some few-shot learning methods described above also employ this approach and use task-specific generation or modulation of the weights of the final classification model.
    For example, in LGM-Net \citep{li2019lgm} the matching network approach is used to generate a few layers on top of a task-agnostic embedding.
    Another approach abbreviated as LEO \citep{rusu2019metalearning} utilized a similar weight generation method to generate initial model weights from the training dataset in a few-shot learning setting, much like what is proposed in this article.
    However, in \citet{rusu2019metalearning}, the generated weights were also refined using several SGD steps similar to how it is done in \maml{}.
    Here we explore a similar idea, but largely inspired by the \textsc{HyperNetwork} approach \citep{ha2016hypernetworks}, we instead propose to directly generate an entire task-specific CNN model.
    Unlike LEO, we do not rely on pre-computed embeddings for images and generate the model in a single step without additional SGD steps, which simplifies and stabilizes training.
    
    \smallskip
    \paragraph{Transformers in Computer Vision and Few-Shot Learning.}
    Transformer models \citep{vaswani2017attention} originally proposed for NLP applications, had since become a useful tool in practically every field of deep learning.
    In computer vision, Transformers have recently seen an explosion of applications ranging from state-of-the-art classification results \citep{dosovitskiy2021image,touvron2021deit} to object detection \citep{carion2020detection,zhu2020detr}, segmentation \citep{ye2019segmentation}, image super-resolution \citep{yang2020super}, image generation \citep{chen2021} and many others.
    There are also several notable applications in few-shot image classification.
    For example, in \citet{liu2021universal}, the Transformer model was used for generating universal representations in the multi-domain few-shot learning scenario.
    And closely related to our approach, in \citet{ye2020adaptation}, the authors proposed to accomplish embedding adaptation with the help of Transformer models.
    Unlike our method that generates an entire end-to-end image classification model, this approach applied a task-dependent perturbation to an embedding generated by an independent task-agnostic feature extractor.
    In \citet{gidaris2018dynamic}, a simplified attention-based model was used for the final layer generation.

\smallskip
\section{Analytical Framework}

    Here we establish a general framework that includes few-shot learning as a special case, but allows us to extend it to cases when more information is available beyond few supervised samples, e.g.\ using additional unlabeled data.

    \smallskip
    \subsection{Learning from Generalized Task Descriptions}

        Consider a set of tasks $\{t|t\in \tasks\}$ each of which is associated with a loss $\mathcal{L}(f;t)$ that quantifies the correctness of any model $f$ attempting to solve $t$.
        A task can be associated with a classification, regression, learning a reinforcement learning policy or any other kind of problem.
        Along with the loss, each task also is characterized by a {\em task description} $\tau(t)$ that is sufficient for communicating this task and finding the optimal model that solves it.
        This task description can include any available information about $t$, like labeled and unlabeled samples, image metadata, textual descriptions, etc.
        
        The weight generation algorithm can then be viewed as a method of using a set of training tasks $\trtasks$ for discovering a particular {\em solver} $a_\vphi$ that given $\tau(t)$ for a task $t$ similar to those present in the training set, produces an optimal model $f_*=a_\vphi(\tau)\in \mathcal{F}$ minimizing $\mathcal{L}(f_*,t)$.
        In this paper, we learn $a_\vphi$ by performing gradient-descent optimization of
        \tinyskip
        \begin{gather}
            \label{eq:core}
            \argmin_{\vphi\in \Phi} \E_{t\sim p(t)} \mathcal{L}(a_\vphi(\task(t)),t)
        \end{gather}
        \tinyskip
        with $p(t)$ being the distribution of training tasks from $\trtasks$.

    \smallskip
    \subsection{Special Case of Few-Shot Learning}
        \label{sec:few-shot}
        
        Few-shot learning is a special case of the framework described above.
        In few-shot learning, the loss $\mathcal{L}_t$ of a task $t$ is defined by a labeled {\em query set} $Q(t)$.
        The task description $\task(t)$ is then specified via a {\em support set} of examples.
        In a classical ``$\numclasses$-way-$\numshot$-shot'' setting, each training task $t\in \trtasks$ is sampled by first randomly choosing $\numclasses$ distinct classes $C_t$ from a large training dataset and then sampling examples without replacement from these classes to generate $\task(t)$ and $Q(t)$.
        The support set $\task(t)$ in this setting always contains $\numshot$ labeled samples $\{x_i^{(c)}\in X|i\in [1,\numshot]\}$ for each of classes $c \in C_t$.

        The quality of a particular few-shot learning algorithm is evaluated using a separate test space of tasks $\tstasks$.
        By forming $\tstasks$ from classes unseen at training time, we can evaluate generalization of the trained solver $a_\vphi$ by computing accuracies of models $a_\vphi(t)$ for $t \in \tstasks$.
        Best algorithms are expected to capture the structure present in the training set, extrapolating it to novel, but related tasks.

        Equation~\ref{eq:core} describes the general framework for learning to solve tasks given their descriptions $\task(t)$.
        When $\task$ is given by supervised samples, we recover classic few-shot learning.
        But the freedom in the definition of $\tau$ permits us, for example, to extend the problem to a semi-supervised regime \cite{ren2018semi}, assuming that each $\tau(t)$ contains both labeled and unlabeled examples.
        The approach relying on solving \eqref{eq:core} can be contrasted with classical approaches that typically have to modify their algorithms and optimization objectives in response to any additional type of information supplied in the task specification $\task$.
        For example, if $\task$ contains unlabeled examples, representation-based approaches could use unlabeled samples to make more accurate estimates of embedding centroids for each class, effectively trying to infer the distribution of samples in $Q(t)$.
        Optimization-based methods like MAML would have to introduce new optimization objectives on unlabeled samples in addition to the cross-entropy loss on labeled samples.
        In contrast, our algorithm is able to learn from $\task$ directly.

\smallskip
\section{HyperTransformer}

    \begin{figure}
        \centering
        \includegraphics[width=0.48\textwidth]{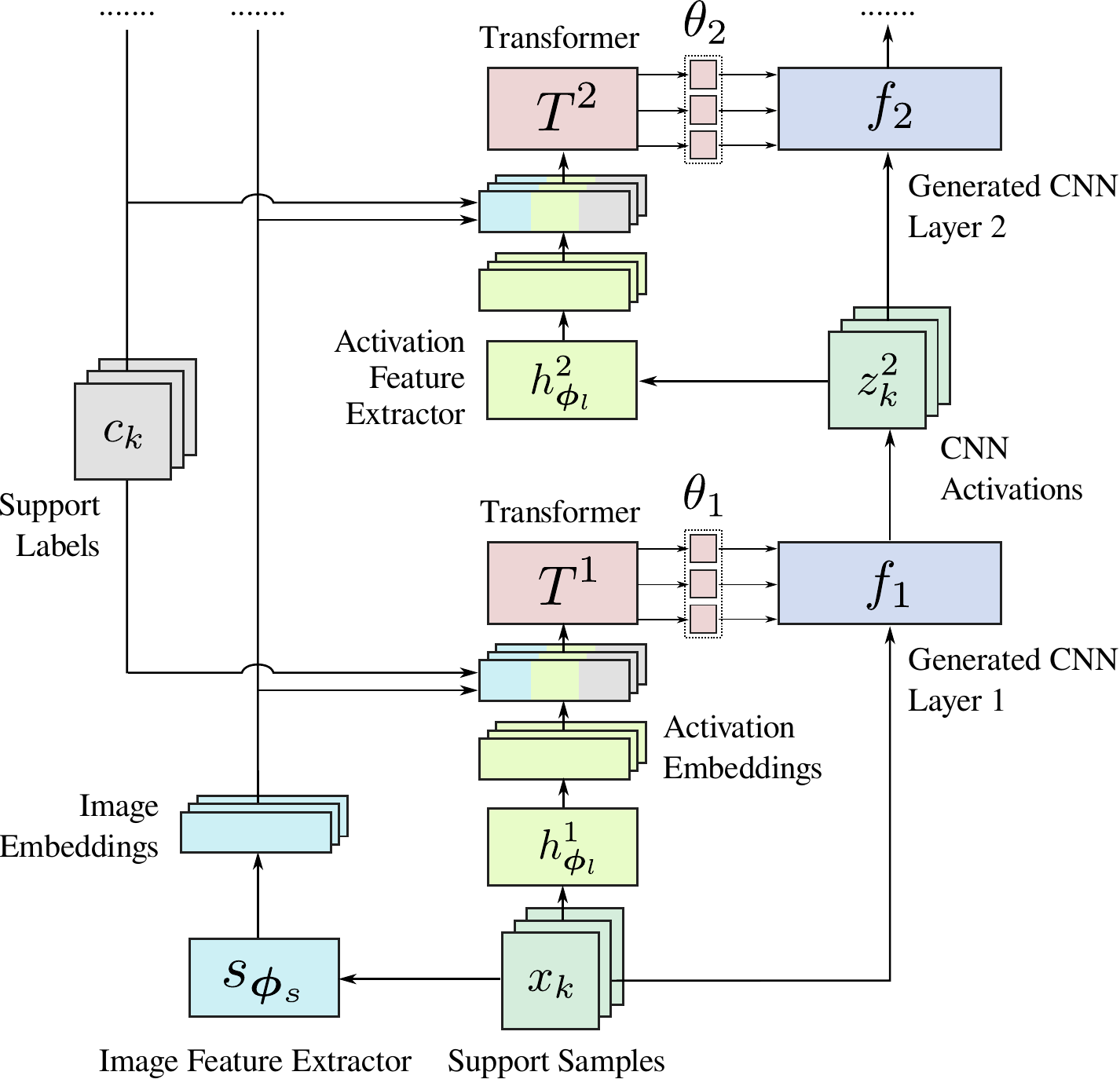}
        \smallskip
        \caption{
            \label{fig:model}
            A diagram of our model showing the generation of two CNN layers: Transformer-based weight generators receive {\em image embeddings} $s_{\phi_s}(\cdot)$ and {\em activation embeddings} $h_{\phi_l}(\cdot)$ along with corresponding labels $c_i$, and produce CNN layer weights ($\theta_1$ and $\theta_2$). After being generated, the CNN model is used to compute the loss on the query set. The gradients of this loss are then used to adjust the weights of the entire weight generation model ($\phi_s$, $\phi_l$, Transformer weights).
            \tinyskip
        }
    \end{figure}

    In a special case of a one-to-one mapping from $t$ to $\tau(t)$, it is generally possible to find $\theta(\tau)$ minimizing $\mathcal{L}(\theta(\tau(t)),t)$ numerically or analytically by solving an ordinary differential equation that effectively ``tracks'' the local minimum of $\mathcal{L}$ along an arbitrary curve $\hat{t}:[0,1]\to \tasks$ in the task space (see Appendix~\ref{app:wg}):
    \begin{gather*}
        \frac{d \theta}{d\gamma} = 
        -\left( \frac{\partial^2 \mathcal{L}}{\partial \theta^2} \right)^{-1}
        \frac{\partial^2 \mathcal{L}}{\partial \theta \, \partial t}
        \frac{d \hat{t}}{d\gamma},
    \end{gather*}
    where $\gamma\in [0,1]$ is a coordinate along the curve, $\theta(\gamma) := \theta(\tau(\hat{t}(\gamma))$, and all derivatives are computed at $\hat{t}(\gamma)$, $\theta(\gamma)$.
 
    Empirical solution of \eqref{eq:core} for $a_\vphi(\tau)$ represented by a deep neural network can be obtained by solving this optimization problem directly.
    In this section, we describe the design of the model $a_\vphi(\tau)$ that we call a \textsc{HyperTransformer} (\hyper{}).
    Choosing Transformer as the core component of \hyper{}, we make it possible for $a_\vphi$ to process any complex multi-modal task description $\tau(t)$ assuming that it can be encoded as an unordered set of Transformer tokens.

    \smallskip
	\subsection{Few-Shot Learning Model}
	    \label{sec:model}

		A solver $a_\vphi$ is the core of a few-shot learning algorithm since it encodes the knowledge of the training task distribution within its weights $\vphi$.
		We choose $a_\vphi$ to be a Transformer-based model (see Fig.~\ref{fig:model}) that takes a task description $\task$ containing the information about labeled and unlabeled support-set samples as input and produces weights for some or all layers $\{ \theta_\ell | \ell\in [1,L]\}$ of the generated CNN model.
		Layer weights that are not generated are instead {\em learned} end-to-end together with \hyper{} weights as ordinary task-agnostic variables.
		In other words, these learned layers are modified during the training phase and remain static during the evaluation phase (i.e.\ not dependent of the support set).
		In our experiments generated CNN models contain a set of convolutional layers and a final fully-connected logits layer.
		Here $\theta_\ell$ are the parameters of the $\ell$-th layer and $L$ is the total number of layers including the final logits layer (with index $L$).
		The weights are generated layer-by-layer starting from the first layer: $\theta_1(\task) \to \theta_2(\theta_1;\task) \to \dots \to \theta_L(\theta_{1,\dots,L-1};\task)$.
		Here we use $\theta_{a,\dots,b}$ as a short notation for $(\theta_a,\theta_{a+1},\dots,\theta_{b})$.

        \smallskip
		\paragraph{Image and activation embeddings.}
		The weights for the layer $\ell$ are either: (a) simply learned as a task-agnostic trainable variable, or (b) generated by the Transformer that receives a concatenation of {\em image embeddings}, {\em activation embeddings} and support sample labels $c_i$:
 		\begin{gather*}
 			\mathcal{I}^\ell := \left\{ \Big( \,\, s_{\vphi_s}(x_i), \,\, h_{\vphi_l}^{\ell}\big(z_i^\ell\big), \,\, c_i \,\, \Big) \right\}_{i = 1,\dots,\numshot \numclasses}.
 		\end{gather*}
		The {\em activation embeddings} at layer $\ell$ are produced by a convolutional {\em feature extractor} $h_{\vphi_l}^{\ell}(z^\ell_i)$ applied to the activations of the previous layer $z_i^\ell := f_{\ell-1}(x_i;\theta_{1,\dots,\ell-1})$ for $\ell>1$ and $z_i^1 := x_i$.
		The intuition behind using the activation embeddings is that the choice of the layer weights should primarily depend on the inputs received by this layer.
		
    \begin{figure}
        \centering
        \includegraphics[width=0.45\textwidth]{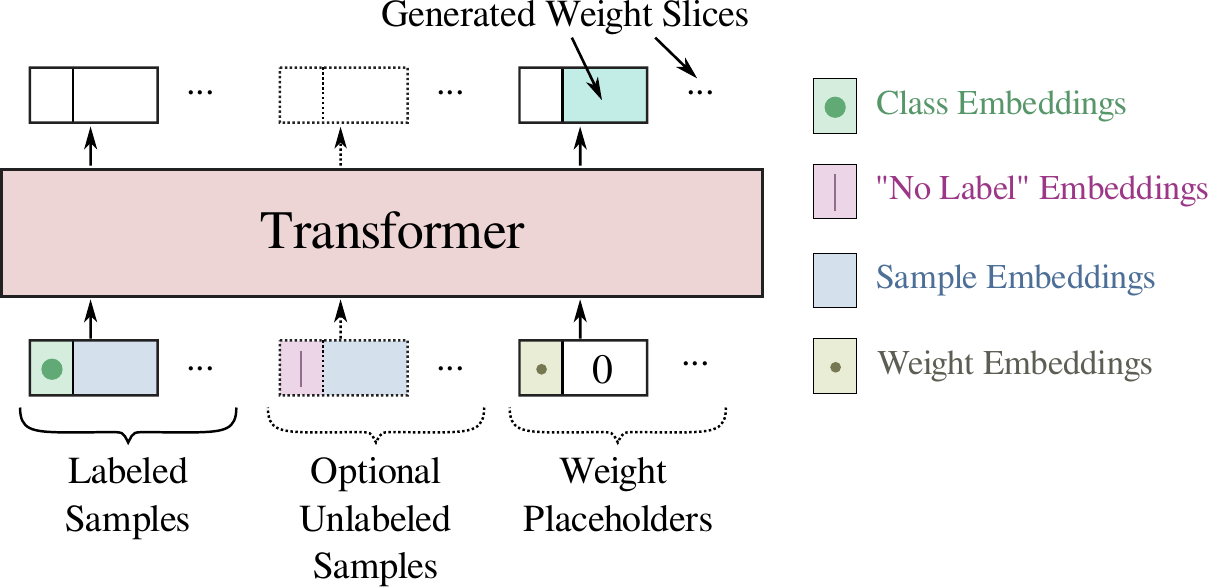}
        \smallskip
        \caption{
            \label{fig:tokens}
            Structure of the tokens passed to and received from a Transformer model.
            Both labeled and unlabeled (optional) samples can be passed to the Transformer as inputs.
            Empty output tokens indicate ignored outputs.
        }
        \smallskip
    \end{figure}

		The {\em image embeddings} are produced by a separate trainable convolutional neural network $s_{\vphi_s}(x_i)$ that is shared by all the layers.
		Their purpose are to modulate each layer's weight generator with a global high-level view of the sample that, unlike the activation embedding, is independent of the generated weights and is shared between generators.

        \smallskip
		\paragraph{Encoding and decoding Transformer inputs and outputs.}
		In the majority of our experiments, the input samples were encoded by concatenating image and activation embeddings from $\mathcal{I}^\ell$ to trainable label embeddings $\xi(c)$ with $\xi:[1, \numclasses]\to \R^\embeddingdim$.
		Here $\numclasses$ is the number of classes per episode and $\embeddingdim$ is a chosen size of the label encoding.
		Note that the class embeddings do not contain semantic information, but rather act as placeholders to differentiate between distinct classes.
		In addition to supervised few-shot learning, we also considered a semi-supervised scenario when some of the support samples are provided without the associated class information.
		Such unlabeled samples were fed into the Transformer using the same general encoding approach, but we used an auxiliary learned ``unlabeled'' token $\hat{\xi}$ in place of the label encoding $\xi(c)$ to indicate the fact that the class of the sample is unknown.
		
		Along with the input samples, the sequence passed to the Transformer was also populated with special learnable placeholder tokens, each associated with a particular slice of the to-be-generated weight tensor.
		Each such token was a learnable $d$-dimensional vector padded with zeros to the size of the input sample token.
		After the entire input sequence was processed by the Transformer, we read out model outputs associated with the weight slice placeholder tokens and assembled output weight slices into the final weight tensors (see Fig.~\ref{fig:tokens}).
		
		In our experiments we considered two different ways of encoding $k\times k \times n_{\rm input} \times n_{\rm output}$ convolutional kernels: (a) {\em ``output allocation''} generates $n_{\rm output}$ tokens with weight slices of size $k^2 \times n_{\rm input}$ and (b) {\em ``spatial allocation''} generates $k^2$ weight slices of size $n_{\rm input} \times n_{\rm output}$.
		We show comparison results in Supplementary Materials.
		
		\smallskip
		\paragraph{Training the model.}
		The weight generation model uses the support set to produce the weights of some or all CNN model layers.
		Then, the cross-entropy loss is computed for the query set samples that are passed through the generated CNN model.
		The weight generation parameters $\vphi$ (including the Transformer model and image/activation feature extractor weights) are learned by optimizing this loss functwion using stochastic gradient descent. 

        \begin{table*}[h]
        	\small
            \begin{center} \begin{tabular}{l|l|ccccc|ccccc}
                \multirow{2}{7em}{\em Dataset} & \multirow{2}{2em}{\em Approach} & \multicolumn{5}{|c}{\em 1-shot (channels)} & \multicolumn{5}{|c}{\em 5-shot (channels)} \\
                & & $8$ & $16$ & $32$ & $48$ & $64$ & $8$ & $16$ & $32$ & $48$ & $64$ \\ \hline
                \multirow{2}{7em}{\omni} & \small \mamlpp & $81.4$ & $88.6$ & $\mathbf{95.6}$ & $\mathbf{95.8}$ & $\mathbf{97.7}^\dagger$ & $83.2$ & $94.9$ & $\mathbf{98.6}$ & $\mathbf{98.8}$ & $\mathbf{99.3}^\dagger$ \\
                & \small \hyper & $\mathbf{87.2}$ & $\mathbf{93.7}$ & $\mathbf{95.5}$ & $\mathbf{95.7}$ & $96.2$ & $\mathbf{94.7}$ & $\mathbf{98.0}$ & $\mathbf{98.6}$ & $\mathbf{98.8}$ & $98.8$ \\
                \hline
                \multirow{3}{8em}{\miniim{}} & 
                \small \mamlpp & $43.9$ & $46.6$ & $49.4$ & $52.2^\dagger$ & -- & $\mathbf{59.0}$ & $\mathbf{64.6}$ & $\mathbf{66.8}$ & $68.3^\dagger$ & -- \\
                & \small \rfs & $44.0$ & $49.4$ & $51.5$ & $54.2$ & -- & $56.1$ & $63.5$ & $\mathbf{67.1}$ & $\mathbf{69.1}$ & -- \\
                & \small \hyper & $\mathbf{45.5}$ & $\mathbf{50.2}$ & $\mathbf{53.8}$ & $\mathbf{55.1}$ & -- & $\mathbf{59.3}$ & $64.2$ & $\mathbf{67.1}$ & $68.1$ & -- \\
                \hline
                \multirow{2}{8em}{\tiered{}} & 
                \small \rfs & $44.1$ & $47.7$ & $51.5$ & $54.6$ & $\mathbf{56.8}^\star$ & $55.5$ & $62.0$ & $66.3$ & $69.3$ & $73.2^\star$ \\
                & \small \hyper & $\mathbf{49.1}$ & $\mathbf{51.9}$ & $\mathbf{54.0}$ & $\mathbf{55.0}$ & $56.3$ & $\mathbf{61.9}$ & $\mathbf{65.8}$ & $\mathbf{70.2}$ & $\mathbf{71.1}$ & $\mathbf{73.9}$ \\
            \end{tabular} \end{center}
            \smallskip
            \caption{
                \label{tab:supervised}
        		Comparison of \hyper{} with \mamlpp{} and \rfs{} on models of different sizes and different datasets: (a) 20-way \omni{}, (b) 5-way \miniim{} and (c) 5-way \tiered{}.
        		Results for \mamlpp{} and \rfs{} were obtained using GitHub codes accompanying \citet{antoniou2019maml} and \citet{tian2020rethinking} correspondingly (results marked with $\dagger$ and $\star$ were taken from corresponding papers).
        		\hyper{} generally outperforms both \mamlpp{} and \rfs{} for smaller models.
        		Accuracy confidence intervals: \omni{} -- between $0.1\%$ and $0.3\%$, \miniim{} and \tiered{} -- between $0.2\%$ and $0.5\%$.
        	}
        \smallskip
        \end{table*}
        
    \smallskip
    \subsection{Reasoning Behind the Self-Attention Mechanism}
        \label{sec:reasoning}

        The choice of self-attention mechanism for the weight generator is not random.
        One reason behind this choice is that the output produced by generator with the basic self-attention is by design invariant to input permutations, i.e., permutations of samples in the training dataset.
        This also makes it suitable for processing unbalanced batches and batches with a variable number of samples (see Sec.~\ref{sec:semisupervised}).
        Now we show that the calculation performed by a self-attention model with properly chosen parameters can mimic basic few-shot learning algorithms further motivating its utility. 

        \smallskip
        \paragraph{Supervised learning.}
        Self-attention in its rudimentary form can implement a method similar to cosine-similarity-based sample weighting encoded in the logits layer\footnote{we assume that the embeddings $\ve$ are unbiased, i.e., $\langle e_i \rangle=0$} with weights $W_{i,j} \sim \sum_{m=1}^{n} y_i^{(m)} e_j^{(m)}$ 
        which can also be viewed as a result of applying a single gradient descent step on the cross-entropy loss (see Appendix~\ref{app:unlabeled}).
    	Here $n$ is the total number of support-set samples $\{x^{(m)}|m\in [1,n]\}$ and $\ve^{(m)}$, $\vy^{(m)}$ are the embedding vector and the one-hot label corresponding to $x^{(m)}$.
        
        The approach can be outlined (see more details in Appendix~\ref{app:unlabeled}) as follows.
        The self-attention operation receives encoded input samples $\mathcal{I}_k=(\xi(c_k),\ve_k)$ and weight placeholders $(\mu(i),0)$ as its input.
        If each weight slice $W_{i,\cdot}$ represented by a particular token $(\mu(i),0)$ produces a {\em query} $Q_i$ that only attends to {\em keys} $K_k$ corresponding\footnote{in other words, the self-attention layer should match tokens $(\mu(i),0)$ with $(\xi(i),\dots)$.} to samples $\mathcal{I}_k$ with labels $c_k$ matching $i$ and the {\em values} of these samples are set to their embeddings $\ve_k$, then the self-attention operation will essentially average the embeddings of all samples assigned label $i$ thus matching the first term in $\mW$.
    
        \smallskip
        \paragraph{Semi-supervised learning.}
    	A similar self-attention mechanism can also be designed to produce logits layer weights when the support set contains some unlabeled samples.
    	The proposed mechanism first propagates classes of labeled samples to similar unlabeled samples.
    	This can be achieved by a single self-attention layer choosing the {\em queries} and the {\em keys} of the samples to be proportional to their embeddings.
    	The attention map for sample $i$ would then be defined by a softmax of $\ve_i \cdot \ve_j$, or in other words would be proportional to $\exp(\ve_i \cdot \ve_j)$.
    	Choosing sample {\em values} to be proportional to the class tokens, we can then propagate a class of a labeled sample $\ve_j$ to a nearby unlabeled sample with embedding $\ve_i$, for which $\ve_i \cdot \ve_j$ is sufficiently large.
    	If the self-attention module is ``residual'', i.e., the output of the self-attention operation is added to the original input, like it is done in the Transformer model, then this additive update would essentially ``mark'' an unlabeled sample by the propagated class (albeit this term might have a small norm).
    	The second self-attention layer can then be designed similarly to the supervised case.
    	If label embeddings are orthogonal, then even a small component of a class embedding propagated to an unlabeled sample can be sufficient for a weight slice to attend to it thus adding its embedding to the final weight (resulting in the averaging of embeddings of both labeled and proper unlabeled examples).

\smallskip
\section{Experiments}

    In this section, we present \textsc{HyperTransformer} (\hyper{}) experimental results and discuss the implications of our empirical findings.

    \smallskip
    \subsection{Datasets and Setup}

        \paragraph{Datasets.}
        For our experiments, we chose several most widely used few-shot datasets including \omni{}, \miniim{} and \tiered{}.
        \miniim{} contains a relatively small set of labels and is arguably the simplest to overfit to.
        Because of this and since in many recent publications \miniim{} was replaced with a larger \tiered{} dataset, we conduct many of our experiments and ablation studies using \omni{} and \tiered{}.

        \begin{table*}[h]
        \renewcommand{\arraystretch}{1.1}
        \begin{center} \begin{tabular}{l|cc||l|cc||l|cc}
            \multicolumn{6}{c||}{\textsc{miniImageNet}} & \multicolumn{3}{c}{\textsc{tieredImageNet}} \\
            {\bf Method} & {\bf 1-S} & {\bf 5-S} & {\bf Method} & {\bf 1-S} & \bf{5-S} & {\bf Method} & {\bf 1-S} & {\bf 5-S} \\
            \hline
            \underline{\hyper} & $54.1$ & $68.5$ & \underline{\hyper-48} & $\mathbf{55.1}$ & $68.1$ & \underline{\hyper-32} & $\mathbf{54.0}$ & $\mathbf{70.2}$ \\
            MN & $43.6$ & $55.3$ & SAML & $52.2$ & $66.5$ & MAML-32 & $51.7$ & $\mathbf{70.3}$ \\ \cline{7-9}
            IMP & $49.2$ & $64.7$ & GCR & $53.2$ & $\mathbf{72.3}$ & \underline{\hyper} & $\mathbf{56.3}$ & $\mathbf{73.9}$ \\
            PN & $49.4$ & $68.2$ & KTN & $54.6$ & $71.2$ & PN & $53.3$ & $72.7$ \\
            MELR & $\mathbf{55.4}$ & $\mathbf{72.3}$ & PARN  & $\mathbf{55.2}$ & $71.6$ & MELR & $\mathbf{56.4}$ & $\mathbf{73.2}$ \\
            TAML & $51.8$ & $66.1$ & PPA & $54.5$ & $67.9$ & RN & $54.5$ & 71.3  \\
        \end{tabular} \end{center}
        \smallskip        
        \caption{
            \label{tab:comparison}
    		Comparison of \miniim{} and \tiered{} 1-shot (1-S) and 5-shot (5-S) 5-way results for \hyper{} (underlined) and other widely known methods with a 64-64-64-64 model including \citep{tian2020rethinking}: Matching Networks \citep{vinyals2016matching}, IMP \citep{allen2019infinite}, Prototypical Networks \citep{snell2017prototypical}, TAML \citep{jamal2019ml}, SAML \citep{hao2019collect}, GCR \citep{li2019global}, KTN \citep{peng2019fewshot}, PARN \citep{wu2019parn}, Predicting Parameters from Activations \citep{qiao2018ppa}, Relation Net \citep{sung2018relation}, MELR \citep{fei2021melr}.
    		We also include results for CNNs with fewer channels (``-32'' for 32-channel models, etc.).
    	}
    	\smallskip
        \end{table*}

        \smallskip
        \paragraph{Models.}
        \textsc{HyperTransformer} can in principle generate arbitrarily large weight tensors by producing low-dimensional embeddings that can then be fed into another trainable model to generate the entire weight tensors.
        In this work, however, we limit our experiments to \hyper{} models that generate weight tensor slices encoding individual output channels directly.
        For the target models we focus on 4-layer CNN architectures identical to those used in \mamlpp{} and numerous other papers. 
        More precisely, we used a sequence of four $3\times 3$ convolutional layers with the same number of output channels followed by batch normalization (BN) layers, nonlinearities and max-pooling stride-2 layers. 
        All BN variables were learned and not generated.
        Experiments with generated BN variables did not show much difference with this simpler approach.
        Generating larger architectures such as \textsc{ResNet} and \textsc{WideResNet} will be the subject of our future work.

        \smallskip
        \subsection{Supervised Results with Logits Layer Generation}
        \label{sec:supervised}
        As discussed in Section~\ref{sec:reasoning}, using a simple self-attention mechanism to generate the CNN logits layer can be a basis of a simple few-shot learning algorithm.
        Motivated by this observation, in our first experiments, we compared the proposed \hyper{} approach with \mamlpp{} and \rfs{} \cite{tian2020rethinking} on \omni{}, \miniim{} and \tiered{} datasets (see Table~\ref{tab:supervised}) with \hyper{} limited to generating only the final fully-connected logits layer.

        In our experiments, the dimensionality of the activation embedding was chosen to be the same as the number of model channels and the image embedding had a dimension of $32$ regardless of the model size.
        The image feature extractor was a simple 4-layer convolutional model with batch normalization and stride-$2$ $3\times 3$ convolutional kernels.
        The activation feature extractors were two-layer convolutional models with outputs of both layers averaged over the spatial dimensions and concatenated to produce the final activation embedding.
        For all tasks except 5-shot \miniim{} our Transformer had $3$ layers, used a simple sequence of encoder layers (Figure~\ref{fig:tokens}) and used the ``output allocation'' of weight slices (Section~\ref{sec:model}).
        Experiments with the encoder-decoder Transformer architecture can be found in Appendix~\ref{app:ablation}.
        The 5-shot \miniim{} and \tiered{} results presented in Table~\ref{tab:supervised} were obtained with a simplified Transformer model that had 1 layer, and did not have the final fully-connected layer and nonlinearity.
        This proved necessary for reducing model overfitting of this smaller dataset.
        Other model parameters are described in detail in Appendix~\ref{app:details}.

        Results obtained with our method in a few-shot setting (see Table~\ref{tab:supervised}) are frequently better than \mamlpp{} and \rfs{} results, especially on smaller models, which can be attributed to parameter disentanglement between the weight generator and the CNN model.
        While the improvement over \mamlpp{} and \rfs{} gets smaller with the growing size of the generated CNN, our results for large models appear to be comparable to those obtained with \mamlpp{}, \rfs{} and numerous other methods (see Table~\ref{tab:comparison}).
        Discussion of additional comparisons to LGM-Net \citep{li2019lgm} and LEO \citep{rusu2019metalearning} using a different setup (which is why they could not be included in Table~\ref{tab:comparison}) and showing an almost identical performance can be found in Appendix~\ref{app:supervised}.

        While the learned \hyper{} model could perform a relatively simple calculation on high-dimensional sample embeddings, our brief analysis of the parameters space (see Appendix~\ref{app:ablation}) shows that using simpler 1-layer Transformers leads to a modest decrease of the test accuracy and a greater drop in the training accuracy for smaller models.
        However, in our experiments with 5-shot \miniim{} dataset, which is generally more prone to overfitting, we observed that increasing the Transformer model complexity improves the model training accuracy (on episodes that only use classes seen at the training time), but the test accuracy relying on classes unseen at the training time, generally degrades.
        We also observed that the results in Table~\ref{tab:supervised} could be improved even further by increasing the embedding sizes (see Appendix~\ref{app:ablation}), but we did not pursue an exhaustive optimization in the parameter space.

        Note that overfitting characterized by a good performance on tasks composed of seen categories, but poor generalization to unseen categories, may still have practical applications for pesonalization.
        Specifically, if the actual task relies on classes seen at the training time, we can generate an accurate model customized to a particular task in a single pass without having to perform any SGD steps to fine-tune the model.
        This is useful if, for example, the client model needs to be adjusted to a particular set of known classes most widely used by this client.
        We also anticipate that with more complex data augmentations and additional synthetic tasks, more complex Transformer-based models can further improve their performance on the test set.

    \smallskip
    \subsection{Semi-Supervised Results}
        \label{sec:semisupervised}

        \begin{table}[b]
			\smallskip
			\begin{center} \scalebox{0.8}{\begin{tabular}{c|c|c|c|c|c|c|c}
                $(u,L_T)$ & {\em 1-shot} & {\em 5-shot} & $(2,3)$ & $(4,1)$ & $(4,2)$ & $(4,3)$ & $(9,3)$ \\
                \hline \\[-0.8em]
                Accuracy & $56.0$ & $69.9$ & $58.3$ & $56.6$ & $59.9$ & $59.9$ & $61.5$
            \end{tabular}} \end{center} \smallskip
            \caption{
                \label{tab:semi}            
                Test accuracy on \tiered{} of supervised 1-shot and 5-shot models and semi-supervised 1-shot models with $u$ additional unlabeled samples per class.
				The weight generation Transformer model uses $3$ encoder layer for supervised tasks and $L_T$ encoder layers in semi-supervised experiments.
				Notice a performance improvement of semi-supervised learning over the 1-shot supervised results.
				Accuracy is seen to grow with the number of unlabeled samples and the maximum accuracy is reached when the encoder has at least two layers.
			}
        \end{table}
        
        \begin{figure}[h]
            \centering
            \includegraphics[width=0.28\textwidth]{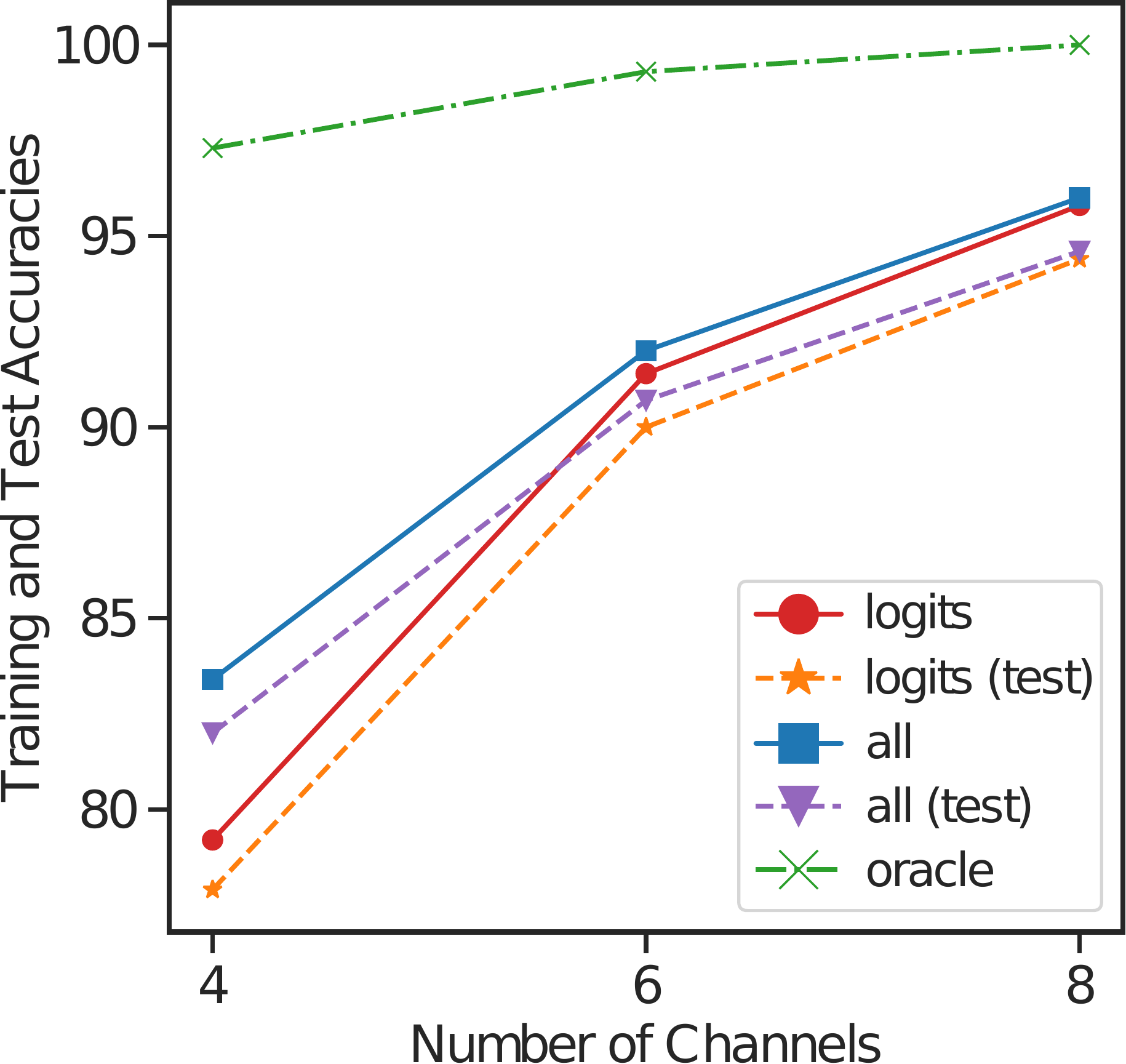}
            \smallskip
            \caption{
                \label{fig:tiny}
                5-shot-20-way \omni{} training/test accuracies as a function of the CNN model complexity:
                only the final logits layer being generated ({\em logits}),
                all layers being generated ({\em all}),
                training the model on all available samples for a random set of few classes ({\em oracle}).
                A model that generates CNN weights by memorizing all samples (being able to determine their classes) and also memorizing optimal trained weights for any selection of classes would reach the {\em oracle} accuracy, but would not generalize.
            }
            \smallskip
        \end{figure}

        In our approach, the weight generation model is trained by optimizing the query set loss and therefore any additional information about the task, including unlabeled samples, can be provided as a part of the task description $\task$ to the weight generator without having to alter the optimization objective.
        This allows us to tackle a semi-supervised few-shot learning problem without making any substantial changes to the model or the training approach.
        In our implementation, we simply added unlabeled samples into the support set and marked them with an auxiliary learned ``unlabeled'' token $\hat{\xi}$ in place of the label encoding $\xi(c)$.
        
        Since \omni{} is typically characterized by very high accuracies in the $97\%$--$99\%$ range, we conducted all our experiments with \tiered{}.
        As shown in Table~\ref{tab:semi}, adding unlabeled samples results in a substantial increase of the final test accuracy.
        Furthermore, notice that the model achieves its best performance when the number of Transformer layers is greater than one.
        This is consistent with the basic mechanism discussed in Section~\ref{sec:reasoning} that required two self-attention layers to function.

        It is worth noticing that adding more unlabeled samples into the support set makes our model more difficult to train and it gets stuck producing CNNs with essentially random outputs.
        Our solution was to introduce unlabeled samples incrementally during training.
        This was implemented by masking out some unlabeled samples in the beginning of the training and then gradually reducing the masking probability over time\footnote{we linearly interpolated the probability of ignoring an unlabeled sample from $0.7$ to $0.0$ in half the training steps}.

        \begin{figure*}[h]
            \centering
            Layer 1 \hspace{6em} Layer 2 \hspace{6em} Layer 3 \hspace{6em} Layer 4 \hspace{6em} All layers
            \includegraphics[width=\textwidth]{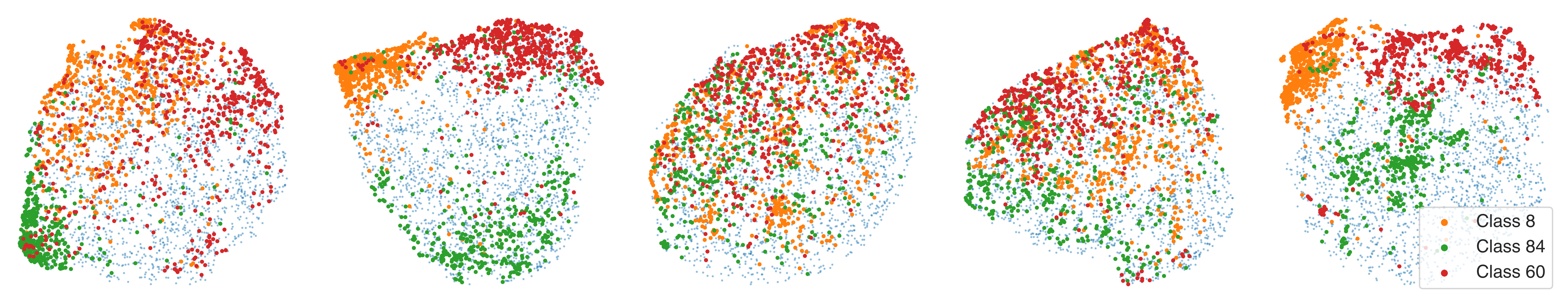}
            \smallskip
            \caption{
                \label{fig:umap}
                UMAP embedding of weights for each convolutional layer of a 6-channel CNN generated by \hyper{} for $1\,242$ different episodes from \tiered{}. Each point corresponds to 2d embedding of the combined weights for a given layer (or concatenated for all layers) generated for a given episode. We color some of the point according to the classes contained in the episodes. For highlighted classes, the generated weights appear to be correlated between episodes where these classes are present. We selected classes specifically to demonstrate this correlation. For most of the other classes, this correlation was minor. Note that since there are 5 classes in each episode, the coloring for some of the episodes might be ambiguous. See Appendix~\ref{app:umap} for more classes and samples for each class. 
            }
        \end{figure*} 

    \smallskip
    \subsection{Generating Additional Model Layers}
        \label{sec:more}

        We demonstrated that \hyper{} model can outperform \mamlpp{} on common few-shot learning datasets by generating just the last logits layer of the CNN model.
        But is it advantageous to be generating additional CNN layers (ultimately fully utilizing the capability of the \hyper{} model)?

        We explored this question by comparing the performance of models, in which all, or only some of the convolutional layers were generated, while others were learned (typically all or few first convolutional layers of the CNN).
        We observed a significant performance improvement for models that generated all convolutional layers in addition to the CNN logits layer, but only for CNN models below a particular size.
        For \omni{} dataset, we saw that both training and test accuracies for a 4-channel and a 6-channel CNNs increased with the number of generated layers (see Fig.~\ref{fig:tiny} and Table~\ref{tab:tiny} in Appendix) and using more complex Transformer models with $2$ or more encoder layers improved both training and test accuracies of fully-generated CNN models of this size (see Appendix~\ref{app:ablation}).
        However, as the size of the model increased and reached $8$ channels, generating the last logits layer alone proved to be sufficient for getting the best results on \omni{} and \tiered{}.
        By separately training an ``oracle'' CNN model using all available data for a random set of $\numclasses$ classes, we observed the gap between the training accuracy of the generated model and the {\em oracle} model (see Fig.~\ref{fig:tiny}), indicating that the Transformer does not fully capture the dependence of the optimal CNN model weights on the support set samples.
        A hypothetical weight generator reaching maximum training accuracy could, in principle, memorize all training images being able to associate them with corresponding classes and then generate an optimal CNN model for a particular set of classes in the episode matching ``oracle'' model performance.
        
        We visualized the distribution of the weights generated by \hyper{} for different episodes by using UMAP~\citep{umap} embeddings of the generated weights for a 6-channel CNN model (see Fig.~\ref{fig:umap}).
        We highlighted some of the classes present in the evaluation set and while the general structure may be hard to interpret, the distribution of the highlighted classes is somewhat clustered indicating the importance of semantic information for generated CNN weights.
        More details can be found in Appendix~\ref{app:umap}.

        The positive effect of generating convolutional layers can also be observed in shallow models with large convolutional kernels and large strides where the model performance can be much more sensitive to a proper choice of model weights.
        For example, in a 16-channel model with two convolutional kernels of size $9$ and the stride of $4$, the overall test accuracy for a model generating only the final convolutional layer was about $1\%$ lower than the accuracies of the models generating at least one additional convolutional filter.
        We also speculate that as the complexity of the task increases, generating some or all intermediate network layers should become more important for achieving optimal performance.
        Verifying this hypothesis and understanding the ``boundary'' in the model space between two regimes where a static backbone is sufficient or not will be the subject of our future work.

\smallskip
\section{Conclusions}

    In this work, we proposed {\em a HyperTransformer} (\hyper{}), a novel Transformer-based model that generates all weights of a CNN model directly from a few-shot support set.
    This approach allowed us to use a high-capacity model for encoding task-dependent variations in the weights of a smaller model.
    We demonstrated that generating the last logits layer alone, the Transformer-based weight generator beats or matches performance of multiple traditional learning methods on several few-shot benchmarks and surpasses \mamlpp{} and \rfs{} performance on smaller models.
    More importantly, we showed that \hyper{} can be straightforwardly extended to handle more complex problems like semi-supervised tasks with unlabeled samples present in the support set.
    Our experiments demonstrated a considerable few-shot performance improvement in the presence of unlabeled data.
    Finally, we explored the impact of the Transformer-encoded model diversity in CNN models of different sizes.
    We used \hyper{} to generate some or all convolutional kernels and biases and showed that for sufficiently small models, adjusting all model parameters further improves their few-shot learning performance. 

\section*{Acknowledgements}

We would like to thank Azalia Mirhoseini, David Ha, Bill Mark, Luke Metz, Raviteja Vemulapalli, Philip Mansfield, and Nolan Miller for insightful discussions.

\bibliography{paper}

\begin{thebibliography}{45}
\providecommand{\natexlab}[1]{#1}
\providecommand{\url}[1]{\texttt{#1}}
\expandafter\ifx\csname urlstyle\endcsname\relax
  \providecommand{\doi}[1]{doi: #1}\else
  \providecommand{\doi}{doi: \begingroup \urlstyle{rm}\Url}\fi

\bibitem[Allen et~al.(2019)Allen, Shelhamer, Shin, and
  Tenenbaum]{allen2019infinite}
Allen, K.~R., Shelhamer, E., Shin, H., and Tenenbaum, J.~B.
\newblock Infinite mixture prototypes for few-shot learning.
\newblock In Chaudhuri, K. and Salakhutdinov, R. (eds.), \emph{Proceedings of
  the 36th International Conference on Machine Learning, {ICML} 2019, 9-15 June
  2019, Long Beach, California, {USA}}, volume~97 of \emph{Proceedings of
  Machine Learning Research}, pp.\  232--241. {PMLR}, 2019.

\bibitem[Antoniou et~al.(2019)Antoniou, Edwards, and Storkey]{antoniou2019maml}
Antoniou, A., Edwards, H., and Storkey, A.~J.
\newblock How to train your {MAML}.
\newblock In \emph{7th International Conference on Learning Representations,
  {ICLR} 2019, New Orleans, LA, USA, May 6-9, 2019}. OpenReview.net, 2019.

\bibitem[Carion et~al.(2020)Carion, Massa, Synnaeve, Usunier, Kirillov, and
  Zagoruyko]{carion2020detection}
Carion, N., Massa, F., Synnaeve, G., Usunier, N., Kirillov, A., and Zagoruyko,
  S.
\newblock End-to-end object detection with transformers.
\newblock In Vedaldi, A., Bischof, H., Brox, T., and Frahm, J. (eds.),
  \emph{Computer Vision - {ECCV} 2020 - 16th European Conference, Glasgow, UK,
  August 23-28, 2020, Proceedings, Part {I}}, volume 12346 of \emph{Lecture
  Notes in Computer Science}, pp.\  213--229. Springer, 2020.

\bibitem[Chen et~al.(2021)Chen, Wang, Guo, Xu, Deng, Liu, Ma, Xu, Xu, and
  Gao]{chen2021}
Chen, H., Wang, Y., Guo, T., Xu, C., Deng, Y., Liu, Z., Ma, S., Xu, C., Xu, C.,
  and Gao, W.
\newblock Pre-trained image processing transformer.
\newblock In \emph{{IEEE} Conference on Computer Vision and Pattern
  Recognition, {CVPR} 2021, virtual, June 19-25, 2021}, pp.\  12299--12310.
  Computer Vision Foundation / {IEEE}, 2021.

\bibitem[Dosovitskiy et~al.(2021)Dosovitskiy, Beyer, Kolesnikov, Weissenborn,
  Zhai, Unterthiner, Dehghani, Minderer, Heigold, Gelly, Uszkoreit, and
  Houlsby]{dosovitskiy2021image}
Dosovitskiy, A., Beyer, L., Kolesnikov, A., Weissenborn, D., Zhai, X.,
  Unterthiner, T., Dehghani, M., Minderer, M., Heigold, G., Gelly, S.,
  Uszkoreit, J., and Houlsby, N.
\newblock An image is worth 16x16 words: Transformers for image recognition at
  scale.
\newblock In \emph{9th International Conference on Learning Representations,
  {ICLR} 2021, Virtual Event, Austria, May 3-7, 2021}, 2021.

\bibitem[Fei et~al.(2021)Fei, Lu, Xiang, and Huang]{fei2021melr}
Fei, N., Lu, Z., Xiang, T., and Huang, S.
\newblock {MELR:} meta-learning via modeling episode-level relationships for
  few-shot learning.
\newblock In \emph{9th International Conference on Learning Representations,
  {ICLR} 2021, Virtual Event, Austria, May 3-7, 2021}. OpenReview.net, 2021.

\bibitem[Finn et~al.(2017)Finn, Abbeel, and Levine]{finn2017maml}
Finn, C., Abbeel, P., and Levine, S.
\newblock Model-agnostic meta-learning for fast adaptation of deep networks.
\newblock In Precup, D. and Teh, Y.~W. (eds.), \emph{Proceedings of the 34th
  International Conference on Machine Learning}, volume~70 of \emph{Proceedings
  of Machine Learning Research}, pp.\  1126--1135. PMLR, 06--11 Aug 2017.

\bibitem[Gidaris \& Komodakis(2018)Gidaris and Komodakis]{gidaris2018dynamic}
Gidaris, S. and Komodakis, N.
\newblock Dynamic few-shot visual learning without forgetting.
\newblock In \emph{2018 {IEEE} Conference on Computer Vision and Pattern
  Recognition, {CVPR} 2018, Salt Lake City, UT, USA, June 18-22, 2018}, pp.\
  4367--4375. {IEEE} Computer Society, 2018.
\newblock \doi{10.1109/CVPR.2018.00459}.

\bibitem[Guo \& Cheung(2020)Guo and Cheung]{guo2020attentive}
Guo, Y. and Cheung, N.
\newblock Attentive weights generation for few shot learning via information
  maximization.
\newblock In \emph{2020 {IEEE/CVF} Conference on Computer Vision and Pattern
  Recognition, {CVPR} 2020, Seattle, WA, USA, June 13-19, 2020}, pp.\
  13496--13505. Computer Vision Foundation / {IEEE}, 2020.

\bibitem[Ha et~al.(2017)Ha, Dai, and Le]{ha2016hypernetworks}
Ha, D., Dai, A.~M., and Le, Q.~V.
\newblock {HyperNetworks}.
\newblock In \emph{5th International Conference on Learning Representations,
  {ICLR} 2017, Toulon, France, April 24-26, 2017, Conference Track
  Proceedings}, 2017.

\bibitem[Hao et~al.(2019)Hao, He, Cheng, Wang, Cao, and Tao]{hao2019collect}
Hao, F., He, F., Cheng, J., Wang, L., Cao, J., and Tao, D.
\newblock Collect and select: Semantic alignment metric learning for few-shot
  learning.
\newblock In \emph{2019 {IEEE/CVF} International Conference on Computer Vision,
  {ICCV} 2019, Seoul, Korea (South), October 27 - November 2, 2019}, pp.\
  8459--8468. {IEEE}, 2019.
\newblock \doi{10.1109/ICCV.2019.00855}.

\bibitem[Jamal \& Qi(2019)Jamal and Qi]{jamal2019ml}
Jamal, M.~A. and Qi, G.
\newblock Task agnostic meta-learning for few-shot learning.
\newblock In \emph{{IEEE} Conference on Computer Vision and Pattern
  Recognition, {CVPR} 2019, Long Beach, CA, USA, June 16-20, 2019}, pp.\
  11719--11727. Computer Vision Foundation / {IEEE}, 2019.
\newblock \doi{10.1109/CVPR.2019.01199}.

\bibitem[Koch et~al.(2015)Koch, Zemel, Salakhutdinov, et~al.]{koch2015siamese}
Koch, G., Zemel, R., Salakhutdinov, R., et~al.
\newblock Siamese neural networks for one-shot image recognition.
\newblock In \emph{ICML deep learning workshop}, volume~2. Lille, 2015.

\bibitem[Li et~al.(2019{\natexlab{a}})Li, Luo, Xiang, Huang, and
  Wang]{li2019global}
Li, A., Luo, T., Xiang, T., Huang, W., and Wang, L.
\newblock Few-shot learning with global class representations.
\newblock In \emph{2019 {IEEE/CVF} International Conference on Computer Vision,
  {ICCV} 2019, Seoul, Korea (South), October 27 - November 2, 2019}, pp.\
  9714--9723. {IEEE}, 2019{\natexlab{a}}.
\newblock \doi{10.1109/ICCV.2019.00981}.

\bibitem[Li et~al.(2019{\natexlab{b}})Li, Dong, Mei, Ma, Huang, and
  Hu]{li2019lgm}
Li, H., Dong, W., Mei, X., Ma, C., Huang, F., and Hu, B.
\newblock Lgm-net: Learning to generate matching networks for few-shot
  learning.
\newblock In Chaudhuri, K. and Salakhutdinov, R. (eds.), \emph{Proceedings of
  the 36th International Conference on Machine Learning, {ICML} 2019, 9-15 June
  2019, Long Beach, California, {USA}}, volume~97 of \emph{Proceedings of
  Machine Learning Research}, pp.\  3825--3834. {PMLR}, 2019{\natexlab{b}}.

\bibitem[Liu et~al.(2021)Liu, Hamilton, Long, Jiang, and
  Larochelle]{liu2021universal}
Liu, L., Hamilton, W.~L., Long, G., Jiang, J., and Larochelle, H.
\newblock A universal representation transformer layer for few-shot image
  classification.
\newblock In \emph{9th International Conference on Learning Representations,
  {ICLR} 2021, Virtual Event, Austria, May 3-7, 2021}. OpenReview.net, 2021.

\bibitem[Liu et~al.(2019)Liu, Lee, Park, Kim, Yang, Hwang, and
  Yang]{liu2019transductive}
Liu, Y., Lee, J., Park, M., Kim, S., Yang, E., Hwang, S.~J., and Yang, Y.
\newblock Learning to propagate labels: Transductive propagation network for
  few-shot learning.
\newblock In \emph{7th International Conference on Learning Representations,
  {ICLR} 2019, New Orleans, LA, USA, May 6-9, 2019}, 2019.

\bibitem[Ma(2019)]{lgmnet-github}
Ma, C.
\newblock {LGM-Net}.
\newblock \url{https://github.com/likesiwell/LGM-Net}, 2019.

\bibitem[Mahabadi et~al.(2021)Mahabadi, Ruder, Dehghani, and
  Henderson]{mahabadi2021finetuning}
Mahabadi, R.~K., Ruder, S., Dehghani, M., and Henderson, J.
\newblock Parameter-efficient multi-task fine-tuning for transformers via
  shared hypernetworks.
\newblock In Zong, C., Xia, F., Li, W., and Navigli, R. (eds.),
  \emph{Proceedings of the 59th Annual Meeting of the Association for
  Computational Linguistics and the 11th International Joint Conference on
  Natural Language Processing, {ACL/IJCNLP} 2021, (Volume 1: Long Papers),
  Virtual Event, August 1-6, 2021}, pp.\  565--576. Association for
  Computational Linguistics, 2021.
\newblock \doi{10.18653/v1/2021.acl-long.47}.

\bibitem[McInnes et~al.(2018)McInnes, Healy, and Melville]{umap}
McInnes, L., Healy, J., and Melville, J.
\newblock Umap: Uniform manifold approximation and projection for dimension
  reduction.
\newblock \emph{arXiv preprint arXiv:1802.03426}, 2018.

\bibitem[Nichol et~al.(2018)Nichol, Achiam, and Schulman]{nichol2018meta}
Nichol, A., Achiam, J., and Schulman, J.
\newblock On first-order meta-learning algorithms.
\newblock \emph{CoRR}, abs/1803.02999, 2018.

\bibitem[Oreshkin et~al.(2018)Oreshkin, L{\'{o}}pez, and
  Lacoste]{oreshkin2018tadam}
Oreshkin, B.~N., L{\'{o}}pez, P.~R., and Lacoste, A.
\newblock {TADAM:} task dependent adaptive metric for improved few-shot
  learning.
\newblock In Bengio, S., Wallach, H.~M., Larochelle, H., Grauman, K.,
  Cesa{-}Bianchi, N., and Garnett, R. (eds.), \emph{Advances in Neural
  Information Processing Systems 31: Annual Conference on Neural Information
  Processing Systems 2018, NeurIPS 2018, December 3-8, 2018, Montr{\'{e}}al,
  Canada}, pp.\  719--729, 2018.

\bibitem[Peng et~al.(2019)Peng, Li, Zhang, Li, Qi, and Tang]{peng2019fewshot}
Peng, Z., Li, Z., Zhang, J., Li, Y., Qi, G., and Tang, J.
\newblock Few-shot image recognition with knowledge transfer.
\newblock In \emph{2019 {IEEE/CVF} International Conference on Computer Vision,
  {ICCV} 2019, Seoul, Korea (South), October 27 - November 2, 2019}, pp.\
  441--449. {IEEE}, 2019.
\newblock \doi{10.1109/ICCV.2019.00053}.

\bibitem[Pilault et~al.(2021)Pilault, Elhattami, and
  Pal]{pilault2021mtlearning}
Pilault, J., Elhattami, A., and Pal, C.~J.
\newblock Conditionally adaptive multi-task learning: Improving transfer
  learning in {NLP} using fewer parameters {\&} less data.
\newblock In \emph{9th International Conference on Learning Representations,
  {ICLR} 2021, Virtual Event, Austria, May 3-7, 2021}. OpenReview.net, 2021.

\bibitem[Qi et~al.(2018)Qi, Brown, and Lowe]{qi2018imprinted}
Qi, H., Brown, M., and Lowe, D.~G.
\newblock Low-shot learning with imprinted weights.
\newblock In \emph{2018 {IEEE} Conference on Computer Vision and Pattern
  Recognition, {CVPR} 2018, Salt Lake City, UT, USA, June 18-22, 2018}, pp.\
  5822--5830. Computer Vision Foundation / {IEEE} Computer Society, 2018.
\newblock \doi{10.1109/CVPR.2018.00610}.

\bibitem[Qiao et~al.(2018)Qiao, Liu, Shen, and Yuille]{qiao2018ppa}
Qiao, S., Liu, C., Shen, W., and Yuille, A.~L.
\newblock Few-shot image recognition by predicting parameters from activations.
\newblock In \emph{2018 {IEEE} Conference on Computer Vision and Pattern
  Recognition, {CVPR} 2018, Salt Lake City, UT, USA, June 18-22, 2018}, pp.\
  7229--7238. Computer Vision Foundation / {IEEE} Computer Society, 2018.
\newblock \doi{10.1109/CVPR.2018.00755}.

\bibitem[Ratzlaff \& Li(2019)Ratzlaff and Li]{ratzlaff2019hypergan}
Ratzlaff, N. and Li, F.
\newblock Hypergan: {A} generative model for diverse, performant neural
  networks.
\newblock In Chaudhuri, K. and Salakhutdinov, R. (eds.), \emph{Proceedings of
  the 36th International Conference on Machine Learning, {ICML} 2019, 9-15 June
  2019, Long Beach, California, {USA}}, volume~97 of \emph{Proceedings of
  Machine Learning Research}, pp.\  5361--5369. {PMLR}, 2019.

\bibitem[Ravi \& Larochelle(2017)Ravi and Larochelle]{ravi2017optimization}
Ravi, S. and Larochelle, H.
\newblock Optimization as a model for few-shot learning.
\newblock In \emph{5th International Conference on Learning Representations,
  {ICLR} 2017, Toulon, France, April 24-26, 2017, Conference Track
  Proceedings}. OpenReview.net, 2017.

\bibitem[Ren et~al.(2018)Ren, Triantafillou, Ravi, Snell, Swersky, Tenenbaum,
  Larochelle, and Zemel]{ren2018semi}
Ren, M., Triantafillou, E., Ravi, S., Snell, J., Swersky, K., Tenenbaum, J.~B.,
  Larochelle, H., and Zemel, R.~S.
\newblock Meta-learning for semi-supervised few-shot classification.
\newblock In \emph{6th International Conference on Learning Representations,
  {ICLR} 2018, Vancouver, BC, Canada, April 30 - May 3, 2018, Conference Track
  Proceedings}, 2018.

\bibitem[Requeima et~al.(2019)Requeima, Gordon, Bronskill, Nowozin, and
  Turner]{requeima2019conditional}
Requeima, J., Gordon, J., Bronskill, J., Nowozin, S., and Turner, R.~E.
\newblock Fast and flexible multi-task classification using conditional neural
  adaptive processes.
\newblock In Wallach, H.~M., Larochelle, H., Beygelzimer, A.,
  d'Alch{\'{e}}{-}Buc, F., Fox, E.~B., and Garnett, R. (eds.), \emph{Advances
  in Neural Information Processing Systems 32: Annual Conference on Neural
  Information Processing Systems 2019, NeurIPS 2019, December 8-14, 2019,
  Vancouver, BC, Canada}, pp.\  7957--7968, 2019.

\bibitem[Rusu et~al.(2019)Rusu, Rao, Sygnowski, Vinyals, Pascanu, Osindero, and
  Hadsell]{rusu2019metalearning}
Rusu, A.~A., Rao, D., Sygnowski, J., Vinyals, O., Pascanu, R., Osindero, S.,
  and Hadsell, R.
\newblock Meta-learning with latent embedding optimization.
\newblock In \emph{7th International Conference on Learning Representations,
  {ICLR} 2019, New Orleans, LA, USA, May 6-9, 2019}, 2019.

\bibitem[Snell et~al.(2017)Snell, Swersky, and Zemel]{snell2017prototypical}
Snell, J., Swersky, K., and Zemel, R.~S.
\newblock Prototypical networks for few-shot learning.
\newblock In Guyon, I., von Luxburg, U., Bengio, S., Wallach, H.~M., Fergus,
  R., Vishwanathan, S. V.~N., and Garnett, R. (eds.), \emph{Advances in Neural
  Information Processing Systems 30: Annual Conference on Neural Information
  Processing Systems 2017, December 4-9, 2017, Long Beach, CA, {USA}}, pp.\
  4077--4087, 2017.

\bibitem[Sung et~al.(2018)Sung, Yang, Zhang, Xiang, Torr, and
  Hospedales]{sung2018relation}
Sung, F., Yang, Y., Zhang, L., Xiang, T., Torr, P. H.~S., and Hospedales, T.~M.
\newblock Learning to compare: Relation network for few-shot learning.
\newblock In \emph{2018 {IEEE} Conference on Computer Vision and Pattern
  Recognition, {CVPR} 2018, Salt Lake City, UT, USA, June 18-22, 2018}, pp.\
  1199--1208. {IEEE} Computer Society, 2018.
\newblock \doi{10.1109/CVPR.2018.00131}.

\bibitem[Tay et~al.(2021)Tay, Zhao, Bahri, Metzler, and Juan]{tay2021hypergrid}
Tay, Y., Zhao, Z., Bahri, D., Metzler, D., and Juan, D.
\newblock Hypergrid transformers: Towards {A} single model for multiple tasks.
\newblock In \emph{9th International Conference on Learning Representations,
  {ICLR} 2021, Virtual Event, Austria, May 3-7, 2021}. OpenReview.net, 2021.

\bibitem[Tian et~al.(2020)Tian, Wang, Krishnan, Tenenbaum, and
  Isola]{tian2020rethinking}
Tian, Y., Wang, Y., Krishnan, D., Tenenbaum, J.~B., and Isola, P.
\newblock Rethinking few-shot image classification: {A} good embedding is all
  you need?
\newblock In Vedaldi, A., Bischof, H., Brox, T., and Frahm, J. (eds.),
  \emph{Computer Vision - {ECCV} 2020 - 16th European Conference, Glasgow, UK,
  August 23-28, 2020, Proceedings, Part {XIV}}, volume 12359 of \emph{Lecture
  Notes in Computer Science}, pp.\  266--282. Springer, 2020.
\newblock \doi{10.1007/978-3-030-58568-6\_16}.

\bibitem[Touvron et~al.(2021)Touvron, Cord, Douze, Massa, Sablayrolles, and
  J{\'{e}}gou]{touvron2021deit}
Touvron, H., Cord, M., Douze, M., Massa, F., Sablayrolles, A., and J{\'{e}}gou,
  H.
\newblock Training data-efficient image transformers {\&} distillation through
  attention.
\newblock In Meila, M. and Zhang, T. (eds.), \emph{Proceedings of the 38th
  International Conference on Machine Learning, {ICML} 2021, 18-24 July 2021,
  Virtual Event}, volume 139 of \emph{Proceedings of Machine Learning
  Research}, pp.\  10347--10357. {PMLR}, 2021.

\bibitem[Vaswani et~al.(2017)Vaswani, Shazeer, Parmar, Uszkoreit, Jones, Gomez,
  Kaiser, and Polosukhin]{vaswani2017attention}
Vaswani, A., Shazeer, N., Parmar, N., Uszkoreit, J., Jones, L., Gomez, A.~N.,
  Kaiser, L., and Polosukhin, I.
\newblock Attention is all you need.
\newblock In Guyon, I., von Luxburg, U., Bengio, S., Wallach, H.~M., Fergus,
  R., Vishwanathan, S. V.~N., and Garnett, R. (eds.), \emph{Advances in Neural
  Information Processing Systems 30: Annual Conference on Neural Information
  Processing Systems 2017, December 4-9, 2017, Long Beach, CA, {USA}}, pp.\
  5998--6008, 2017.

\bibitem[Vinyals et~al.(2016)Vinyals, Blundell, Lillicrap, Kavukcuoglu, and
  Wierstra]{vinyals2016matching}
Vinyals, O., Blundell, C., Lillicrap, T., Kavukcuoglu, K., and Wierstra, D.
\newblock Matching networks for one shot learning.
\newblock In Lee, D.~D., Sugiyama, M., von Luxburg, U., Guyon, I., and Garnett,
  R. (eds.), \emph{Advances in Neural Information Processing Systems 29: Annual
  Conference on Neural Information Processing Systems 2016, December 5-10,
  2016, Barcelona, Spain}, pp.\  3630--3638, 2016.

\bibitem[Wu et~al.(2019)Wu, Li, Guo, and Jia]{wu2019parn}
Wu, Z., Li, Y., Guo, L., and Jia, K.
\newblock {PARN:} position-aware relation networks for few-shot learning.
\newblock In \emph{2019 {IEEE/CVF} International Conference on Computer Vision,
  {ICCV} 2019, Seoul, Korea (South), October 27 - November 2, 2019}, pp.\
  6658--6666. {IEEE}, 2019.
\newblock \doi{10.1109/ICCV.2019.00676}.

\bibitem[Yang et~al.(2020)Yang, Yang, Fu, Lu, and Guo]{yang2020super}
Yang, F., Yang, H., Fu, J., Lu, H., and Guo, B.
\newblock Learning texture transformer network for image super-resolution.
\newblock In \emph{2020 {IEEE/CVF} Conference on Computer Vision and Pattern
  Recognition, {CVPR} 2020, Seattle, WA, USA, June 13-19, 2020}, pp.\
  5790--5799. Computer Vision Foundation / {IEEE}, 2020.
\newblock \doi{10.1109/CVPR42600.2020.00583}.

\bibitem[Ye et~al.(2020)Ye, Hu, Zhan, and Sha]{ye2020adaptation}
Ye, H., Hu, H., Zhan, D., and Sha, F.
\newblock Few-shot learning via embedding adaptation with set-to-set functions.
\newblock In \emph{2020 {IEEE/CVF} Conference on Computer Vision and Pattern
  Recognition, {CVPR} 2020, Seattle, WA, USA, June 13-19, 2020}, pp.\
  8805--8814. {IEEE}, 2020.
\newblock \doi{10.1109/CVPR42600.2020.00883}.

\bibitem[Ye et~al.(2019)Ye, Rochan, Liu, and Wang]{ye2019segmentation}
Ye, L., Rochan, M., Liu, Z., and Wang, Y.
\newblock Cross-modal self-attention network for referring image segmentation.
\newblock In \emph{{IEEE} Conference on Computer Vision and Pattern
  Recognition, {CVPR} 2019, Long Beach, CA, USA, June 16-20, 2019}, pp.\
  10502--10511. Computer Vision Foundation / {IEEE}, 2019.
\newblock \doi{10.1109/CVPR.2019.01075}.

\bibitem[Ye \& Ren(2021)Ye and Ren]{ye2021adapters}
Ye, Q. and Ren, X.
\newblock Learning to generate task-specific adapters from task description.
\newblock In Zong, C., Xia, F., Li, W., and Navigli, R. (eds.),
  \emph{Proceedings of the 59th Annual Meeting of the Association for
  Computational Linguistics and the 11th International Joint Conference on
  Natural Language Processing, {ACL/IJCNLP} 2021, (Volume 2: Short Papers),
  Virtual Event, August 1-6, 2021}, pp.\  646--653. Association for
  Computational Linguistics, 2021.
\newblock \doi{10.18653/v1/2021.acl-short.82}.

\bibitem[Zhao et~al.(2020)Zhao, von Oswald, Kobayashi, Sacramento, and
  Grewe]{zhao2020meta}
Zhao, D., von Oswald, J., Kobayashi, S., Sacramento, J., and Grewe, B.~F.
\newblock Meta-learning via hypernetworks.
\newblock 2020.

\bibitem[Zhu et~al.(2021)Zhu, Su, Lu, Li, Wang, and Dai]{zhu2020detr}
Zhu, X., Su, W., Lu, L., Li, B., Wang, X., and Dai, J.
\newblock Deformable {DETR:} deformable transformers for end-to-end object
  detection.
\newblock In \emph{9th International Conference on Learning Representations,
  {ICLR} 2021, Virtual Event, Austria, May 3-7, 2021}. OpenReview.net, 2021.

\end{thebibliography}
\bibliographystyle{icml2022}

\appendix
\onecolumn

\newpage

\section{Example of a Self-Attention Mechanism for Supervised Learning}
\label{app:unlabeled}

    Self-attention in its rudimentary form can implement a cosine-similarity-based sample weighting, which can also be viewed as a simple 1-step \maml-like learning algorithm.
    This can be seen by considering a simple classification model
    \begin{gather*}
        f(x;\theta) = s(\mW \ve(x;\phi) + \vb)
    \end{gather*}
    with $\theta=(\mW,\vb,\phi)$, where $\ve(x;\phi)$ is the embedding and $s(\cdot)$ is a softmax function.
	\maml\ algorithm identifies such initial weights $\theta_0$ that given any task $T$ just a few gradient descent steps with respect to the loss $\mathcal{L}_{T}$ starting at $\theta_0$ bring the model towards a task-specific local optimum of $\mathcal{L}_T$.

    Notice that if any label assignment in the training tasks is equally likely, it is natural for $f(x;\theta_0)$ to not prefer any particular label over the others.
	Guided by this, let us choose $\mW_0$ and $\vb_0$ that are {\em label-independent}.
    Substituting $\theta=\theta_0+\delta\theta$ into $f(x;\theta)$, we obtain
    \begin{gather*}
        f_\ell(x;\theta) =
        f_\ell(x;\theta_0)
        + s_\ell'(\cdot) (\delta \mW_\ell \ve(x;\phi_0) + \delta \vb_\ell +
        \mW_{0,\ell} \ve'(x;\phi_0) \delta \phi)
        + O(\delta \theta^2),
    \end{gather*}
    where $\ell$ is the label index and $\delta \theta=(\delta \mW,\delta \vb,\delta \phi)$.
	We see that the lowest-order label-dependent correction to $f_\ell(x;\theta_0)$ is given simply by $s_\ell'(\cdot) (\delta \mW_\ell \ve(x;\phi_0) + \delta \vb_\ell)$.
	In other words, in the lowest-order, the model only adjusts the final logits layer to adapt the pretrained embedding $\ve(x;\phi_0)$ to a new task.
	It is then easy to calculate that for a simple softmax cross-entropy loss, a single step of the gradient descent results in the following logits weight and bias updates:
    \begin{align}
        \label{eq:app-w}
		\delta W_{ij} = \frac{\gamma}{n} \sum_{m=1}^{n} \left(y_i^{(m)} - \frac{1}{|C|}\right) e_j(x^{(m)};\phi_0), \qquad
		\delta b_{i} = \frac{\gamma}{n} \sum_{m=1}^{n} \left(y_i^{(m)} - \frac{1}{|C|}\right).
    \end{align}
	Here $\gamma$ is the learning rate, $n$ is the total number of support-set samples, $|C|$ is the number of classes and $\vy^{(m)}$ is the one-hot label corresponding to $x^{(m)}$.
	A closely-related idea of extending the logits layer with a vector proportional to an average of the novel class sample embeddings is used in the ``imprinted weights'' approach \cite{qi2018imprinted} allowing to add novel classes into pre-trained models.
	
	Now consider a self-attention module generating the last logits layer and acting on a sequence of processed input samples\footnote{here we use only activation features $h_{\vphi_l}(e_i)$ of the sample embedding vectors $e_i$} $\mathcal{I}^L=\{(\, \xi(c_i) \, , \, h_{\vphi_l}(e_i)\, )\}_{i=1,\dots,n}$ and weight placeholders $\mathcal{W}^L := \{ (\,\mu(k)\,,\,0\,)\}_{k=1,\dots,|C|}$, where $|C|$ is the number of classes and also the number of weight slices of $\mW$ if each slice corresponds to an output layer channel.
	The output of a simple self-attention module for the weight slice with index $i$ is then given by:
	\begin{gather}
	    \label{eq:app-gen-slice}
	    Z^{-1} \sum_{m=1}^{n} e^{\mQ(\mathcal{W}^L_i) \cdot \mK_I(\mathcal{I}^L_m)} \mV_I(\mathcal{I}^L_m) +
	    Z^{-1} \sum_{m=1}^{N_w} e^{\mQ(\mathcal{W}^L_i) \cdot \mK_W(\mathcal{W}^L_m)} \mV_W(\mathcal{W}^L_m),
	\end{gather}
	where $Z := \sum_{m=1}^{n} e^{\mQ(\mathcal{W}^L_i) \cdot \mK_I(\mathcal{I}^L_m)} + \sum_{m=1}^{N_w} e^{\mQ(\mathcal{W}^L_i) \cdot \mK_W(\mathcal{W}^L_m)}$.
	It is easy to see that with a proper choice of query and key matrices attending only to prepended $\xi$ and $\mu$ tokens, the second term in equation~\ref{eq:app-gen-slice} can be made negligible, while $\mQ(\mathcal{W}^L_i) \cdot \mK_I(\mathcal{I}^L_m)$ can make  the softmax function only attend to those components of $\mV_I$ that correspond to samples with label $i$.
	Choosing $\mV_I(\mathcal{I}^L_m)$ to be proportional to $\ve_m$, we can then recover the first term in $\delta \mW$ in \eqref{eq:app-w}.
	The second term in $\delta \mW$ can be produced, for example, with the help of a second head that generates identical attention weights for all samples, thus summing up their embeddings.

\section{Analytical Expression for Generated Weights}
\label{app:wg}

    Supplied with a distribution $p(t)$ of training tasks $t\in \trtasks$, we learn the weight generator $a_\vphi$ by optimizing the following objective:
    \begin{gather*}
        \argmin_\vphi \E_{t\sim p(t)} \mathcal{L}(a_\vphi(\task(t)),t).
    \end{gather*}
    If the family of functions $a_\vphi(\task)$ is sufficiently rich, we can instead try solving an optimization problem of the form $\theta_*(\task(t)) := \argmin_{\theta} \mathcal{L}(\theta,t)$ with $\theta_* \in C^{2}$.
    
    Generally, when $t\mapsto \tau$ is not one-to-one, it is impossible to minimize $\mathcal{L}(\theta_*(\task(t)),t)$ for each $t$ because $\mathcal{L}(\cdot,t_1)$ and $\mathcal{L}(\cdot,t_2)$ will typically differ even if $\tau(t_1)=\tau(t_2)$.
    However, if this mapping is one-to-one, and we can re-parameterize the space of tasks $\tasks$ choosing $t$ to have the same dimension as $\tau$, the condition that $\mathcal{L}(\theta_*(\task(t+\delta t)),t+\delta t)$ is an extremum for any infinitesimal $\delta t$ can be rewritten as:
    \begin{gather*}
        \frac{\partial \mathcal{L}}{\partial \theta}(\theta_*(\task(t+\delta t)),t+\delta t)
        = \frac{\partial^2 \mathcal{L}}{\partial \theta^2} \frac{\partial \theta_*}{\partial \task} \frac{\partial \task}{\partial t} \delta t + \frac{\partial^2 \mathcal{L}}{\partial \theta \, \partial t} \delta t + O(\delta t^2)
        = 0,
    \end{gather*}
    which in turn means that:
    \begin{gather*}
        \frac{\partial^2 \mathcal{L}}{\partial \theta^2} \frac{\partial \theta_*}{\partial \tau} \frac{\partial \tau}{\partial t} = -\frac{\partial^2 \mathcal{L}}{\partial \theta \, \partial t}.
    \end{gather*}
    If the Hessian of $\mathcal{L}$ and $\partial \tau / \partial t$ are non-singular, we can solve this equation for $\partial \theta_* / \partial \tau$:
    \begin{gather}
        \label{eq:der}
        \frac{\partial \theta_*}{\partial \tau} = -\left( \frac{\partial^2 \mathcal{L}}{\partial \theta^2} \right)^{-1}
        \frac{\partial^2 \mathcal{L}}{\partial \theta \, \partial t}
        \left( \frac{\partial \tau}{\partial t} \right)^{-1}.
    \end{gather}
    
    Now assume that we know a solution of $\argmin_{\theta_*} \mathcal{L}(\theta_*,t)$ for some particular task $t=t_0$.
    The derivative \ref{eq:der} can then be used to ``track'' this local minimum at $t_0$ to any other task $t_1$ in a sufficiently small vicinity of $t_0$, where $\mathcal{L}$ remains convex and the Hessian of $\mathcal{L}$ is not singular.
    Choosing a path $\hat{t}:[0,1] \to \tasks$ with $\hat{t}(0)=t_0$ and $\hat{t}(1)=t_1$, we need to integrate $\partial \theta_* / \partial \tau$ along $\tau(\hat{t}(\gamma))$ with $\gamma$ changing from $0$ to $1$, which is equivalent to integrating the following ordinary differential equation:
    \begin{gather*}
        \frac{d \theta_*}{d\gamma} = 
        -\left( \frac{\partial^2 \mathcal{L}}{\partial \theta^2} \right)^{-1}
        \frac{\partial^2 \mathcal{L}}{\partial \theta \, \partial t}
        \frac{d \hat{t}}{d\gamma},
    \end{gather*}
    where $\theta_*(\gamma) = \theta_*(\tau(\hat{t}(\gamma))$ and all derivatives are computed at $\hat{t}(\gamma)$ and $\theta_*(\gamma)$.

\section{Model Parameters}
\label{app:details}

    Here we provide additional information about the model parameters used in our experiments.
    
    \paragraph{Image augmentations and feature extractor parameters.}
    For \omni\ dataset, we used the same image augmentations that were originally proposed in \maml.
    For \miniim\ and \tiered\ datasets, however, we used ImageNet-style image augmentations including horizontal image flipping, random color augmentations and random image cropping.
    This helped us to avoid model overfitting on the \miniim\ dataset and possibly on \tiered.
    
    The dimensionality $\embeddingdim$ of the label encoding $\xi$ and weight slice encoding $\mu$ was typically set to $32$.
    Increasing $\embeddingdim$ up to the maximum number of weight slices plus the number of per-episode labels, would allow the model to fully disentangle examples for different labels and different weight slices, but can also make the model train slower.
    
    \paragraph{Transformer parameters.}
    Since the weight tensors of each layer are generally different, our per-layer transformers were also different.
    The key, query and value dimensions of the transformer were chosen to be equal to a pre-defined fraction $\nu$ of the input embedding size, which in turn was a function of the label, image and activation embedding sizes and the sizes of the weight slices.
    The inner dimension of the final fully-connected layer in the transformer was also chosen using the same approach.
    In our \miniim\ and \tiered\ experiments, $\nu$ was chosen to be $0.5$ and in \omni\ experiments, we used $\nu=1$.
    Each transformer typically contained $2$ or $3$ encoder layers and used $2$ or $8$ heads for \omni\ and \miniim, \tiered, correspondingly.
    
    \paragraph{Learning schedule.}
    In all our experiments, we used gradient descent optimizer with a learning rate in the $0.01$ to $0.02$ range.
    Our early experiments with more advanced optimizers were unstable.
    We used a learning rate decay schedule, in which we reduced the learning rate by a factor of $0.95$ every $10^5$ learning steps.

\section{Additional Supervised Experiments}
    \label{app:supervised}

    While the advantage of decoupling parameters of the weight generator and the generated CNN model is expected to vanish with the growing CNN model size, we compared our approach to two other methods, LGM-Net \citep{li2019lgm} and LEO \citep{rusu2019metalearning}, to verify that our approach can match their performance on sufficiently large models.
    
    For our comparison with the LGM-Net method, we used the same image augmentation technique that was used in \cite{li2019lgm} where it was applied both at the training and the evaluation stages \citep{lgmnet-github}.
    We also used the same CNN architecture with 4 learned 64-channel convolutional layers followed by two generated convolutional layers and the final logits layer.
    In our weight generator, we used 2-layer transformers with activation feature extractors that relied on 48-channel convolutional layers and did not use any image embeddings.
    We trained our model in the end-to-end fashion on the \miniim\ 1-shot-5-way task and obtained a test accuracy of $69.3\%\pm 0.3\%$ almost identical to the $69.1\%$ accuracy reported in \cite{li2019lgm}.

    We also carried out a comparison with LEO by using our method to generate a fully-connected layer on top of the \tiered{} embeddings pre-computed with a WideResNet-28 model employed by \cite{rusu2019metalearning}.
    For our experiments, we used a simpler 1-layer transformer model with 2 heads that did not have the final fully-connected layer and nonlinearity.
    We also used $L_2$ regularization of the generated fully-connected weights setting the regularization weight to $10^{-3}$.
    As a result of training this model, we obtained $66.2\%\pm 0.2\%$ and $81.6\%\pm 0.2\%$ test accuracies on the 1-shot-5-way and 5-shot-5-way \tiered{} tasks correspondingly.
    These results are almost identical to $66.3\%$ and $81.4\%$ accuracies reported in \cite{rusu2019metalearning}.

\section{Dependence on Parameters and Ablation Studies}
    \label{app:ablation}

\begin{figure}
        \centering
        \includegraphics[width=1\textwidth]{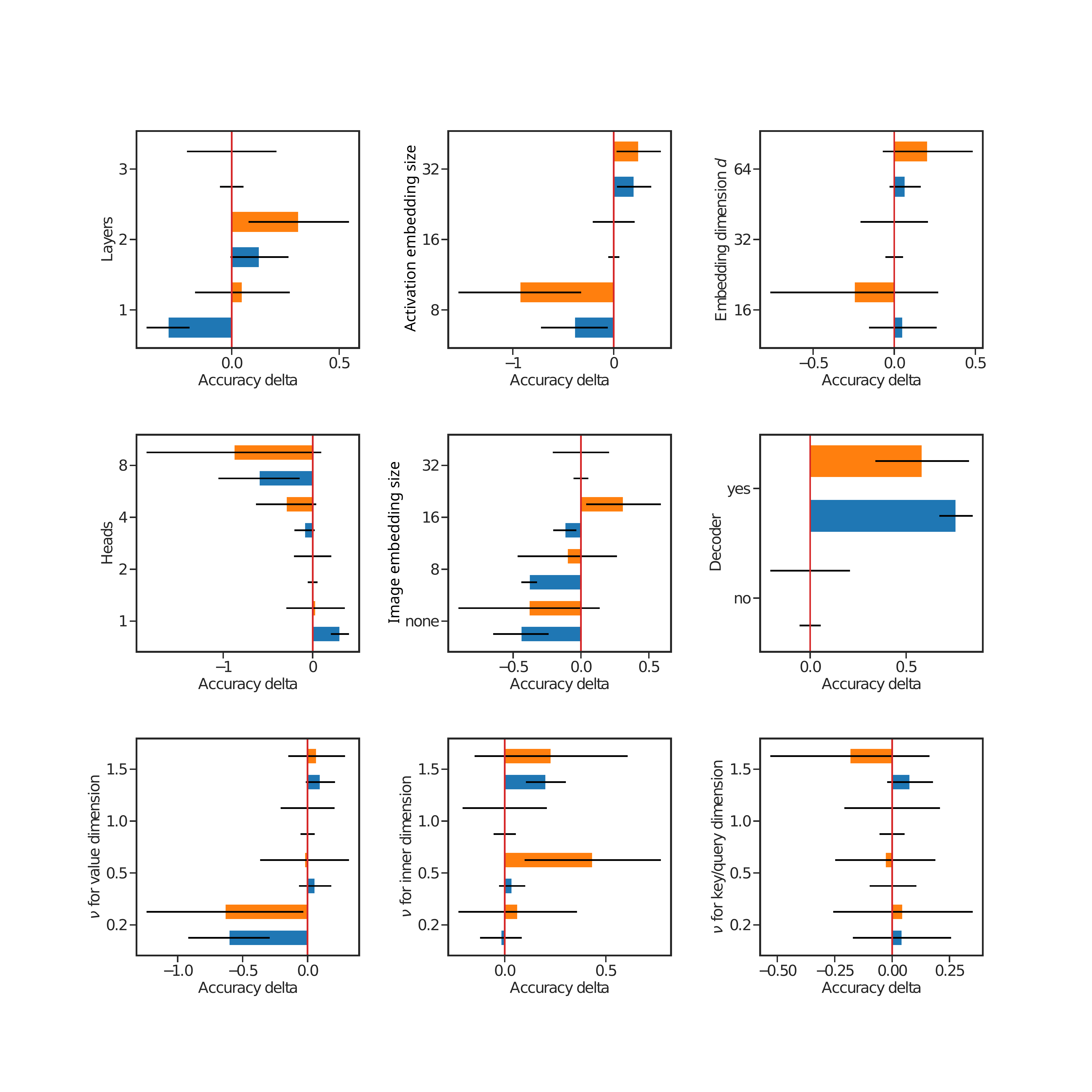}
        \caption{
            Change of the training (blue) and test (orange) accuracies on 1-shot-20-way \omni\ task for a 16-channel model relative to the {\em base} configuration with 3-layer transformer, 16-dimensional activation embedding, $\nu=1.0$, $\embeddingdim=32$, $2$ heads and 32-dimensional image embedding.
            Approximate confidence intervals are shown.
        }
        \label{fig:params-16}
    \end{figure}

\begin{figure}
        \centering
        \includegraphics[width=1\textwidth]{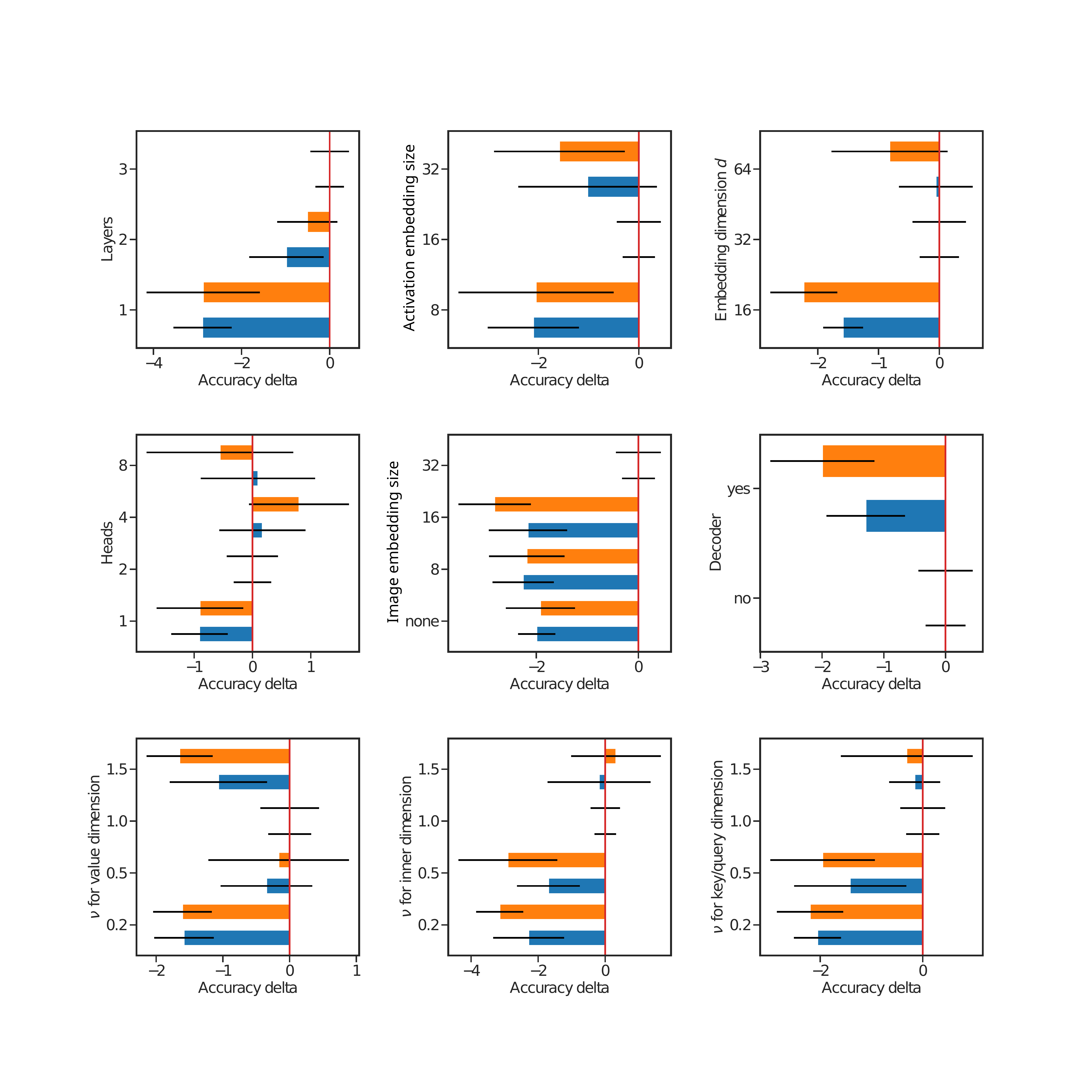}
        \caption{
            Change of the training (blue) and test (orange) accuracies on 5-shot-20-way \omni\ task for a 4-channel model relative to the {\em base} configuration with 3-layer transformer, 16-dimensional activation embedding, $\nu=1.0$, $\embeddingdim=32$, $2$ heads and 32-dimensional image embedding.
            Approximate confidence intervals are shown.
        }
        \label{fig:params-4}
    \end{figure}

    Most of our parameter explorations were conducted for \omni\ dataset.
    We chose a 16-channel model trained on a 1-shot-20-way \omni\ task as an example of a model, for which just the logits layer generation was sufficient.
    We also chose a 4-channel model trained on a 5-shot-20-way \omni\ task for the role of a model, for which generation of all convolutional layers proved to be beneficial.
    Figures~\ref{fig:params-16} and \ref{fig:params-4} show comparison of training and test accuracies on \omni\ for different parameter values for these two models.
    Here we only used two independent runs for each parameter value, which did not allow us to sufficiently reduce the statistical error.
    Despite of this, in the following, we try to highlight a few notable parameter dependencies.
    Note here that in some experiments with particularly large embedding or model sizes, training progressed beyond the target number of steps and there could also be overfitting for very large models.

    \paragraph{Number of transformer layers.}
    Increasing the number of transformer layers is seen to be particularly important in the 4-channel model.
    The 16-channel model also demonstrates the benefit of using $1$ vs $2$ transformer layers, but the performance appears to degrade when we use $3$ transformer layers.

    \paragraph{Activation embedding dimension.}
    Particularly, small activation embeddings can be seen to hurt the performance in both models, while using larger activation embeddings appears to be advantageous in most cases except for the 32-dimensional activation embeddings in the 4-channel model.

    \paragraph{Class embedding dimension.}
    Particularly low embedding dimension of 16 can be seen to hurt the performance of both models.

    \paragraph{Number of transformer heads.}
    Increasing the number of transformer heads leads to performance degradation in the 16-channel model, but does not have a pronounced effect in the 4-channel model.

    \paragraph{Image embedding dimensions.}
    Removing the image embedding, or using an 8-dimensional embedding can be seen to hurt the performance in both cases of the 4- and 16-channel models.

\begin{figure}
        \centering
        \includegraphics[width=.7\textwidth]{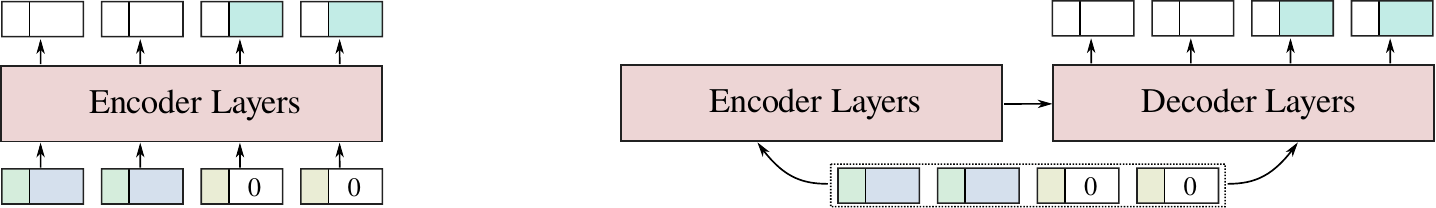}
        \caption{
            Two transformer-based approaches to weight generation studied in our experiments: only encoder layers (left), encoder and decoder layers on the same input sequence (right).
        }
        \label{fig:encoder_decoder}
    \end{figure}

    \paragraph{Transformer architecture.}
    While the majority of our experiments were conducted with a sequence of transformer encoder layers, we also experimented with an alternative weight generation approach, where both encoder and decoder transformer layers were employed (see Fig.~\ref{fig:encoder_decoder}).
    Our experiments with both architectures suggest that the role of the decoder is pronounced, but very different in two models: in the 16-channel model, the presence of the decoder increases the model performance, while in the 4-channel model, it leads to accuracy degradation.

    \paragraph{Inner transformer embedding sizes.}
    Varying the $\nu$ parameter for different components of the transformer model (key/query pair, value and inner fully-connect layer size), we quantify their importance on the model performance.
    Using very low $\nu$ for the value dimension hurts performance of both models.
    The effect of key/query and inner dimensions can be distinctly seen only in the 4-channel model, where using $\nu=1$ or $\nu=1.5$ appears to produce the best results.

    \paragraph{Weight allocation approach.}
    Our experiments with the ``spatial'' weight allocation in 4- and 16-channel models showed slightly inferior performance (both accuracies dropping by about $0.2\%$ to $0.4\%$ in both experiments) compared to that obtained with the ``output'' weight allocation method.

\section{Attention maps of learned transformer models}

\begin{figure}
        \centering
        \includegraphics[width=1\textwidth]{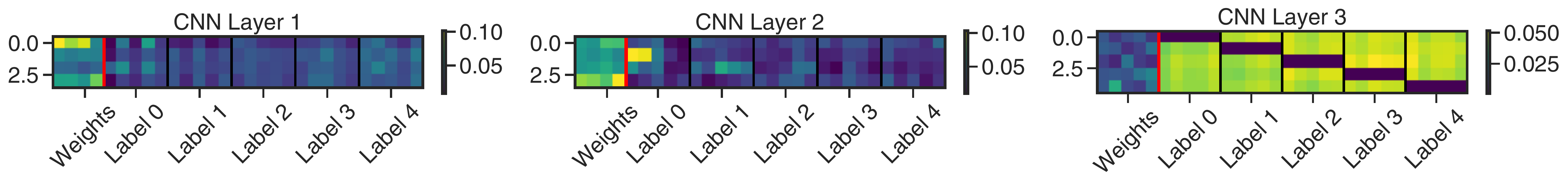}
        \caption{Learned attention maps for 2-layer 4-channel CNN network generated with 1 head, 1 layer transformer for 5-shot \miniim.}
        \label{fig:attention-weights-2-layers}
    \end{figure}

    We visualized the attention maps of several transformer-based models that we used for CNN layer generation.
    Figure~\ref{fig:attention-weights-2-layers} shows attention maps for a 2-layer 4-channel CNN network generated using a 1-head 1-layer transformer on \miniim (labeled samples are sorted in the order of their episode labels).
    Attention map for the final logits layer (``CNN Layer 3'') is seen to exhibit a ``stairways'' pattern indicating that a weight slice $W_{c,\cdot}$ for episode label $c$ is generated by attending to all samples except for those with label $c$.
    This is reminiscent of the supervised learning mechanism outlined in Sec.~\ref{sec:reasoning}.
    While the proposed mechanism would attend to all samples with label $c$ and average their embeddings, another alternative is to average embeddings of samples with other labels and then invert the result.
    We hypothesize that the trained transformer performs a similar calculation with additional learned transformer parameters, which may be seen to result in mild fluctuations of the attention to different input samples.
    
\begin{figure}
        \centering
        \begin{tabular}[c]{@{\hspace{0.0\linewidth}}c@{\hspace{0.0\linewidth}}c@{\hspace{0.0\linewidth}}}
        Transformer Layer 1 & Transformer Layer 2 \\
        \includegraphics[width=.5\textwidth]{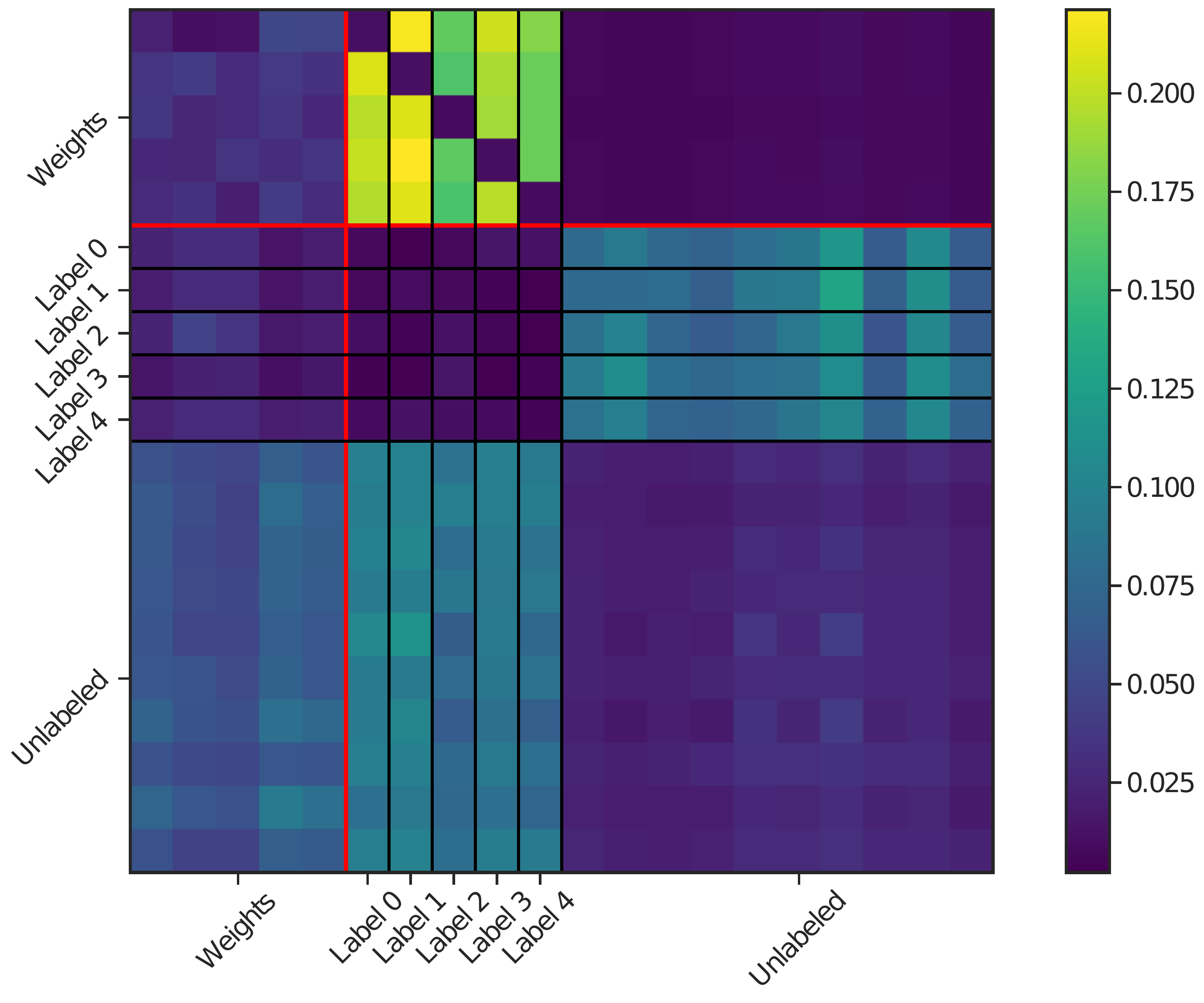} & 
        \includegraphics[width=.5\textwidth]{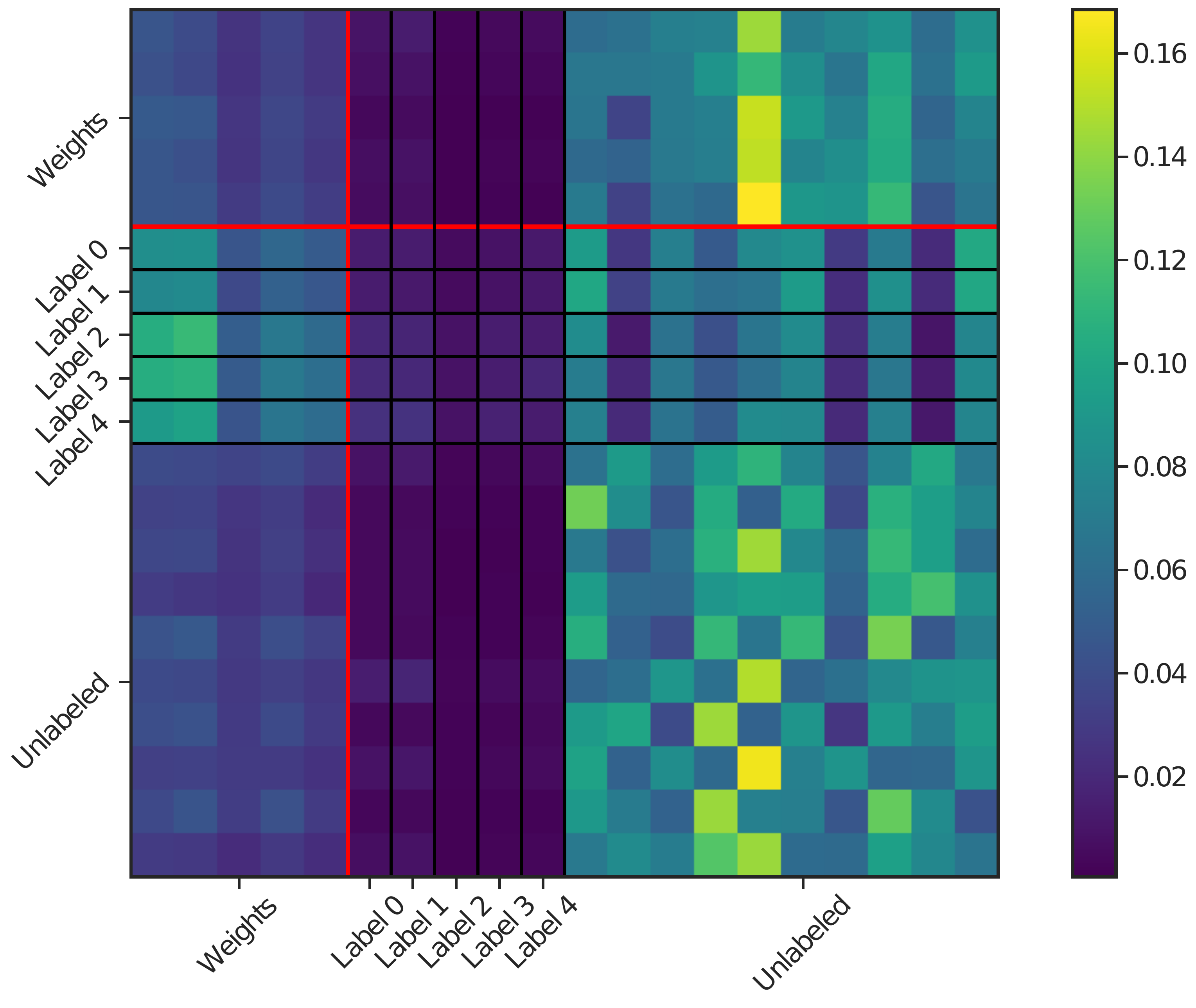}
        \end{tabular}
        \caption{Learned attention maps for 4-layer 8-channel CNN network generated with 1-head, 2-layer transformer for 5-shot \tiered\ with additional unsupervised samples (2 per class). Only the last layer of CNN is generated.}
        \label{fig:attention-weights-2-layers-b}
    \end{figure}

    The attention maps for a semi-supervised learning problem with a 2-layer transformer is shown in Figure~\ref{fig:attention-weights-2-layers-b}.
    One thing to notice is that a mechanism similar to the one described above appears to be used in the first transformer layer, where weight slices $W_{c,\cdot}$ attend to all labeled samples with labels $c_i \ne c$.
    At the same time, unlabeled samples can be seen to attend to labeled samples in layer 1 (see ``Unlabeled'' rows and ``Label \dots'' columns) and the weight slices in layer 2 then attend to the updated unlabeled sample tokens (see ``Weights'' rows and ``Unlabeled'' columns in the second layer).
    This additional pathway connecting labeled samples to unlabeled samples and finally to the logits layer weights is again reminiscent of the simplistic semi-supervised learning mechanism outlined in Sec.~\ref{sec:reasoning}.

    The exact details of these calculations and the generation of intermediate convolutional layers is generally much more difficult to interpret just from the attention maps and a more careful analysis of the trained model is necessary to draw the final conclusions.

\section{UMAP embedding of the generated weights.}
\label{app:umap}

    \begin{figure}
        \centering
        \begin{tabular}[c]{@{\hspace{0.0\linewidth}}c@{\hspace{0.0\linewidth}}c@{\hspace{0.0\linewidth}}}
        & Layer 1 \hspace{4em} Layer 2 \hspace{4em} Layer 3 \hspace{4em} Layer 4 \hspace{4em} All layers \\
        \rotatebox{90}{ \hskip1.5em Class 10 \hskip4.5em Class 85 \hskip3.5em Class 39 \hskip4.5em Class 60 \hskip4em Class 84 \hskip5em Class 8} &
        
        \includegraphics[width=.8\textwidth]{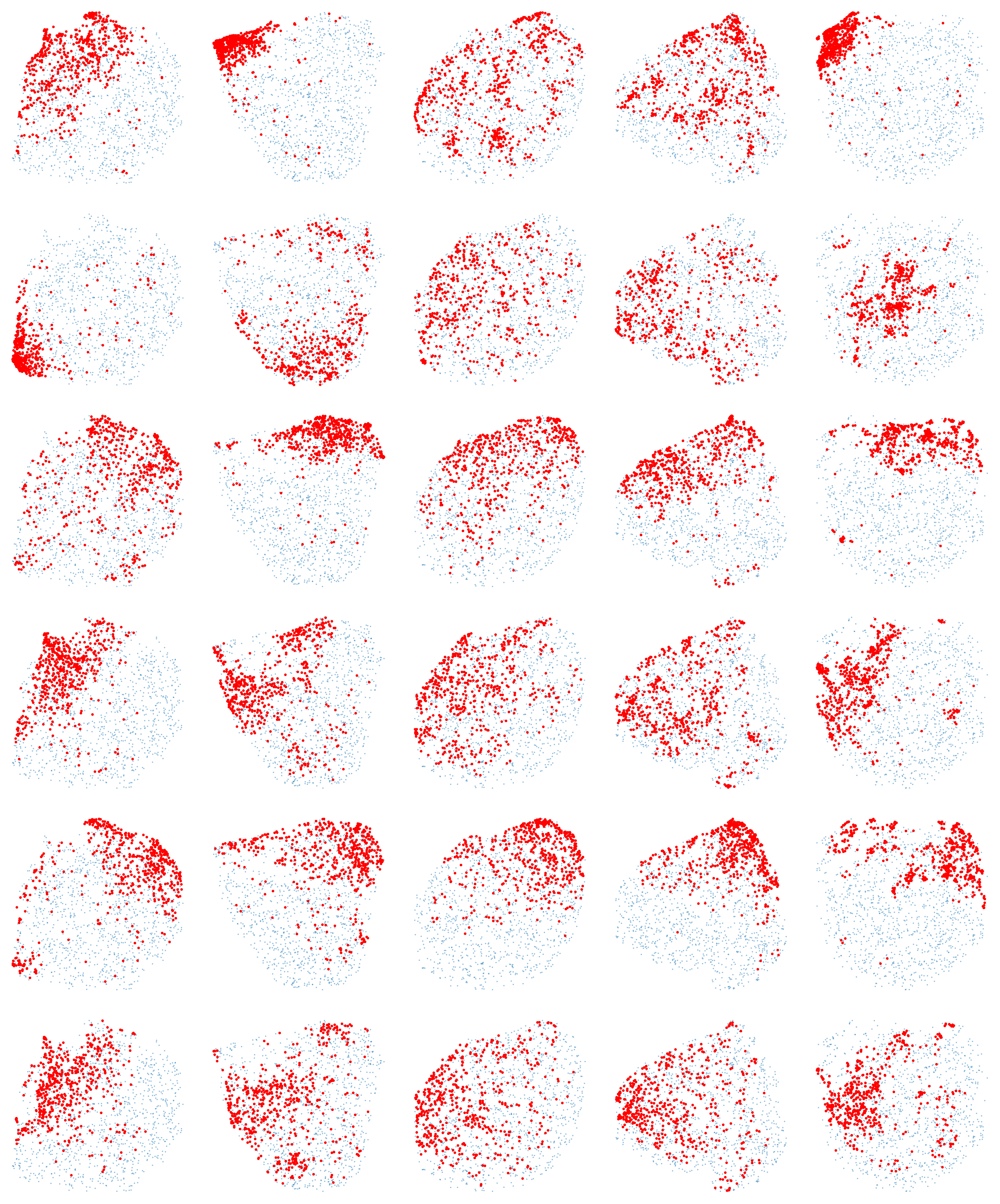}
        \end{tabular}
        \caption{UMAP embedding of weights generated by the \textsc{HyperTransformer}. Each point corresponds to the embedding of the weights of a given layer for a given episode. The original dimensionality for the first layer weights is $162=3\times3\times3\times6$, for each subsequent weights is $324=3\times3\times6\times6$ and for all layers combined is 1\,134. We further selected 6 different classes with the smallest standard deviation in the embeddings and highlighted the episodes that contain these classes.}
        \label{fig:umap-full}
    \end{figure}
    
    \begin{figure}
        \centering
        \begin{tabular}[c]{@{\hspace{0.0\linewidth}}c@{\hspace{0.01\linewidth}}c@{\hspace{0.0\linewidth}}}
        \rotatebox{90}{\hskip2em Class 8} & \includegraphics[width=.8\textwidth]{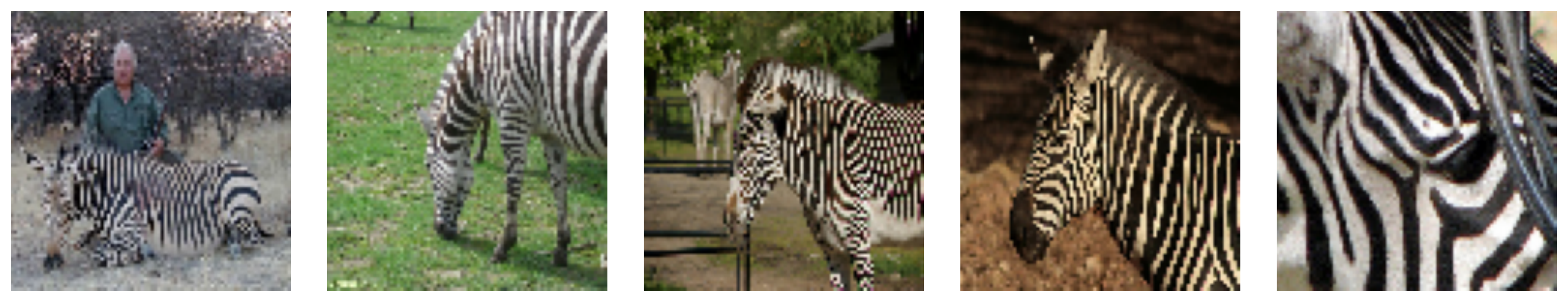} \\ 
        \rotatebox{90}{\hskip2em Class 84} & \includegraphics[width=.8\textwidth]{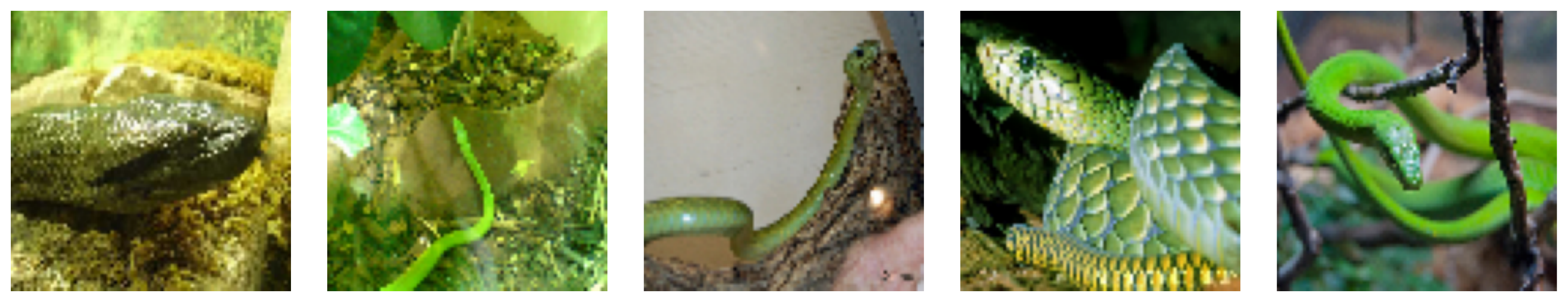} \\
        \rotatebox{90}{\hskip2em Class 60} & \includegraphics[width=.8\textwidth]{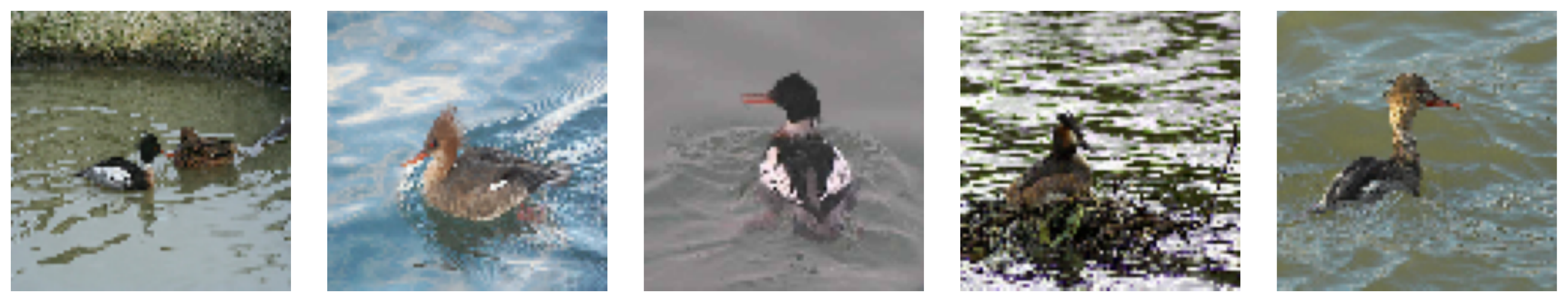} \\
        \rotatebox{90}{\hskip2em Class 39} & \includegraphics[width=.8\textwidth]{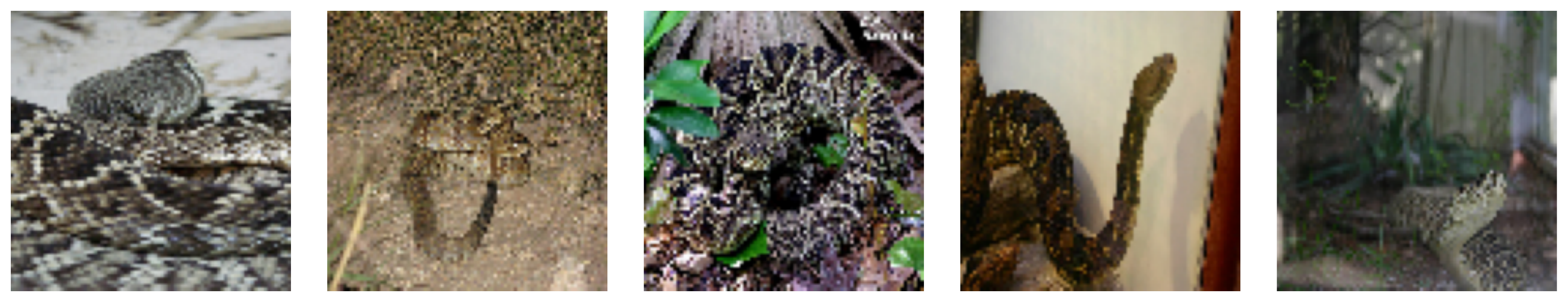} \\
        \rotatebox{90}{\hskip2em Class 85} & \includegraphics[width=.8\textwidth]{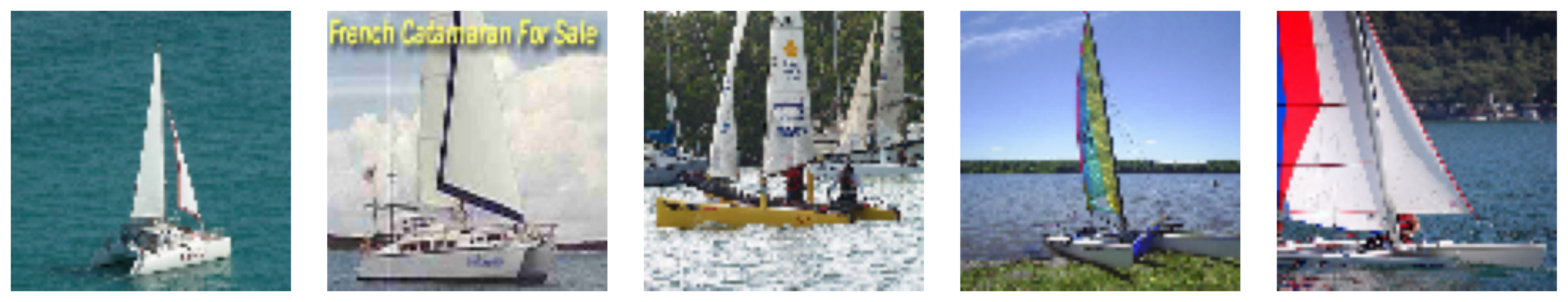} \\
        \rotatebox{90}{\hskip2em Class 10} & \includegraphics[width=.8\textwidth]{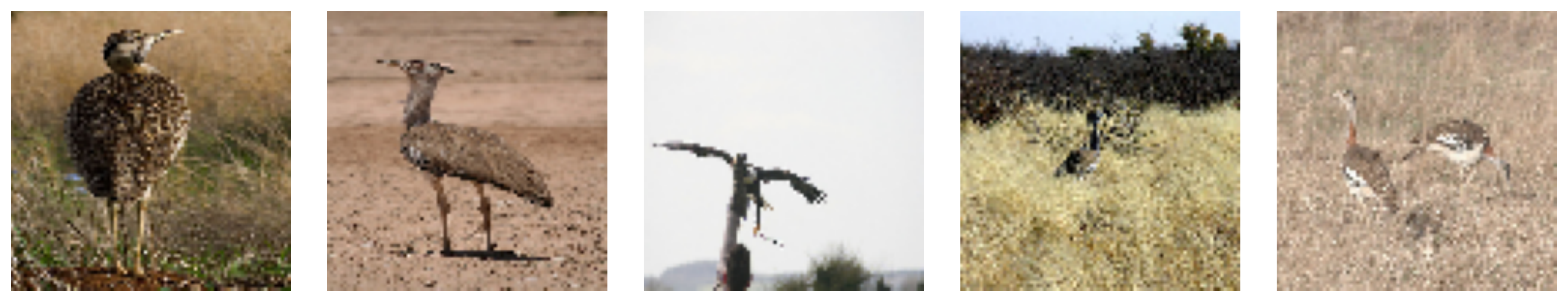} \\
        \end{tabular}
        \caption{Examples of classes from $\tiered{}$ that we used to highlight in UMAP embedding.}
        \label{fig:classes}
    \end{figure}    
    
  Figure~\ref{fig:umap-full} shows additional plots of the UMAP embedding with some of the classes highlighted. Here, we can more clearly see that, at least for those classes, the embedding is quite correlated with the episodes where these classes are included. This suggests that the \textsc{HyperTransformer} does generate meaningful \emph{individualized} weights for each of the episode. Figure~\ref{fig:classes} shows some samples of the highlighted classes. Notice that {\em Class 8} is clustered more tightly for the second layer, which means that the weights for the episodes containing this class are much more similar specifically for that layer. Indeed, {\em Class 8} corresponds to zebras, with their stripes pattern being a distinct feature. In order to be able to distinguish this feature, CNN would need to aggregate information from a wide field of view. This would probably not happen in the first layer, thus we see almost no correlation for the first one, but very tight cluster for the second.
  
\section{Visualization of the Generated CNN Weights.}

\begin{figure}[t]
        \centering
        \begin{tabular}[c]{@{\hspace{0.0\linewidth}}c@{\hspace{0.0\linewidth}}c|@{\hspace{0.01\linewidth}}c@{\hspace{0.0\linewidth}}|c@{\hspace{0.0\linewidth}}}
         & Generated/Generated & Trained/Generated & Trained/Trained \\
        \rotatebox{90}{\hspace{5ex}CNN Layer 1} &
         \includegraphics[width=0.3\textwidth]{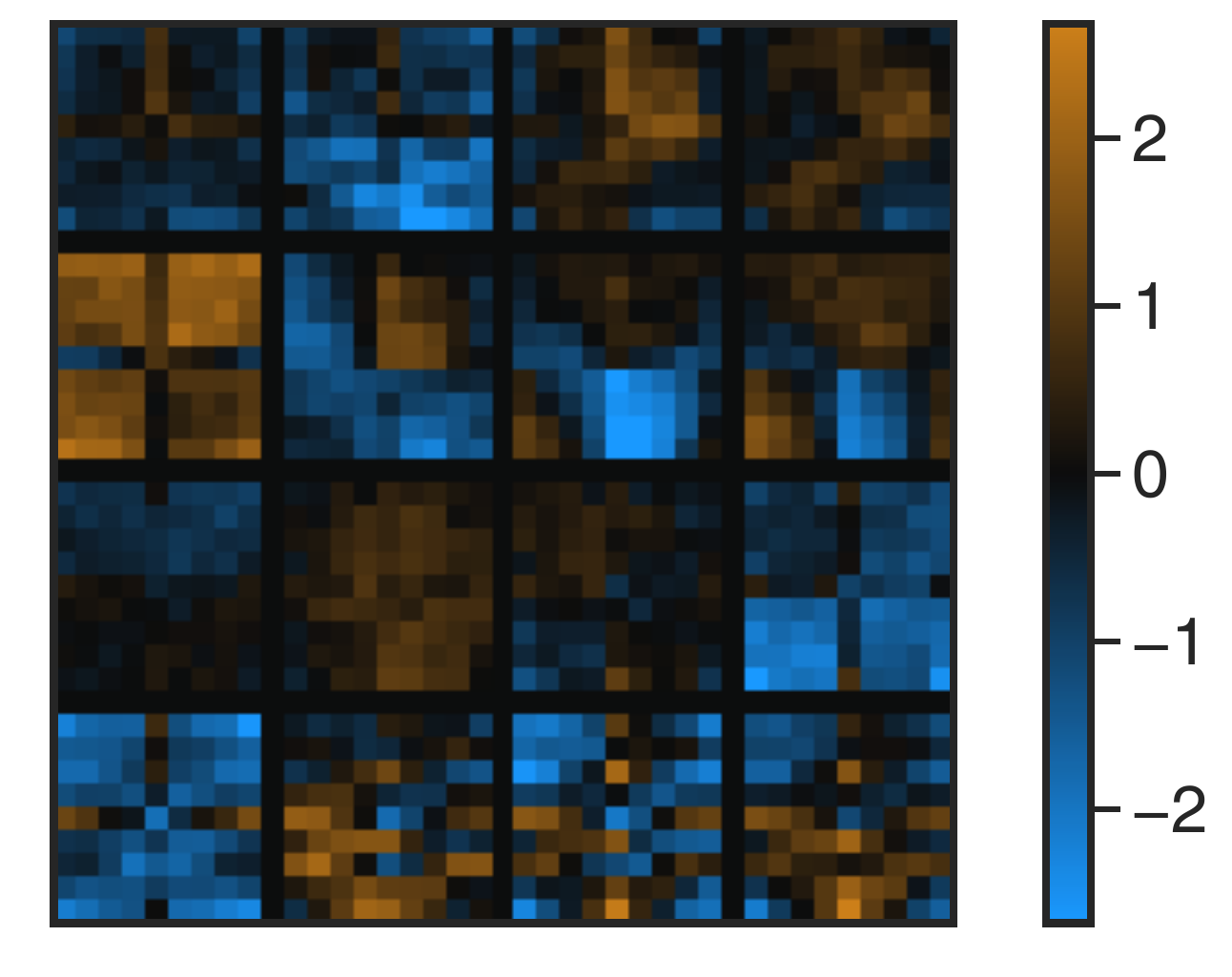} &
         \includegraphics[width=0.3\textwidth]{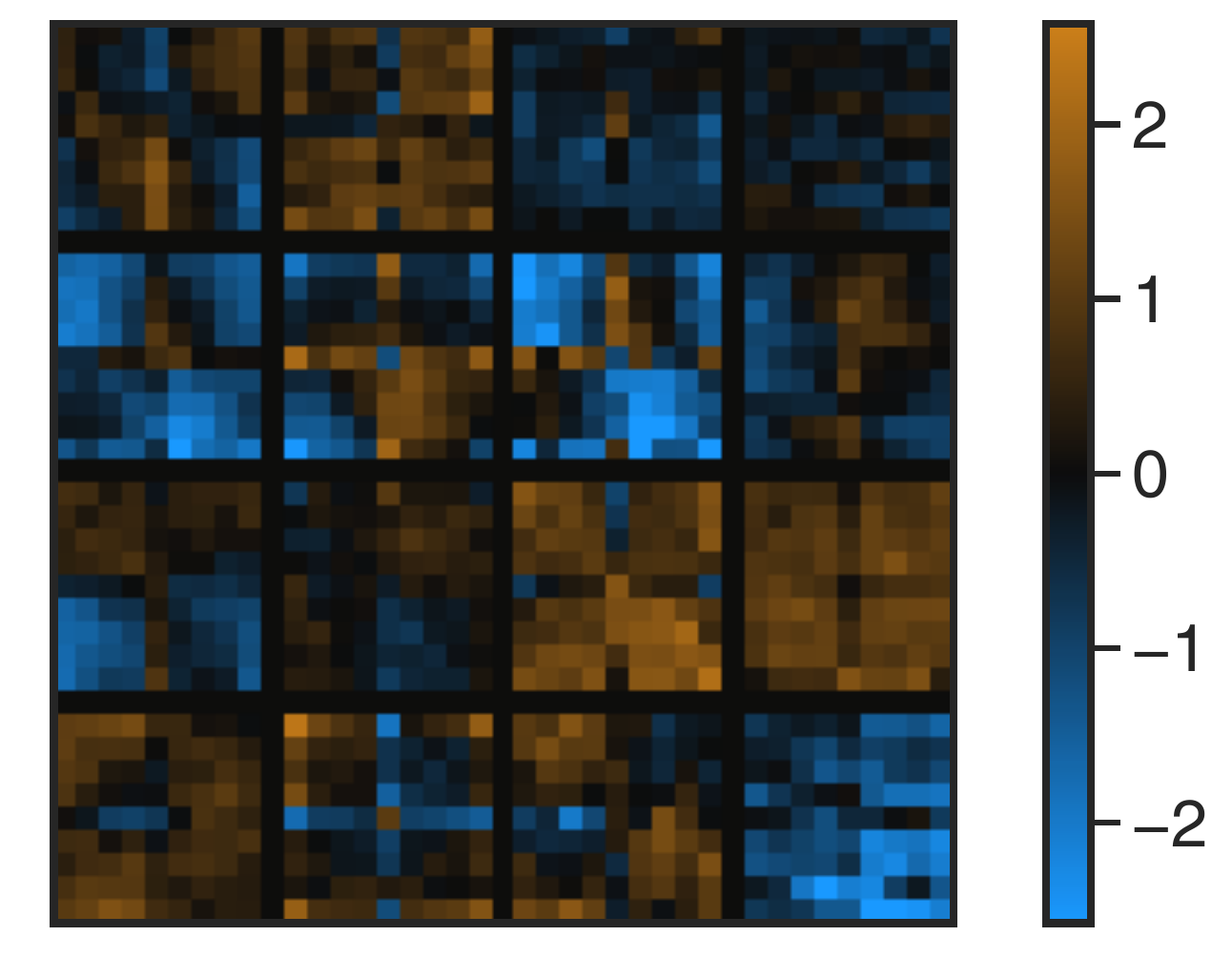} &
         \includegraphics[width=0.3\textwidth]{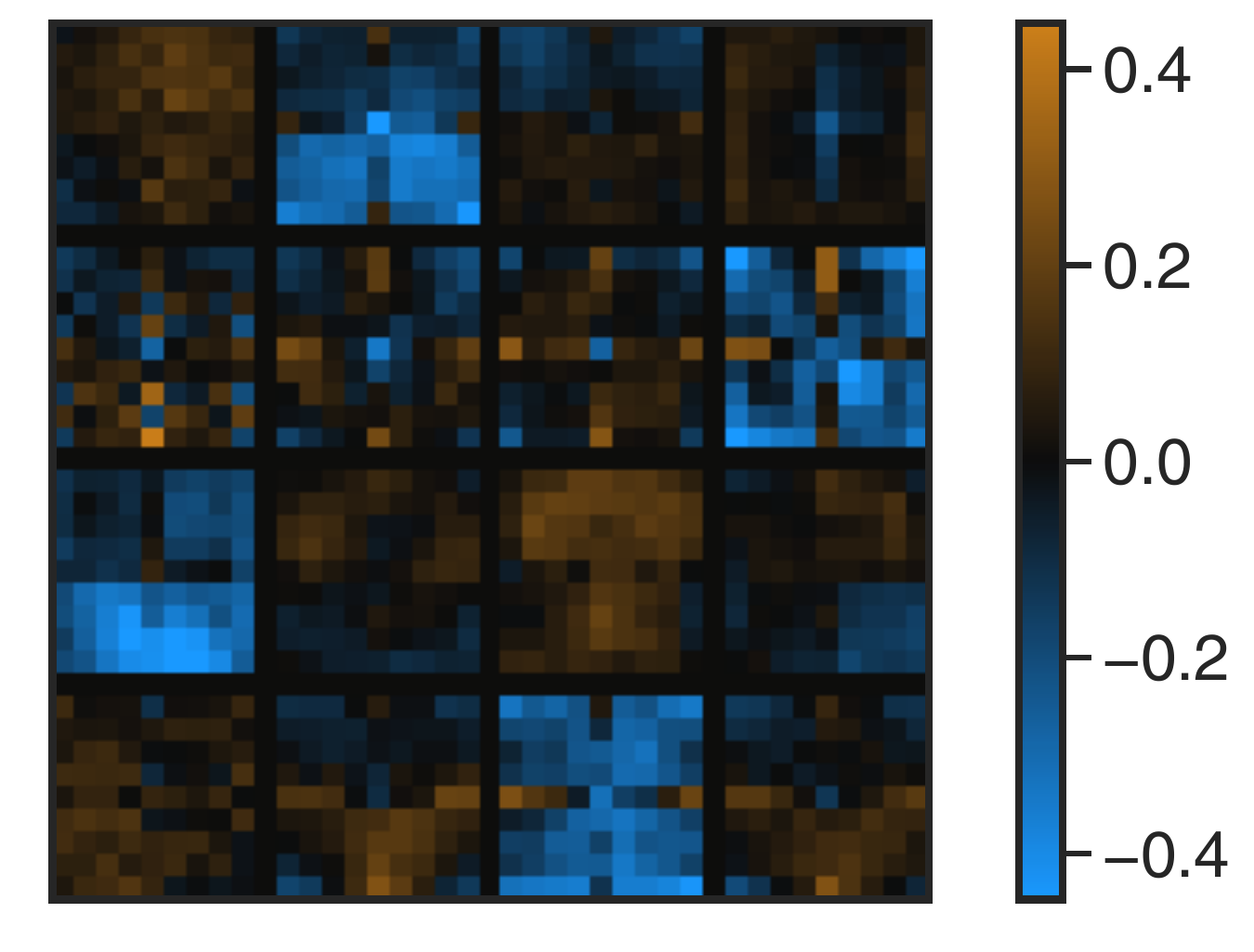} \\
        \rotatebox{90}{\hspace{2ex}CNN Layer 0} &
         \hspace{-1ex}\includegraphics[width=0.29\textwidth]{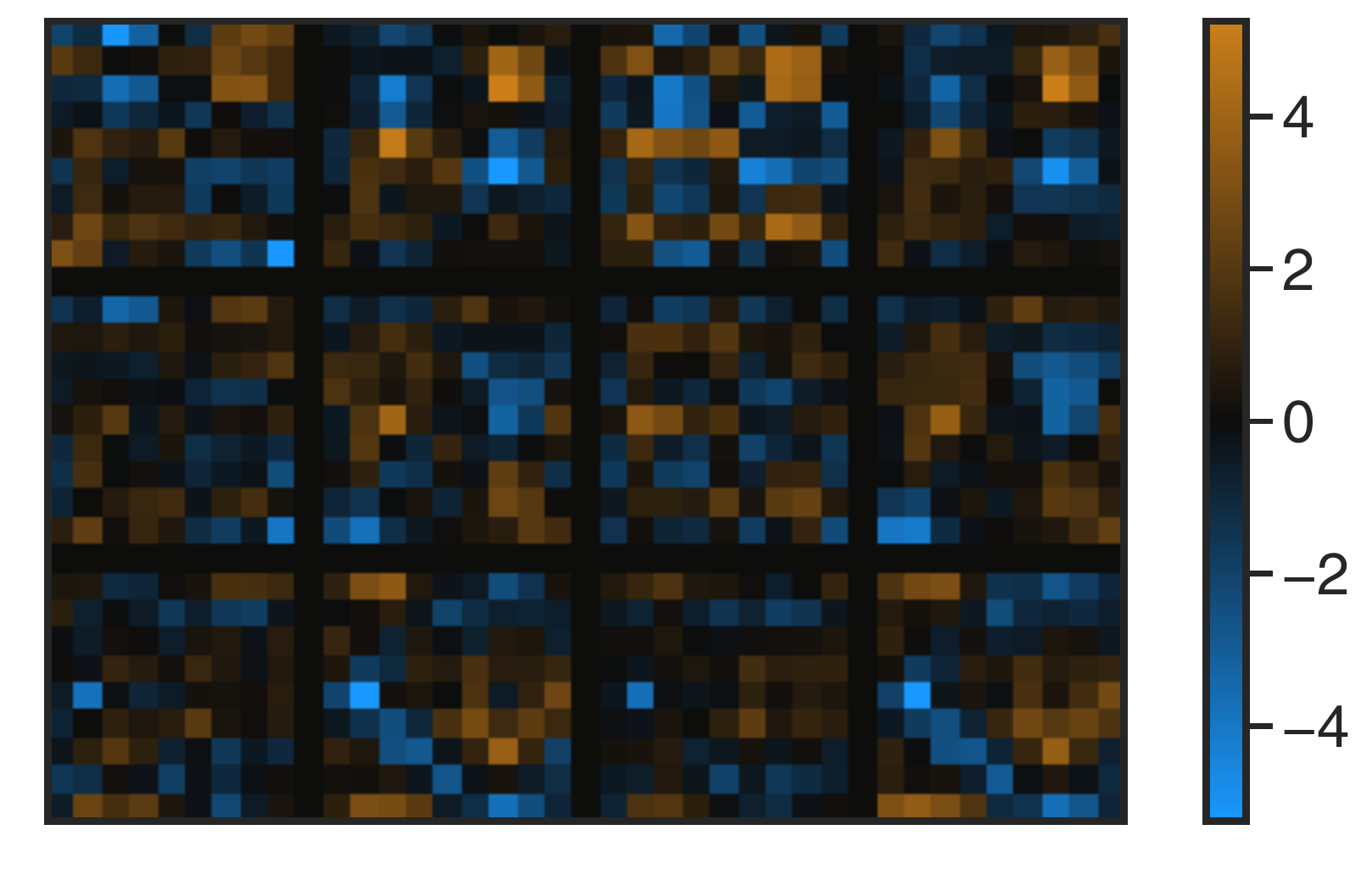} &
         \includegraphics[width=0.29\textwidth]{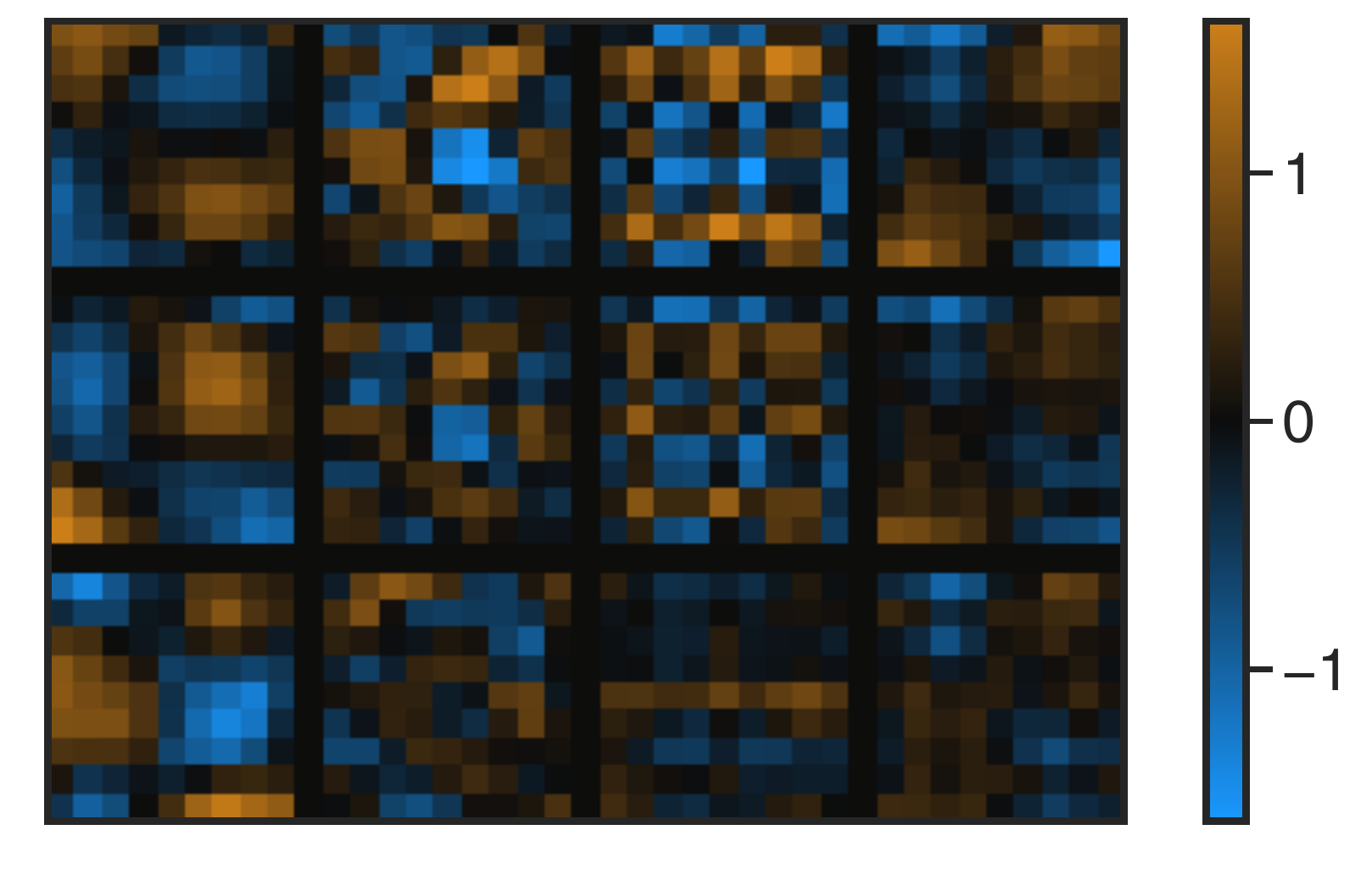} &
         \includegraphics[width=0.29\textwidth]{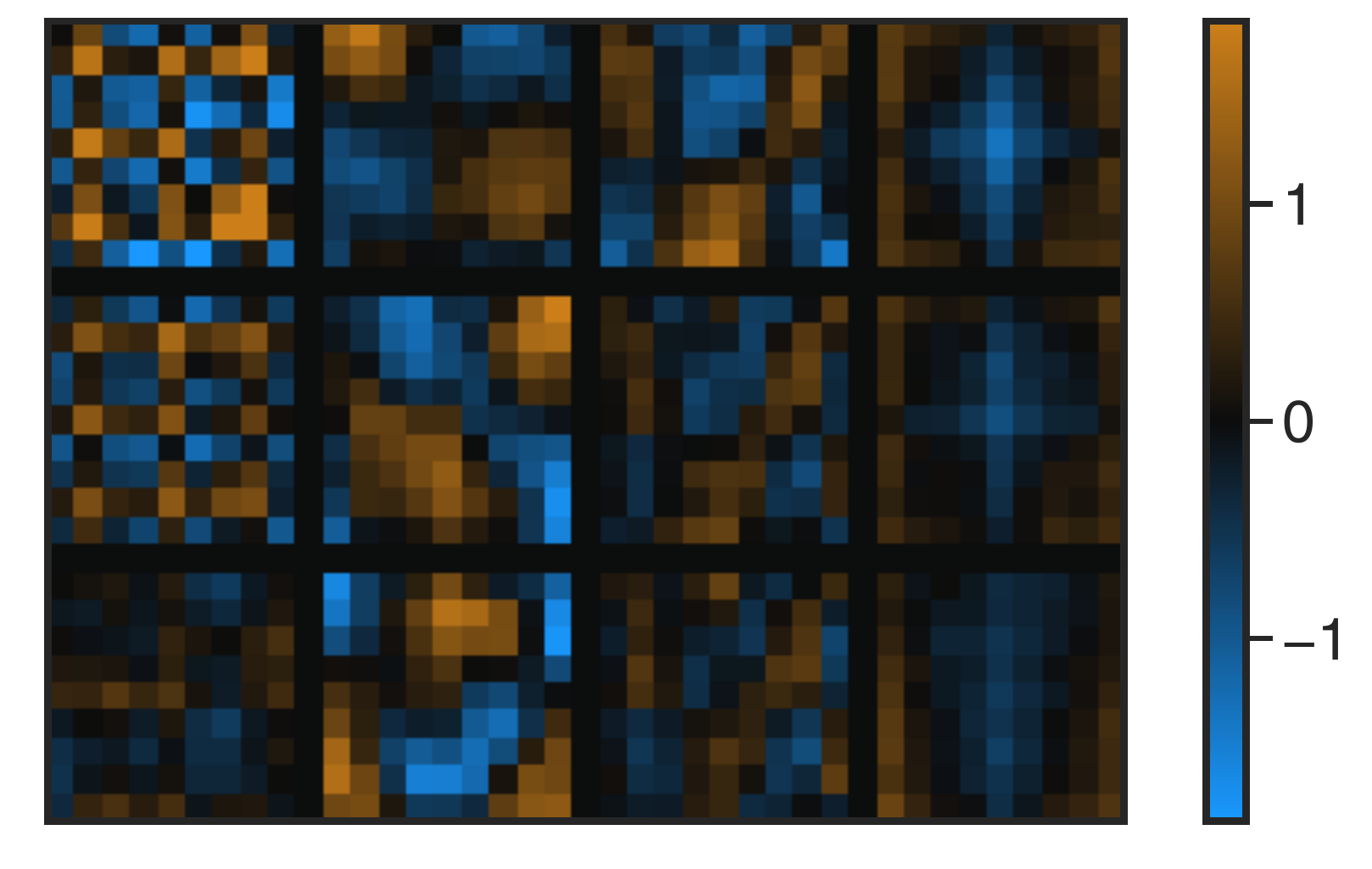}
        \end{tabular}
        \caption{Visualizing convolutional kernels for a 2-layer network with a $9\times 9$ CNN kernel size and stride of 4 trained on \miniim. Each column shows 1 model with two layers and fully-connected head that is always generated by the transformer. \emph{Left:} both CNN layers are generated, \emph{center:} first CNN layer is trained, second is generated, \emph{right:} both CNN layers are trained. Layer weight allocation: ``output''.}
        \label{fig:conv-kernels-output}
    \end{figure}

\begin{figure}
        \centering
        \begin{tabular}[c]{@{\hspace{0.0\linewidth}}c@{\hspace{0.0\linewidth}}c|@{\hspace{0.01\linewidth}}c@{\hspace{0.0\linewidth}}|c@{\hspace{0.0\linewidth}}}
         & Generated/Generated & Trained/Generated & Trained/Trained \\
        \rotatebox{90}{\hspace{5ex}CNN Layer 1} & 
         \includegraphics[width=0.3\textwidth]{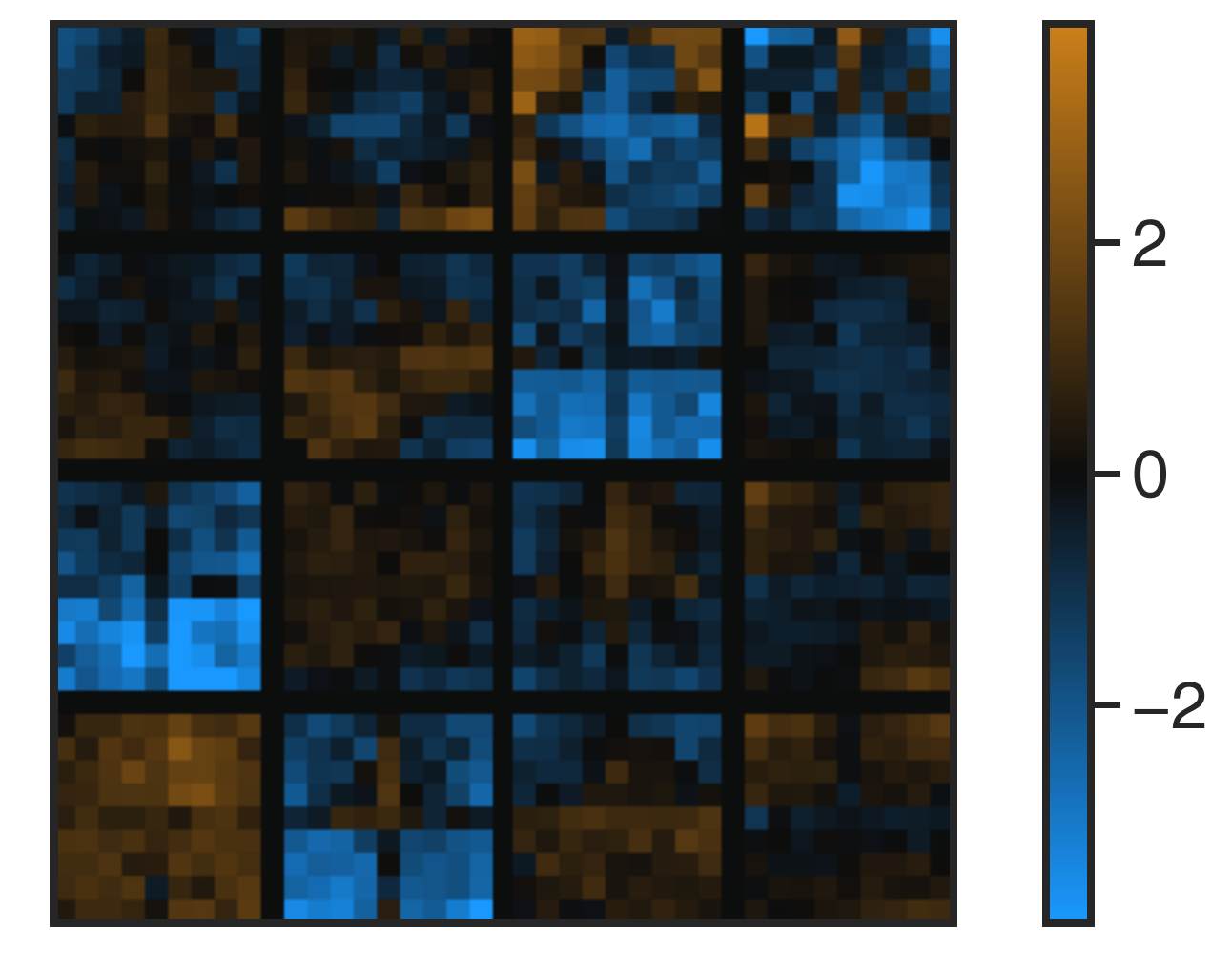} &
         \includegraphics[width=0.3\textwidth]{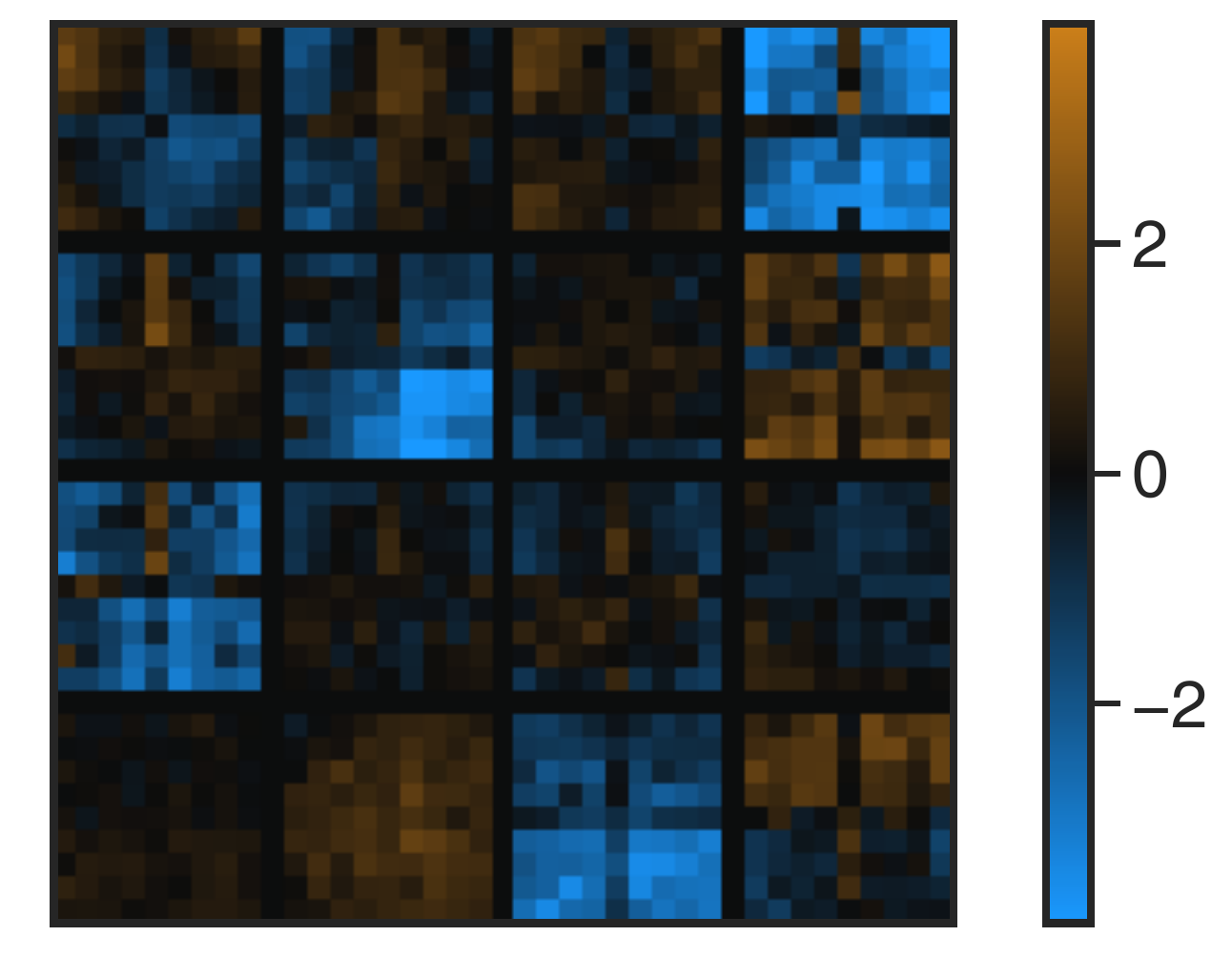} &
         \includegraphics[width=0.3\textwidth]{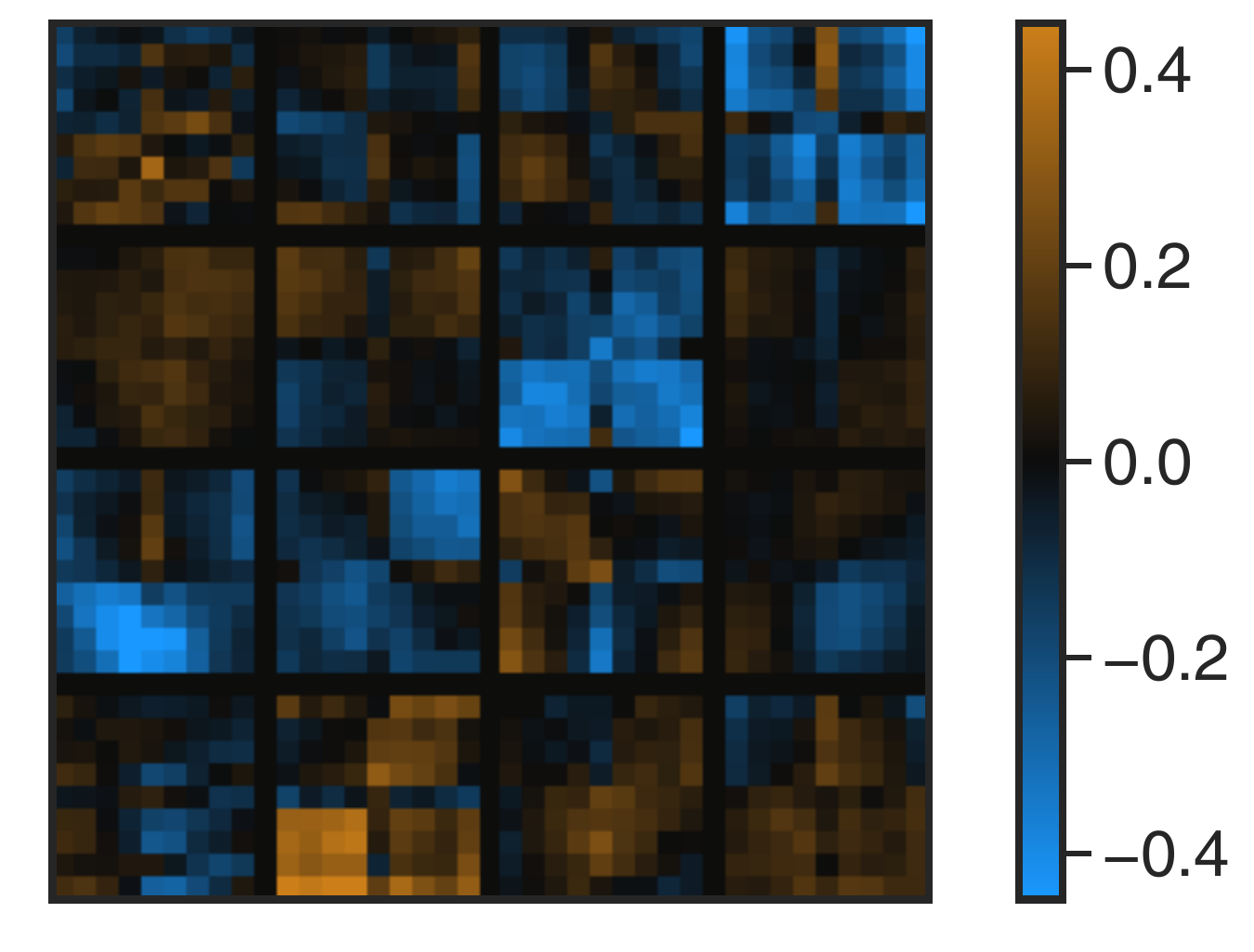} \\
        \rotatebox{90}{\hspace{2ex}CNN Layer 0} &
         \includegraphics[width=0.29\textwidth]{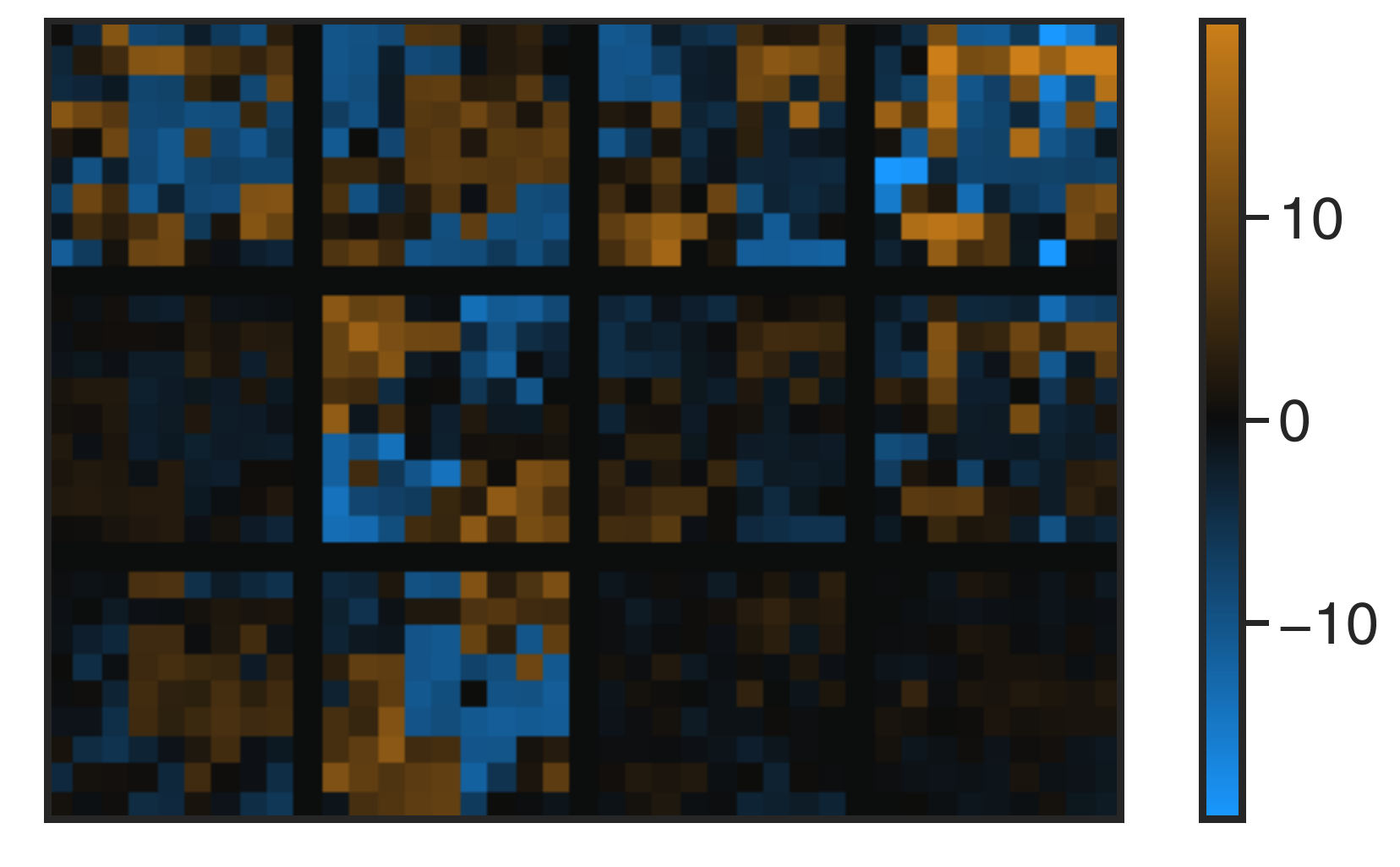} &
         \includegraphics[width=0.29\textwidth]{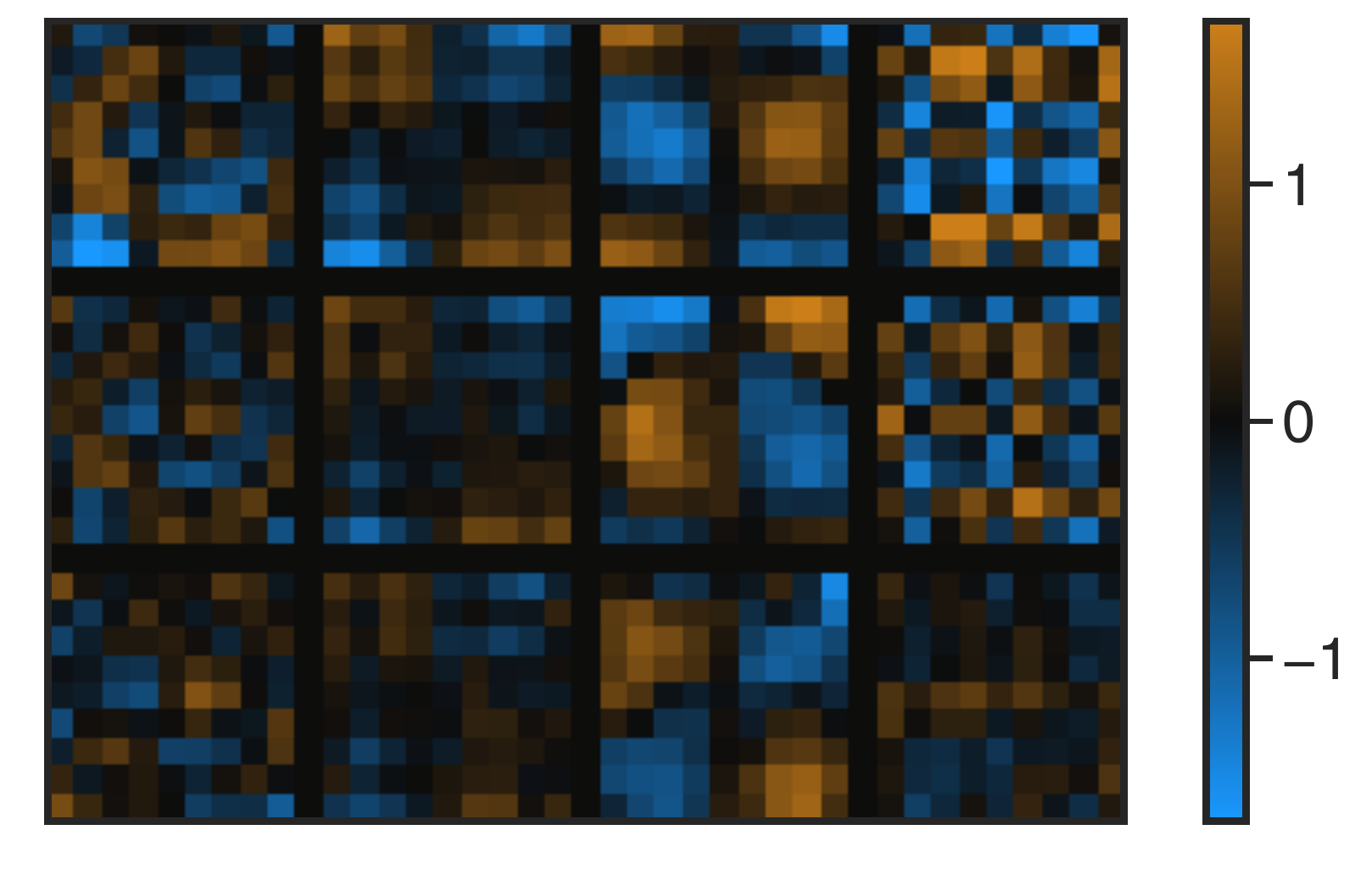} &
         \includegraphics[width=0.29\textwidth]{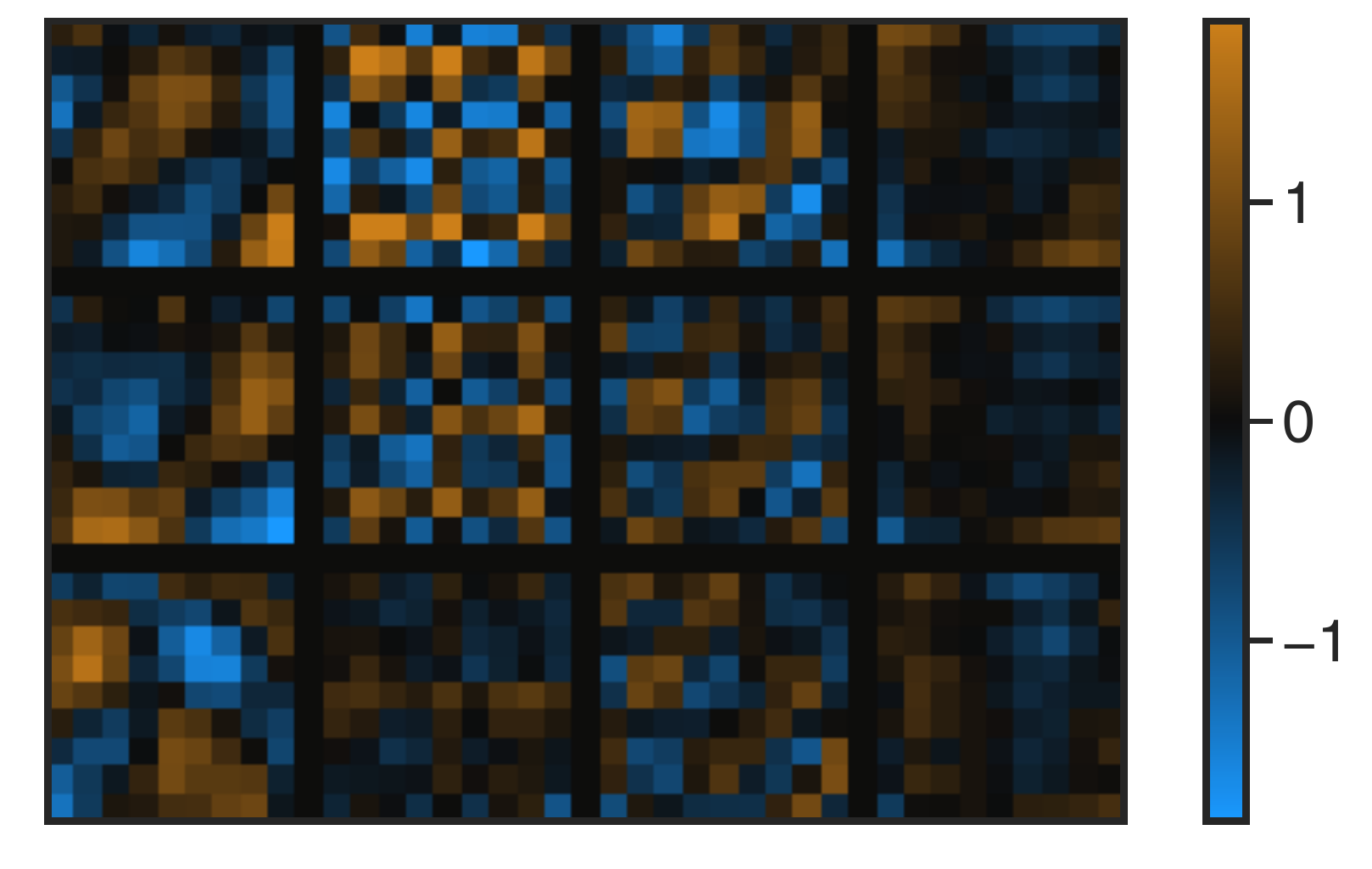} \\
        \end{tabular}
        \caption{Visualizing convolutional kernels for a 2-layer network with a $9\times 9$ CNN kernel size and stride of 4 trained on \miniim. Each column shows 1 model with two layers and fully-connected head that is always generated by the transformer. \emph{Left:} both CNN layers are generated, \emph{center:} first CNN layer is trained, second is generated, \emph{right:} both CNN layers are trained. Layer weight allocation: ``spatial''.}
        \label{fig:conv-kernels-spatial}
    \end{figure}

    Figures~\ref{fig:conv-kernels-output} and~\ref{fig:conv-kernels-spatial} show the examples of the CNN kernels that are generated by a single-head, 1-layer transformer for a simple 2-layer CNN model with $9 \times 9$ stride-4 kernels.
    Different figures correspond to different approaches to re-assembling the weights from the generated slices: using ``output'' allocation or ``spatial'' allocation (see Section~\ref{sec:model} in the main text for more information).
    Notice that ``spatial'' weight allocation produces more homogeneous kernels for the first layer when compared to the ``output'' allocation.
    In both figures we show the difference of the final generated kernels for 3 variants: model with both layers generated, one generated and one trained and both trained. 
    
\begin{figure}
        \centering
        \begin{tabular}[c]{@{\hspace{0.0\linewidth}}c@{\hspace{0.0\linewidth}}c@{\hspace{0.01\linewidth}}c@{\hspace{0.0\linewidth}}||c@{\hspace{0.0\linewidth}}}
         & Episode 0 & Episode 1 & Difference \\
        \rotatebox{90}{\hspace{5ex}Layer 1} &
         \includegraphics[width=0.31\textwidth]{figures/kernels/kernel_layer_1_tcnn_0_lwa_output_ep_0.pdf} &
         \includegraphics[width=0.31\textwidth]{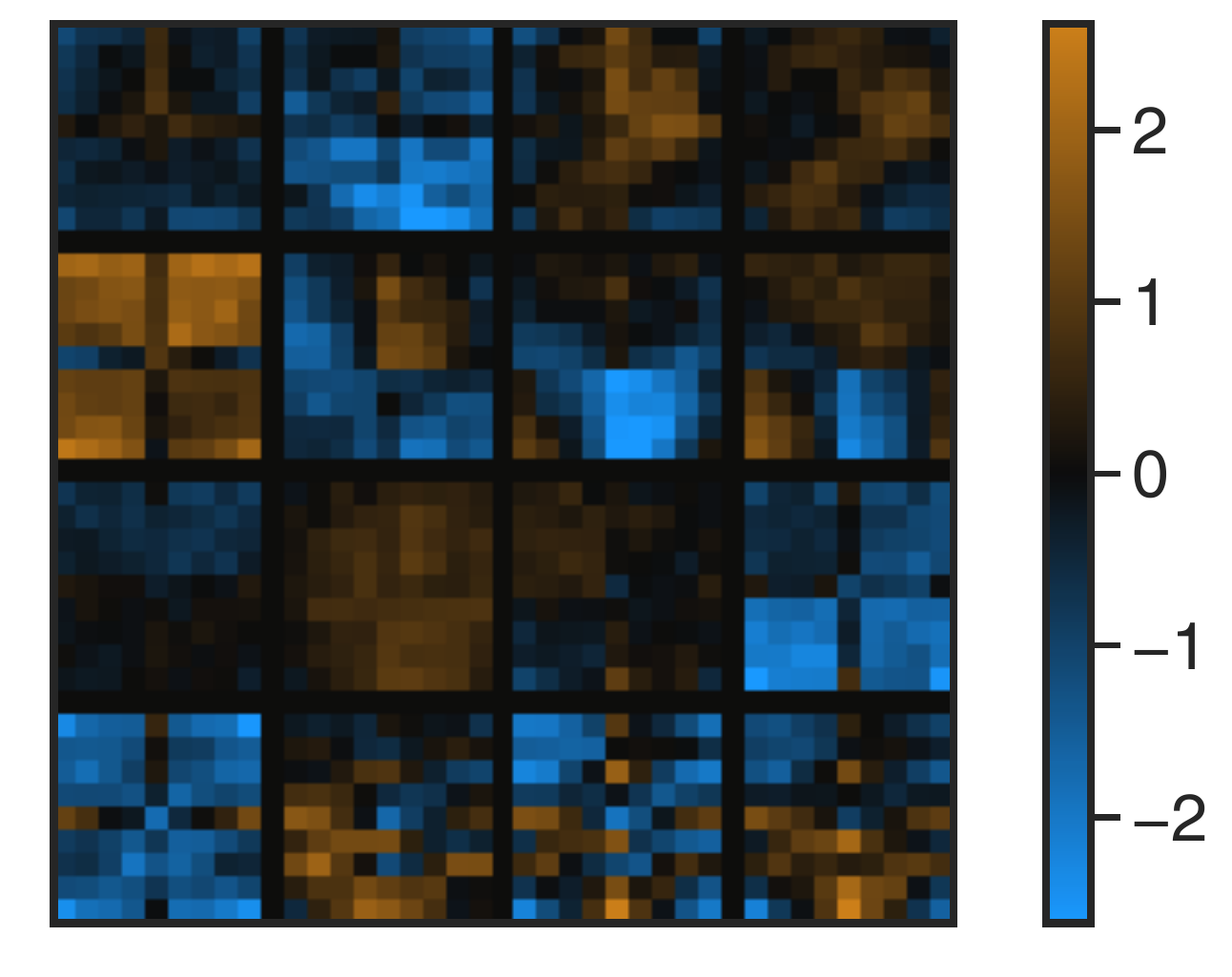} &
         \includegraphics[width=0.33\textwidth]{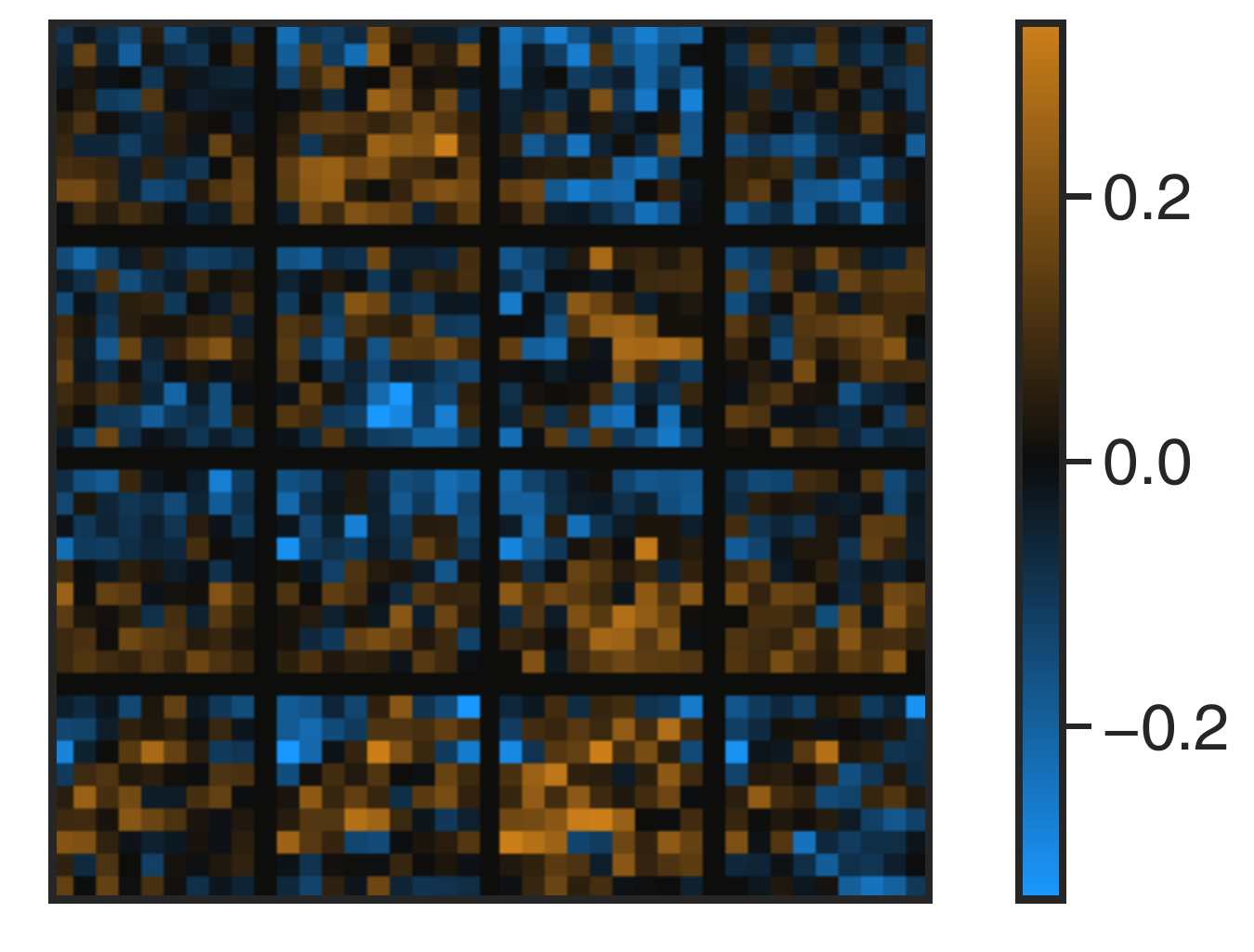} \\ 
        \rotatebox{90}{\hspace{5ex}Layer 0} &
         \includegraphics[width=0.3\textwidth]{figures/kernels/kernel_layer_0_tcnn_0_lwa_output_ep_0.pdf} &
         \includegraphics[width=0.3\textwidth]{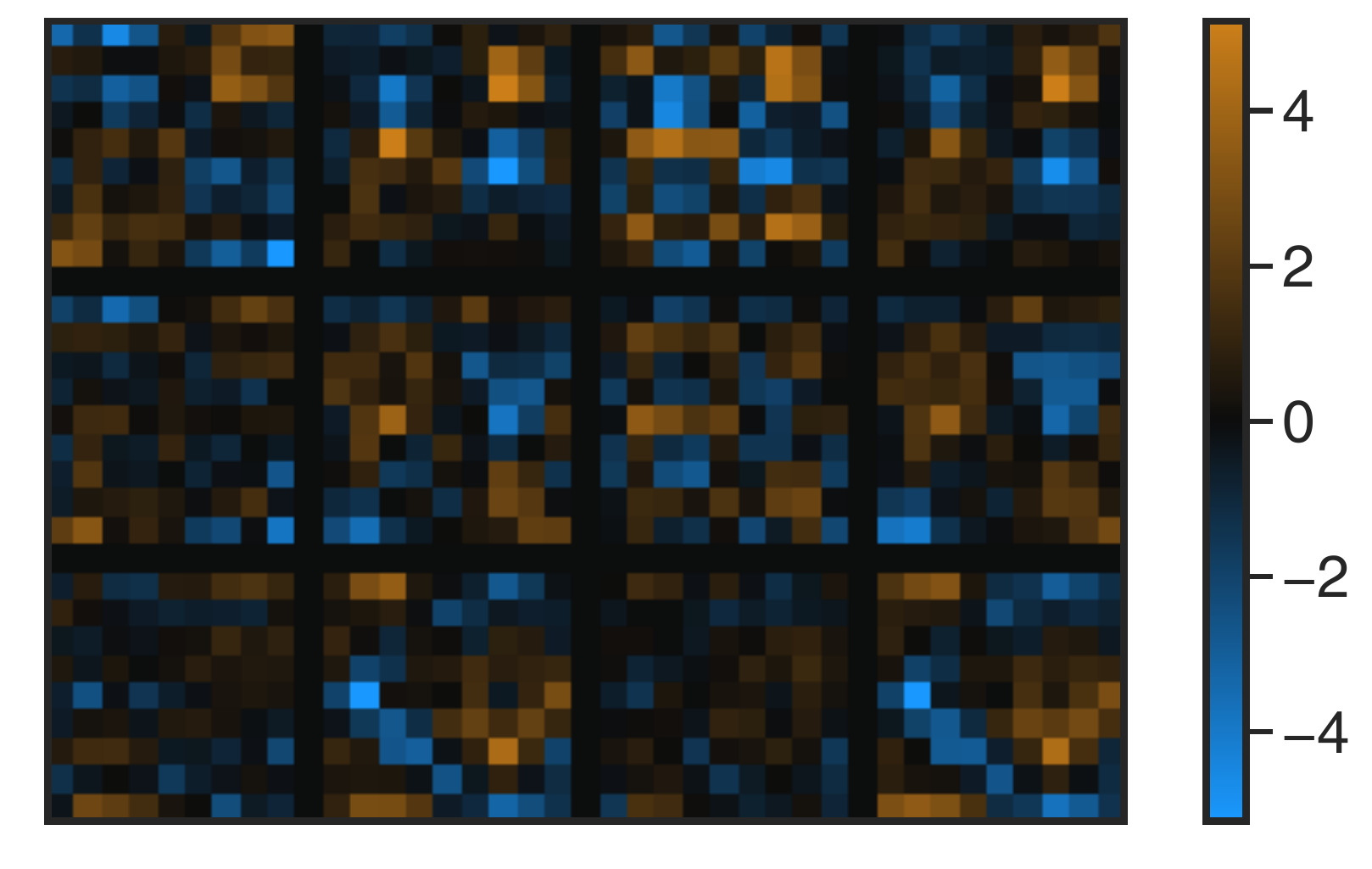} &
         \includegraphics[width=0.32\textwidth]{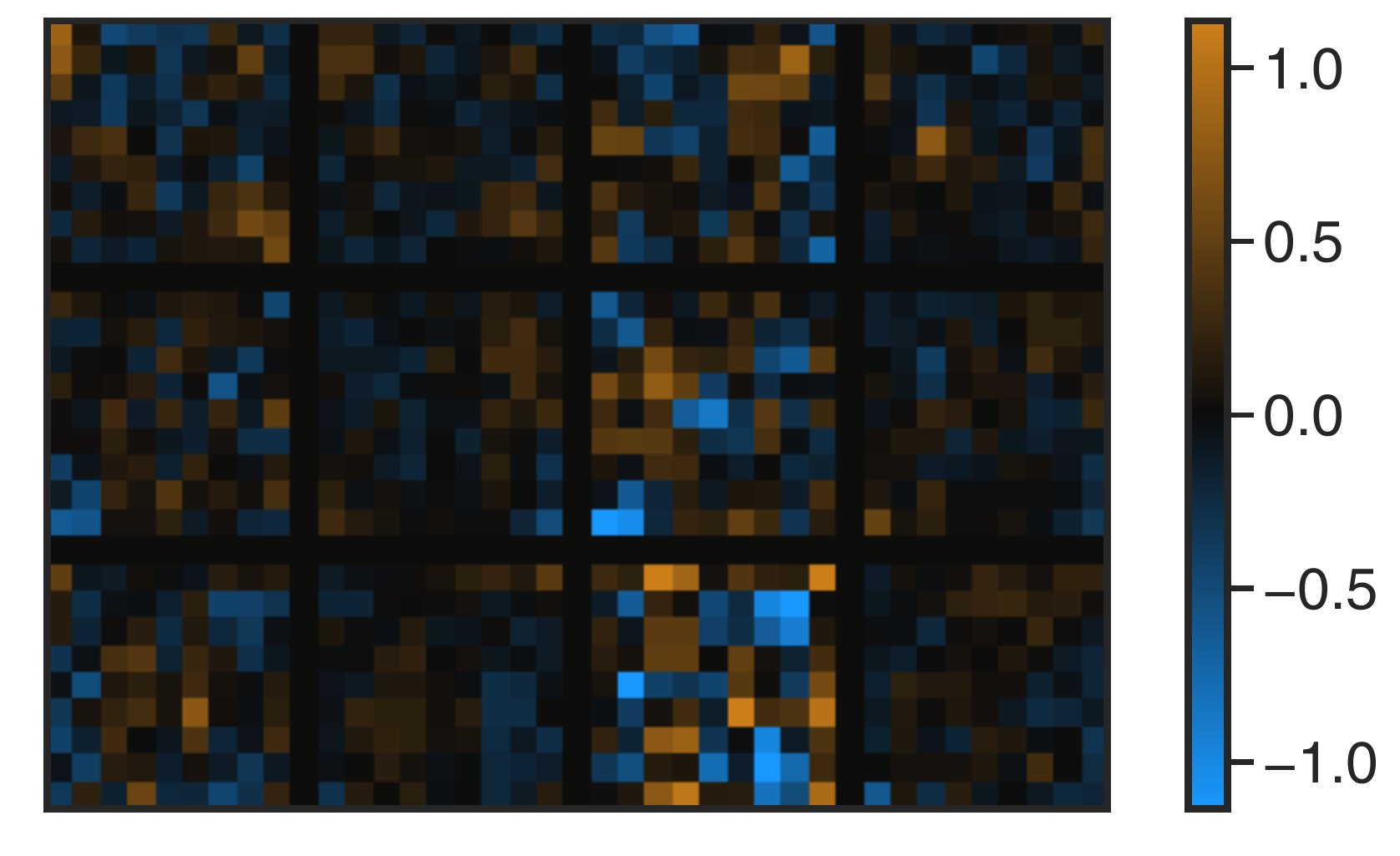}
        \end{tabular}
        \caption{Visualizing generated convolutional kernels in a 2-layer model for two different episodes. \emph{Left two plots:} kernels for two random episodes of 5 classes, \emph{right:} the difference in generated kernels for two episodes. Layer weight allocation: ``output''.}
        \label{fig:conv-kernelsepisode-diff-output}
    \end{figure}

\begin{figure}
        \centering
        \begin{tabular}[c]{@{\hspace{0.0\linewidth}}c@{\hspace{0.0\linewidth}}c@{\hspace{0.01\linewidth}}c@{\hspace{0.0\linewidth}}|c@{\hspace{0.0\linewidth}}}
         & Episode 0 & Episode 1 & Difference \\
        \rotatebox{90}{\hspace{5ex}Layer 1} &
         \includegraphics[width=0.32\textwidth]{figures/kernels/kernel_layer_1_tcnn_0_lwa_spatial_ep_0.pdf} &
         \includegraphics[width=0.32\textwidth]{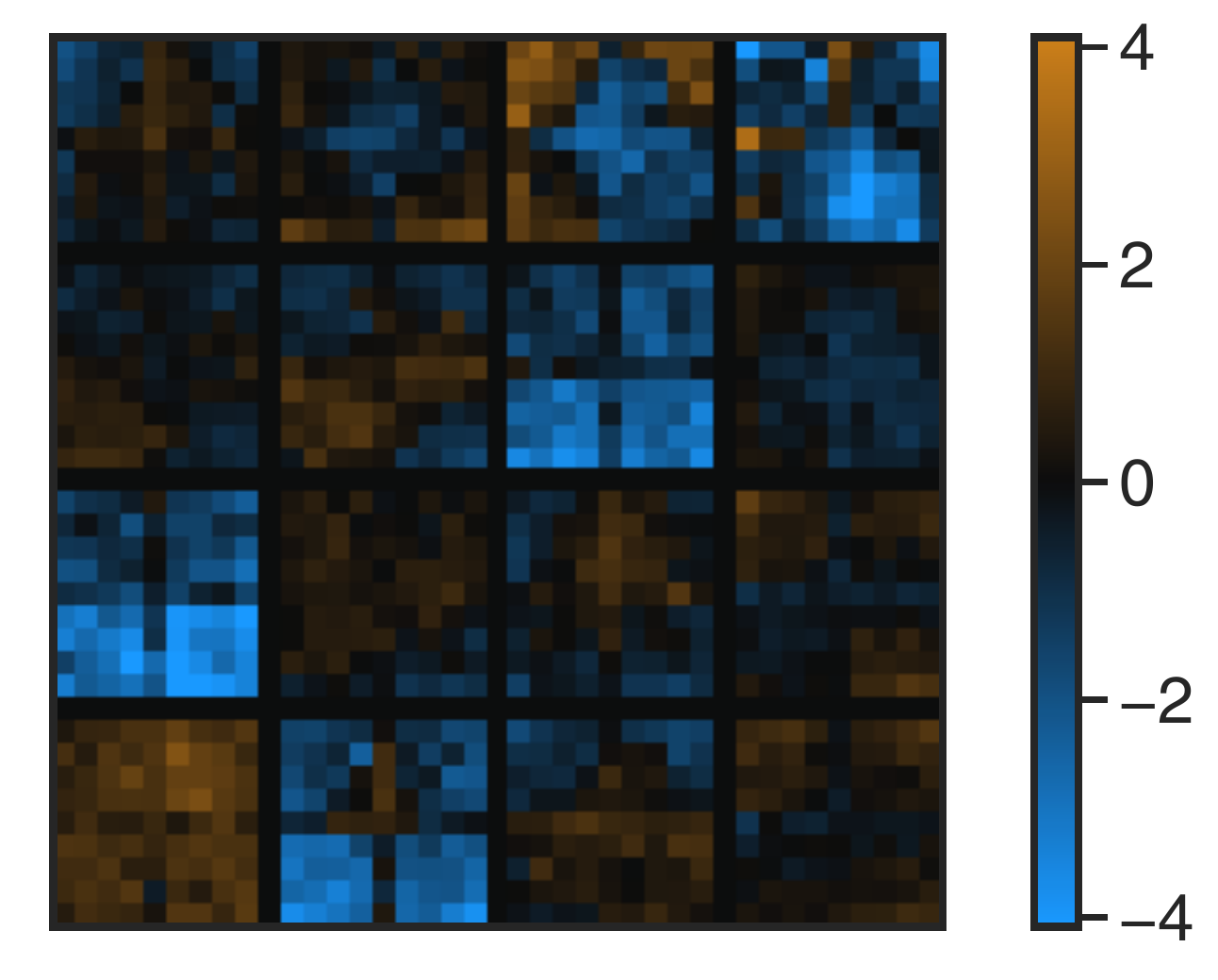} &
         \includegraphics[width=0.32\textwidth]{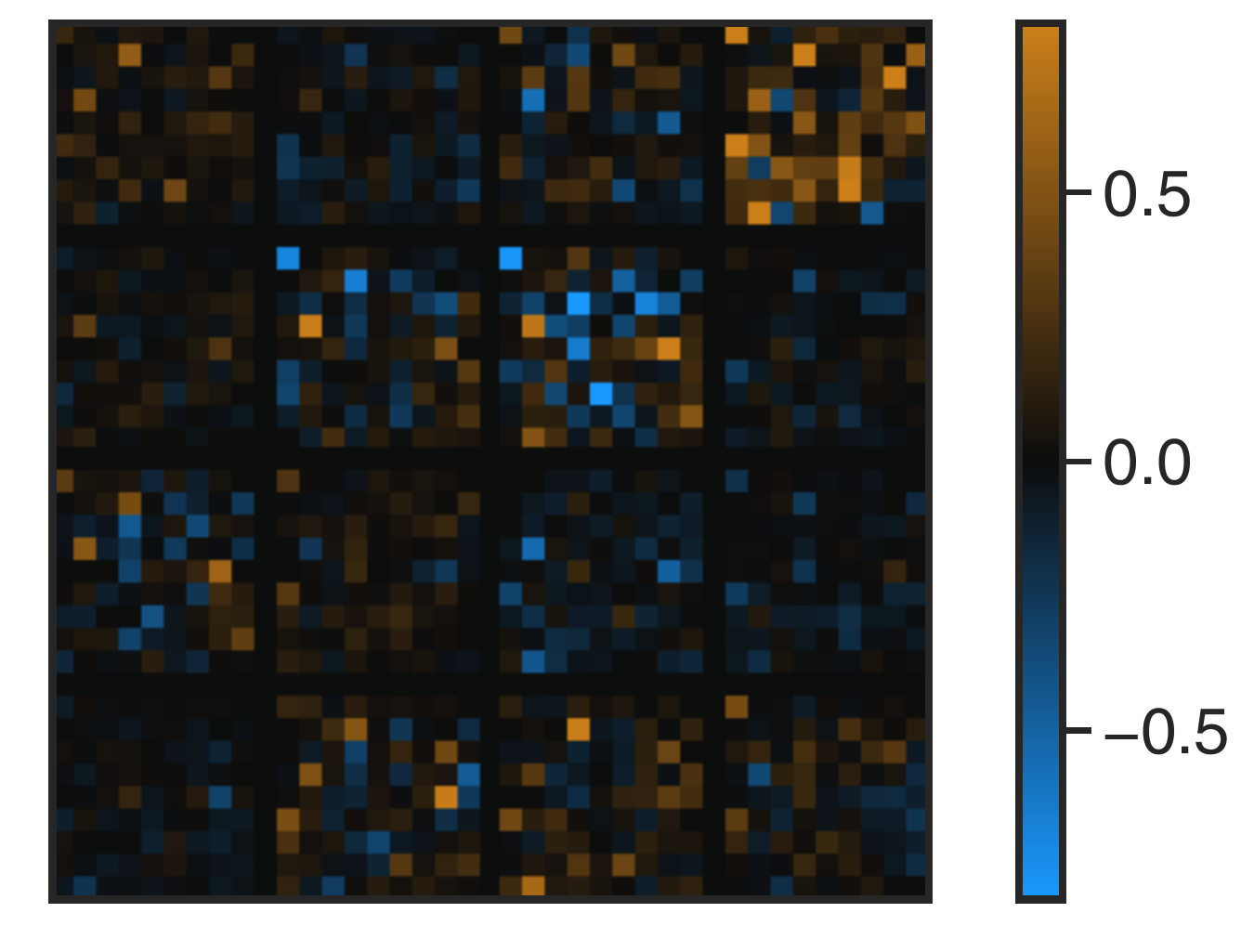} \\ 
        \rotatebox{90}{\hspace{5ex}Layer 0} &
         \includegraphics[width=0.3\textwidth]{figures/kernels/kernel_layer_0_tcnn_0_lwa_spatial_ep_0.pdf} &
         \includegraphics[width=0.3\textwidth]{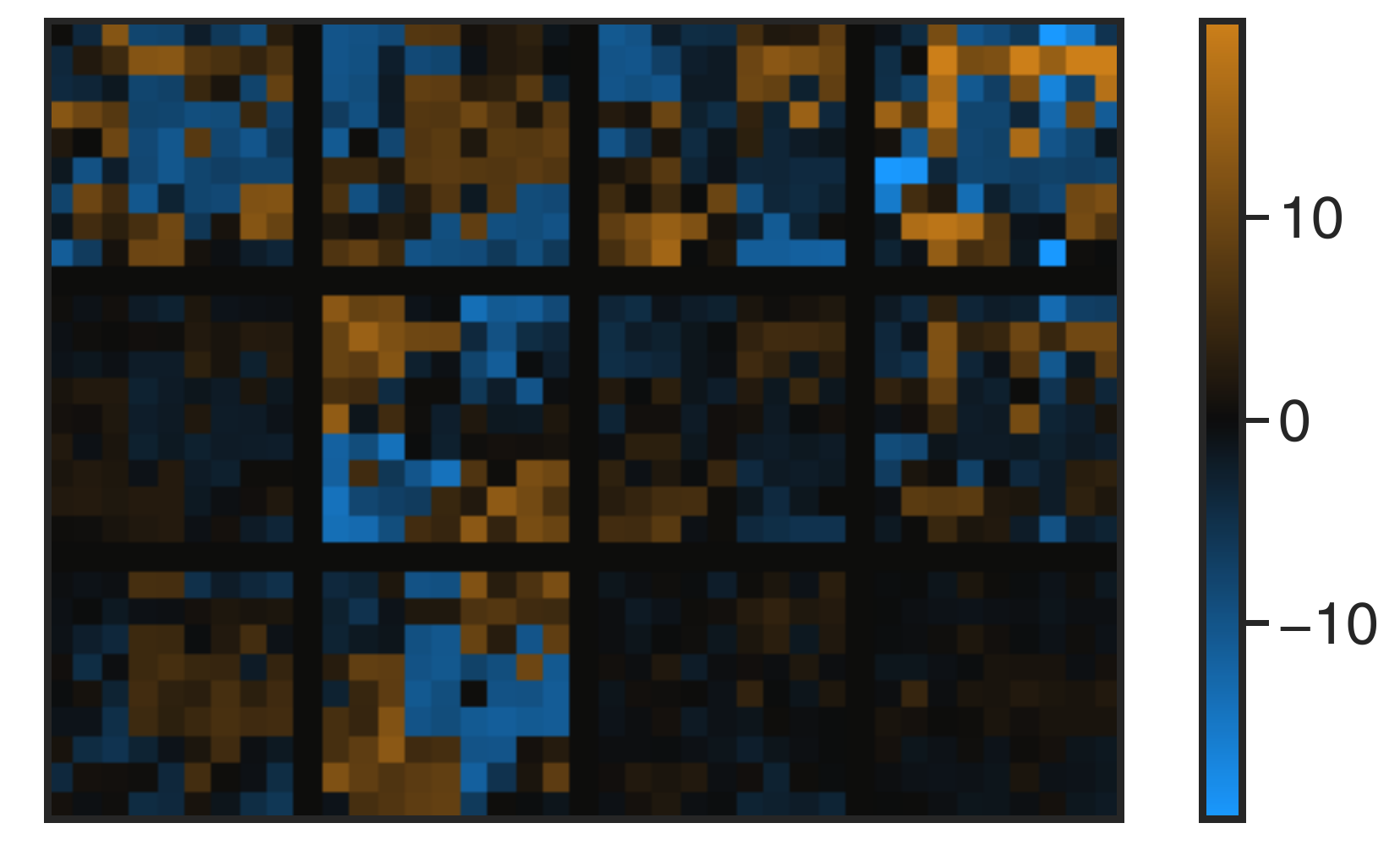} &
         \includegraphics[width=0.32\textwidth]{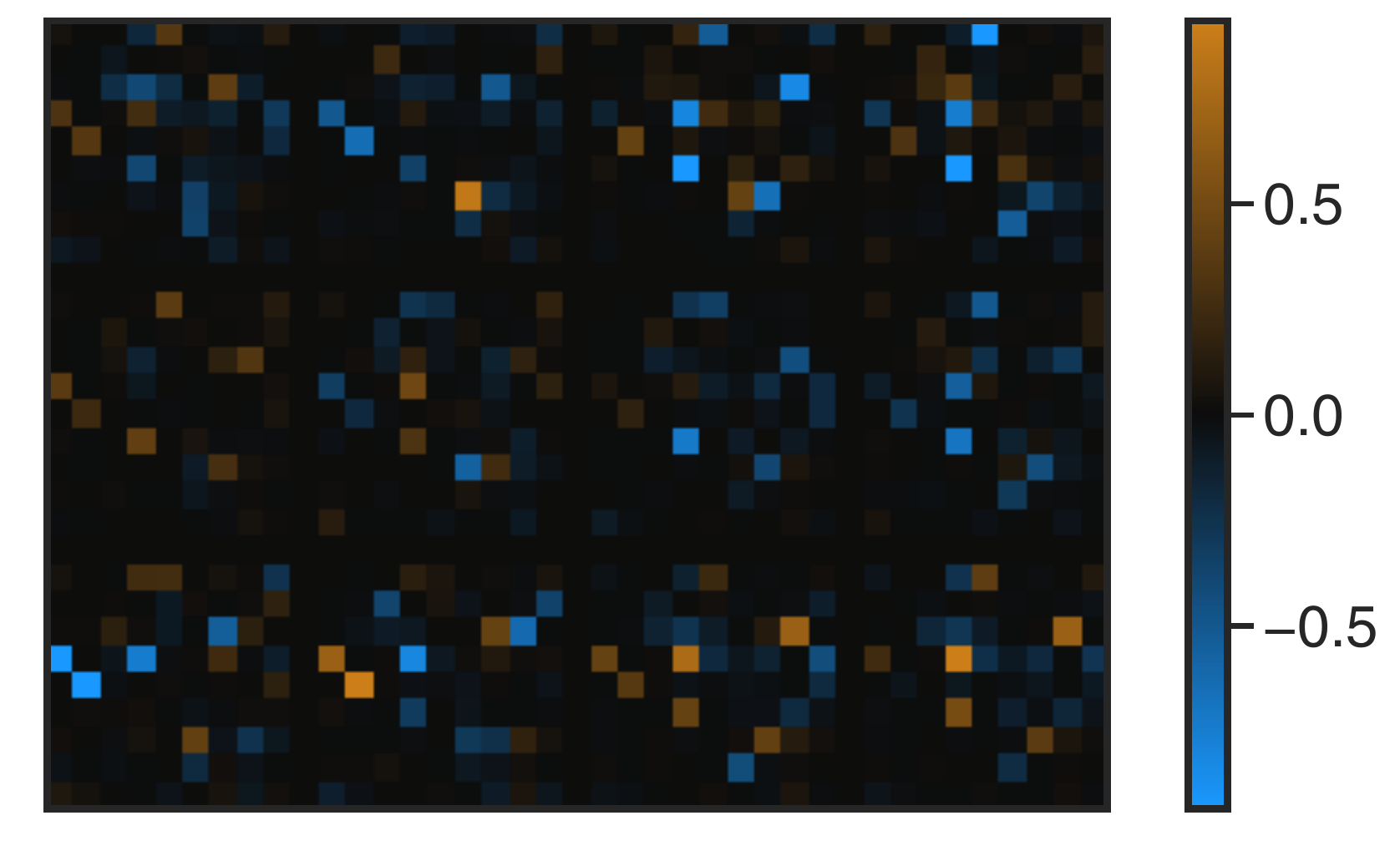}
        \end{tabular}
        \caption{Visualizing generated convolutional kernels in a 2-layer model for two different episodes. \emph{Left two plots:} kernels for two random episodes of 5 classes, \emph{right:} the difference in generated kernels for two episodes. Layer weight allocation: ``spatial''.}
        \label{fig:conv-kernelsepisode-diff-spatial}
    \end{figure}

    Trained layers are always fixed for the inference for all the episodes, but the generated layers vary, albeit not significantly.
    In Figures~\ref{fig:conv-kernelsepisode-diff-output} and \ref{fig:conv-kernelsepisode-diff-spatial} we show the generated kernels for two different episodes and, on the right, the difference between them.
    It appears that the generated convolutional kernel change withing $10-15\%$ form episode to episode.
        
\section{Additional Tables and Figures}

    \begin{table*}[h]
        \caption{
			Average model test and training accuracies on \omni{} (separated by a slash) for the models of different sizes.
			``Logits'' row shows accuracies for the model with only the fully-connected logits layer generated from the support set.
			It can be interpreted as a method based on a learned embedding.
			``All'' row reports accuracies of the models with some or all convolutional layers being generated.
			We were not able to see a statistically significant evidence of an advantage of generating more than one convolutional layers.
		}
        \label{tab:tiny}
        \begin{center} \begin{tabular}{l|c|c|c}
            & 4-channel & 6-channel & 8-channel \\
            \hline \\[-.8em]
            {\bf Logits} & $77.9$ / $79.2$ & $90.0$ / $91.4$ & $94.4$ / $95.8$ \\
            {\bf All} & $82.0$ / $83.4$ & $90.7$ / $92.0$ & $94.6$ / $96.0$
        \end{tabular} \end{center}
    \end{table*}

\end{document}